\documentclass{article}

\usepackage{microtype}
\usepackage{graphicx}
\usepackage{subcaption}
\usepackage{booktabs} 

\usepackage{hyperref}

\usepackage[preprint]{icml2026}

\usepackage{amsmath}
\usepackage{amssymb}
\usepackage{mathtools}
\usepackage{amsthm}

\usepackage{caption}
\usepackage{multirow}
\usepackage{subfiles}
\usepackage{array}
\usepackage{graphicx}
\usepackage{wrapfig}
\usepackage{booktabs} 
\usepackage{subcaption}
\usepackage{makecell}
\usepackage{adjustbox}
\usepackage{wrapfig}
\usepackage{tabularx}

\usepackage[capitalize,noabbrev]{cleveref}

\theoremstyle{plain}

\theoremstyle{definition}

\theoremstyle{remark}

\usepackage[textsize=tiny]{todonotes}

\begin{document}

\twocolumn[
  
 \icmltitle{Measuring Dataset Diversity from a Geometric Perspective}




  \icmlsetsymbol{equal}{*}

  \begin{icmlauthorlist}
    \icmlauthor{Yang Ba}{yyy}
    \icmlauthor{Mohammad Sadeq Abolhasani}{yyy}
    \icmlauthor{Michelle V Mancenido}{comp}
    \icmlauthor{Rong Pan}{yyy}
  \end{icmlauthorlist}

  \icmlaffiliation{yyy}{School of Computing and Augmented Intelligence, Arizona State University}
  \icmlaffiliation{comp}{School of Mathematical and Natural Sciences}

  \icmlcorrespondingauthor{Yang Ba}{yangba@asu.edu}

  \icmlkeywords{Machine Learning, ICML}

  \vskip 0.3in
]



\printAffiliationsAndNotice{}  

\begin{abstract}
Diversity can be broadly defined as the presence of meaningful variation across elements, which can be viewed from multiple perspectives, including statistical variation and geometric structural richness in the dataset. 
Existing diversity metrics, such as feature-space dispersion and metric-space magnitude, primarily capture distributional variation or entropy, while largely neglecting the geometric structure of datasets. To address this gap, we introduce a framework based on topological data analysis (TDA) and persistence landscapes (PLs) to extract and quantify geometric features from data.
This approach provides a theoretically grounded means of measuring diversity beyond entropy, capturing the rich geometric and structural properties of datasets. Through extensive experiments across diverse modalities, we demonstrate that our proposed PLs-based diversity metric (PLDiv) is powerful, reliable, and interpretable, directly linking data diversity to its underlying geometry and offering a foundational tool for dataset construction, augmentation, and evaluation.
\end{abstract}

\section{Introduction}

Life itself depends on diversity, as an ecosystem may collapse when a few species vanish, yet a single new species may reshape balance by either enriching resilience or triggering instability. In machine learning and artificial intelligence, data diversity plays a similar role.
Studying diversity has long been a central concern throughout the ML/AI life cycle, particularly in data collection to ensure representational balance, in data and model evaluation for assessing fairness and robustness \citep{rolf2021representation,clemmensen2022data,pmlr-v258-kim25f}, in model training to prevent overfitting, and in model generalization to reduce the gap between training distributions and real-world deployment \citep{wang2020improving, bian2021does,  yu2022can, ortega2022diversity, liu2023can}.
It is well known that exposure to a wide range of data structures, styles, and semantic patterns supports the learning of more abstract, transferable representations, allowing for more capable and resilient models \citep{ zhang2017mixup, shorten2019survey, NEURIPS2021_fb4c4860}. 
Recent work further demonstrates that diversity in training data influences the weight matrices of neural networks, directly affecting both in-distribution and out-of-distribution performance \citep{ba2024does}.

\begin{figure*}[t]
    \centering
    \includegraphics[width=1\linewidth]{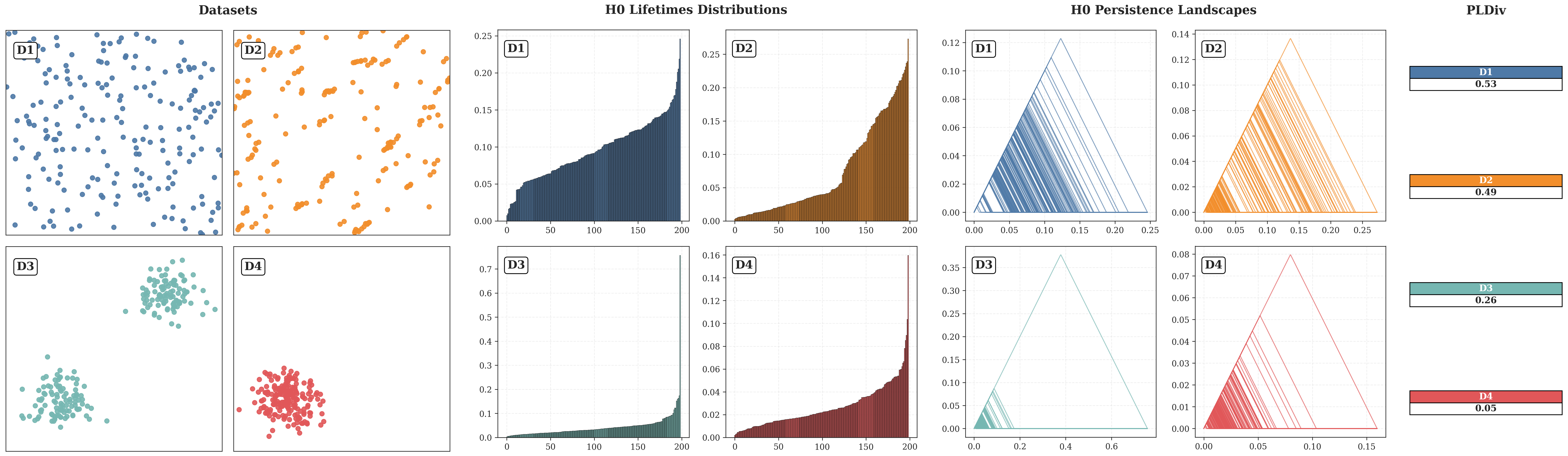}
    \caption{Illustration of PLDiv on four synthetic datasets.
D1: uniformly scattered points; D2: less evenly spread distribution; D3: two separated clusters; D4: a single compact cluster with minimal diversity. 
We extract $H_0$ features via persistent homology, where lifetimes measure how long clusters persist before merging with their closest neighbors. Persistence landscapes capture these patterns, and PLDiv, defined as the sum of their integrals, reflects both scale and persistence, aligning with the datasets’ decreasing diversity.}
    \label{fig:pipeline}
\end{figure*}

Yet beyond traditional performance measures, there is arguably a more urgent motivation to study dataset diversity. Today's generative models are trained on overlapping, internet-scale corpora, then reused and adapted across various applications. As these models are increasingly integrated into real-world writing, content creation, visual and audio materials, and codes, their outputs feed back into the very data streams that will train the next generation of models. Recent studies show that alignment-tuned models such as InstructGPT already exhibit significant reductions in lexical and conceptual diversity \citep{padmakumar2023does}. This homogenization is self-reinforcing i.e., models trained on uniform outputs further reinforce uniformity in subsequent models \citep{bertrand2023stability,alemohammad2024self, jiang2025artificial}. These risks are not limited to text generation, as generative models use the same sources on the Internet, standardized pipelines, and optimization objectives across many data modalities.

At this stage, data diversity is not just a desirable property, but a necessity for innovative, responsible, and human-centered AI design. The challenge of addressing data diversity across the ML/AI life cycle requires a theoretically grounded, empirically measurable definition of what constitutes substantive diversity. Reliable measurement allows the detection of latent homogenization effects in generative models and the design of appropriate interventions such as diversity-aware data collection, synthetic data generation, data augmentation strategies, and dataset–task alignment.

Approaches for quantifying diversity include the Vendi score \citep{dan2023vendi}, a metric inspired by ecological and biological models \citep{daly2018ecological, leinster2021entropy}. More recently, researchers have proposed metrics that operate on similarity matrices, using either aggregated similarity magnitudes  \citep{limbeck2024metric} or probability distributions over pairwise similarities \citep{zhu2025measuring}. While these metrics capture some aspects of similarities in representation space, they do not explicitly model the intrinsic geometry of the data manifold, such as local structures or spatial organization.

In contrast, we posit a closer connection between the geometric structure of data and its diversity. In particular, fundamental geometric properties such as curvature have been shown to have an impact on dataset diversity:
positive curvature, as on a sphere, compresses points and restricts possible configurations, while negative curvature, as in hyperbolic geometry, spreads space out faster, enabling richer variation \citep{limbeck2024metric}. Topological data analysis (TDA) provides tools for capturing the shape of data and encoding its structural geometry. Recognizing the connection between the persistent homology (PH) merging process \citep{edelsbrunner2002topological, edelsbrunner2008persistent} and agglomerative hierarchical clustering \citep{murtagh2012algorithms}, we use a vectorized representation of PH called persistence landscapes (PLs) to propose a novel, geometry-aware measure of dataset diversity that we refer to as \textbf{persistence landscapes-based diversity (PLDiv)}. Fig.\ref{fig:pipeline} illustrates the estimation of PLDiv as the cumulative integral of the PL tent functions. 

In this work, we establish the theoretical grounding of PLDiv in topological principles, demonstrate its ability to measure substantive diversity in practice, and show that it facilitates clear and intuitive interpretations. To the best of our knowledge, this work is the first to leverage topological data analysis as a principled framework for measuring dataset diversity. Our specific contributions are threefold: 
\begin{itemize}
    \item A geometry-aware definition of data diversity: we introduce a principled framework for measuring data diversity that explicitly accounts for the geometric and topological structure of datasets. By leveraging persistent homology and persistence landscapes, our approach captures structural variation that is not accessible to purely distributional measures.
    \item A new topological diversity metric with theoretical grounding: we propose PLDiv, a persistence landscapes–based diversity measure and show that it satisfies core axioms of diversity \cite{leinster2012measuring}, including effective size, redundancy invariance, symmetry, and multi-scale consistency. 
    \item Empirical evidence linking data geometry to ground-truth data diversity: through extensive experiments on synthetic data sets and high-dimensional text and image embeddings, we demonstrate that PLDiv reliably captures substantive dataset diversity and outperforms existing alternatives. 
\end{itemize}

\section{Related Work}

\subsection{Diversity Measurement}



Several reference-based metrics compare generated data with human or gold-standard corpora. The Fréchet Inception Distance (FID) \citep{heusel2017gans} and related Inception Score were among the first to use pretrained embeddings to measure alignment between real and synthetic data distributions. More recently, MAUVE \citep{pillutla2021mauve} quantified distributional gaps between model and human text, while precision–recall metrics \citep{kynkaanniemi2019improved,bronnec2024exploring} provided a decomposition into fidelity (precision) and diversity (recall). Extensions such as density and coverage metrics \citep{naeem2020reliable} improved robustness against outliers and unstable density estimates. Nevertheless, these methods are fundamentally tied to reference datasets, often entangle fidelity with diversity, and remain sensitive to embedding choices or manifold approximations.

A different line of work has explored representation-level measures that aim to be reference-free. Early proposals such as diversity, density, and homogeneity \cite{lai2020diversity} assessed dispersion in embedding spaces, but they remained limited to simple distributional statistics. More principled approaches emerged with entropy- or kernel-based methods: the Vendi Score \citep{dan2023vendi} measures diversity as the exponential of Shannon entropy derived from the similarity spectrum, while Renyi Kernel Entropy (RKE) and its variant RRKE \citep{jalali2023information} extend this perspective using quantum information theory. However, such approaches often require expensive eigenvalue or singular-value decompositions, limiting their scalability to large datasets. Building on efficiency and separability, DCScore \citep{zhu2025measuring} reframes diversity measurement as a classification problem, avoiding eigenvalue computations and yielding faster, more scalable estimates. Complementary to this, magnitude-based methods \citep{limbeck2024metric} quantify effective dataset size across scales, offering metrics such as MAGAREA (reference-free) and MAGDIFF (reference-based). While these methods provide multi-scale summaries, they depend on tuning scale parameters and still abstract away the geometric or topological structures that can differentiate datasets with the same dispersion.

\subsection{Persistent Homology}

Persistent Homology (PH) \citep{edelsbrunner2002topological,edelsbrunner2008persistent} is a central tool in TDA for uncovering the underlying shape of data, typically represented as point clouds. By constructing nested simplicial complexes across scales and applying homology, PH tracks the birth and death of topological features such as connected components, loops, and voids. The result is a multi-scale summary, often visualized as barcodes or persistence diagrams, which distinguishes significant long-lived features from noise and is provably stable to perturbations. 

Building on these foundations, subsequent efforts have explored scalar invariants and geometric inference from persistence. \citet{govc2021persistent} introduced persistent magnitude, a signed, exponentially weighted sum over barcode intervals that refines classical magnitude theory. This approach provides interpretable scalar summaries encoding geometric complexity, including curvature, but it compresses the full topological signature into a single number, limiting its ability to capture heterogeneity or higher-order organization. In parallel, \citet{bubenik2020persistent} demonstrated that persistence can recover curvature information from sampled manifolds by combining diagrams with persistence landscapes, showing that even short-lived features carry meaningful geometric signals. While powerful, this line of work primarily targets smooth continuous geometry rather than irregular or combinatorial variation common in real-world datasets. Together, these directions underscore the expressive capacity of PH, yet also highlight an open gap: existing uses either oversimplify persistence or focus narrowly on geometric inference, leaving the systematic role of PH in quantifying dataset diversity underexplored.

\section{Preliminaries}
\subsection{Persistence Diagrams}

\begin{figure*}
    \centering
    \includegraphics[width=1\textwidth]{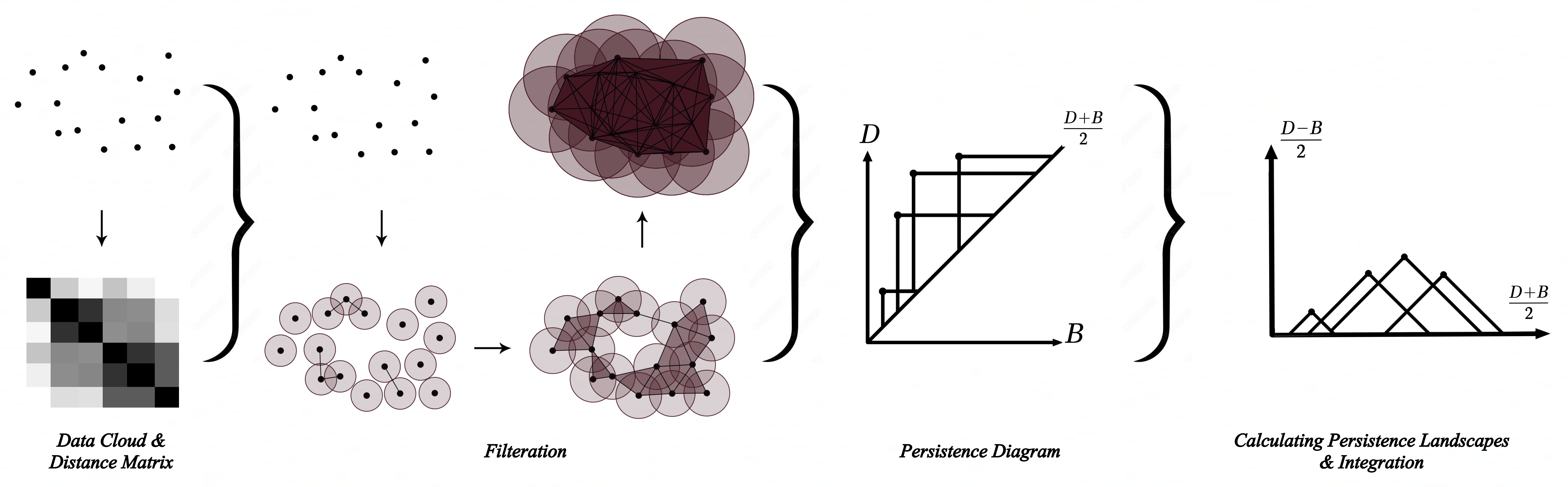}
    \caption{
    The pipeline of PLDiv. Using a data cloud or its distance matrix, we build a filtration of simplicial complexes and track the birth and death of $H_0$ components by persistent homology. The resulting persistence diagram is then used to calculate persistence landscapes. Lastly, PLDiv is obtained by integrating these landscapes and serves as a metric for the dataset diversity.}
    \label{fig:ph}
\end{figure*}

PH provides a multiscale description of the topological structure of data. 
Starting from a point cloud $\mathcal{X} = \{x_1, \dots, x_n\}$, it builds a nested sequence of simplicial complexes (a filtration), such as the Vietoris--Rips filtration. 
This filtration can be understood as growing balls (or ``bubbles'') of radius $r$ around each data point and increasing $r$ gradually. 
As the radius grows, the bubbles begin to overlap, creating higher-dimensional simplices (see Fig. \ref{fig:ph}). 
In this process, new topological features such as connected components, loops, and voids appear and eventually vanish when the bubbles merge or fill in. 
This viewpoint highlights that persistent homology captures how the topology of the data evolves across scales of the underlying radius parameter.

Formally, each topological feature is associated with a birth time $b_i$, the smallest radius at which it appears, and a death time $d_i$, the radius at which it disappears (for instance, when two connected components merge or when a loop becomes filled). 
The difference $\ell_i = d_i - b_i$ is called the \emph{lifetime} (or persistence) of the feature and quantifies its robustness across scales. 

The output of persistent homology is summarized in a \emph{persistence diagram}, defined as the multiset
\[
\mathcal{D} = \{ (b_i, d_i) \}_{i=1}^m, \quad b_i < d_i,
\]
where each point $(b_i, d_i)$ represents the birth and death scales of a feature. 
The diagram is typically plotted in the plane $\mathbb{R}^2$, with each feature as a point above the diagonal $b = d$. 
Features with long lifetimes (points far from the diagonal) are often interpreted as meaningful structural signals in the data, while short-lived features (points near the diagonal) are commonly attributed to noise. 
Persistence diagrams thus provide a compact and interpretable summary of the multiscale topological properties of the dataset.


\subsection{Persistence Landscapes}

Although persistence diagrams provide a geometric summary of topological features, they are multisets, represented by points on a plane, which makes it challenging to apply classical statistical and machine learning techniques directly. 
To address this problem, \citet{bubenik2015statistical} introduced \emph{persistence landscapes}, a functional summary of persistent homology that embeds the information of a persistence diagram into a Banach space, enabling the use of standard statistical tools.

Given a persistence diagram $\mathcal{D} = \{ (b_i, d_i) \}_{i=1}^m$,
we first associate each birth-death pair $(b_i, d_i)$ with a piecewise linear ``tent'' function.

\[
\lambda_{(b,d)}(t) =
\begin{cases}
t-b, & b \leq t \leq \tfrac{b+d}{2}, \\
d-t, & \tfrac{b+d}{2} < t \leq d, \\
0, & \text{otherwise}.
\end{cases}
\]
This function attains its maximum value, $\tfrac{d_i - b_i}{2}$, at the midpoint of the interval.
The persistence landscape is then defined as the sequence of functions
\[
\lambda_k(t)
= \text{$k$-th largest of }
\{\lambda_{(b_i,d_i)}(t)\}_{i=1}^m,
\quad k=1,2,\dots
\]
for each $t \in \mathbb{R}$. 
Thus, $\lambda_1$ records the largest ``tent'' value at each $t$, $\lambda_2$ records the second largest, and so forth. 
Collectively, the functions $\{\lambda_k\}_{k \geq 1}$ constitute the persistence landscape.

Persistence landscapes inherit stability from persistence diagrams and have the advantage of lying in the $L^p$ function space. The persistence landscape is a vectorized form of a persistence diagram, equivalent to a 45° rotation that preserves all information, with X = $(d+b)/2$ and Y = $(d-b)/2$ (see Fig. \ref{fig:ph}).


\section{Methodology}

\subsection{Diversity Measure via Persistence Landscapes}


\textbf{Definition 4.1.} Let $\mathcal{X} = \{x_1, \dots, x_n\}$ be a dataset and let $\Lambda(\mathcal{X}) = \{ \lambda_k \}_{k \geq 1}$
denote its persistence landscape obtained from persistent homology. 
The \emph{persistence landscapes based diversity} score, PLDiv$(\mathcal{X})$, is defined as
\begin{equation}
\mathrm{PLDiv}(\mathcal{X}) 
= \sum_{k=1}^\infty \int_{\mathbb{R}} \lambda_k(t) \, dt.
\end{equation}
The summation is typically finite, as only a finite number of $\lambda_k$  terms are actually non-zero.
PLDiv$(\mathcal{X})$ measures the cumulative ``area under the triangles'' of the persistence landscape and quantifies the richness of topological features across all scales.

\textbf{Proposition 4.2.} A closed form of PLDiv can be derived. Let $\mathcal{D}=\{(b_i,d_i)\}_{i=1}^m$ be the set of birth--death pairs produced by persistence homology, then
\[
\begin{aligned}
\mathrm{PLDiv}(\mathcal{X})
&= \sum_{k=1}^\infty \!\int_{\mathbb{R}} \lambda_k(t)\,dt
 = \sum_{i=1}^m \!\int_{\mathbb{R}} \lambda_{(b_i,d_i)}(t)\,dt \\
&= \frac{1}{4}\sum_{i=1}^m (d_i-b_i)^2 .
\end{aligned}
\]

\textit{Proof.} Each tent function with its supports on the interval $[b_i,d_i]$ is a symmetric isosceles triangle of base length $d_i-b_i$ and height $(d_i-b_i)/2$, hence its area is
\[
\int_{\mathbb{R}} \lambda_{(b_i,d_i)}(t)\,dt \;=\; \tfrac12\cdot (d_i-b_i)\cdot \tfrac{d_i-b_i}{2} \;=\; \frac{(d_i-b_i)^2}{4}.
\]

Summing them yields the closed form above. We provide a detailed proof in Appendix \ref{appex:b}.

\textbf{Remark 4.3.}
The area under $\lambda_k$ measures both the \emph{scale} and the \emph{persistence} of topological features, representing how long and how strongly features persist across scales. 
Summing across $k$ aggregates contributions across all topological structures, capturing both \emph{local fluctuations} (short lifetimes) and \emph{global connectivity} (long lifetimes).

\textbf{Remark 4.4.}
A large PLDiv$(\mathcal{X})$ indicates that features such as clusters or loops are well-separated and persist across scales, reflecting high structural diversity. Conversely, a smaller value corresponds to a dataset where data points collapse quickly into clusters, eliminating persistent features. In particular, by Proposition 4.2, PLDiv $(\mathcal{X})$ coincides with the second moment of lifetimes of topological features, up to scaling. 





\textbf{Remark 4.5.}
Since the persistence landscape lies in $L^p(\mathbb{R})$, the integral
$
\int_{\mathbb{R}} \lambda_k(t) \, dt
$
can be interpreted as the ``expected persistence'' of the $k$-th most prominent feature across random scales $t$. 
From the probabilistic perspective, PLDiv$(\mathcal{X})$  represents the total expected persistence across all topological features, analogous to computing an energy functional over the data manifold.

PLDiv$(\mathcal{X})$ should be understood as a holistic measure of dataset complexity. 
Unlike conventional approaches in topological data analysis that treat short-lived features as noise, this measure incorporates the full spectrum of topological features, emphasizing that both long- and short-lived structures contribute to the geometry of the data (follows the insights in \citet{turkes2022effectiveness}). 
In this sense, PLDiv$(\mathcal{X})$ provides a unified framework that balances mathematical rigor with interpretability.

In practice, there are many choices for the filtration and the degree of persistent homology. 
For most tasks, 0-dimensional persistent homology is sufficient, because it efficiently captures the connectivity structure of the dataset while keeping computational costs low. Therefore, our metric (PLDiv) is computed based on $H_0$ features in the following experiments.

\subsection{Axiomatic Properties of Diversity}

Among core diversity axiomatic properties provided by \citet{leinster2012measuring} and  \citet{leinster2021entropy}, our proposed diversity measure, PLDiv, satisfies four fundamental axioms: effective size, monotonicity, twin property, 
and symmetry. These axioms provide a foundation for reasonable and robust diversity evaluation. A description of these axioms is provided below, while the formal proofs of these properties on PLDiv are presented in Appendix \ref{appex:b}.


\begin{itemize}
    \item \textbf{Effective size.} 
    For a fixed number of points, PLDiv$(\mathcal{X})$ increases when data points are well-separated and decreases as they cluster, reaching a maximum when all points are distinct and a minimum when all are identical. (An illustration is presented in the Appendix \ref{appendix:tail}) 


    \item \textbf{Monotonicity.} 
    Decreasing similarity increases diversity. Fix $n$ and let $\mathcal{X}$ be a point cloud in a metric space. If all pairwise distances in $\mathcal{X}$ are scaled by a factor $\alpha$ (i.e. replace the metric $d(\cdot,\cdot)$ by $\alpha d(\cdot,\cdot)$), then


\[
\mathrm{PLDiv}(\alpha \mathcal{X})
\begin{cases}
> \alpha^{2}\,\mathrm{PLDiv}(\mathcal{X}), & \alpha>1, \\
< \alpha^{2}\,\mathrm{PLDiv}(\mathcal{X}), &  0 < \alpha < 1 .
\end{cases}
\]
    

    

    \item \textbf{Twin property.} Adding an exact duplicate of a point does not change PLDiv$(\mathcal{X})$. 
    The duplicate induces a trivial birth--death pair $(0,0)$, contributing zero to the diversity score.
    Let $\mathcal{X}$ be a dataset and let $x_i \in \mathcal{X}$. For the set $\mathcal{X}' = \mathcal{X} \cup \{x_n\}$ where $x_n = x_i$, the diversity is unchanged:
    \[
    \mathrm{PLDiv}(\mathcal{X}') = \mathrm{PLDiv}(\mathcal{X}).
    \]

    \item \textbf{Symmetry.} PLDiv is invariant to the ordering of data points (permutation invariance). 
    Since persistent homology depends only on the metric structure of $\mathcal{X}$ and PLDiv$(\mathcal{X})$ is computed from the multiset of intervals $\{(b_i,d_i)\}$, relabeling or reordering points does not affect the value of the score. 
    Let $\mathcal{X}=(x_1,\dots,x_n)$ be an ordered sequence of points and let $\pi$ be any permutation of $\{1,\dots,n\}$. For the permuted sequence $\mathcal{X}_\pi=(x_{\pi(1)},\dots,x_{\pi(n)})$, we have
    \[
    \mathrm{PLDiv}(\mathcal{X}_\pi) = \mathrm{PLDiv}(\mathcal{X}).
    \]
    
\end{itemize}

\section{Experiment \& Result}

\subsection{Diversity Assessment in Synthetic Data Clouds}
\label{sec:synthetic}

\begin{figure*}[!htb]
    \centering
    \includegraphics[width=\textwidth]{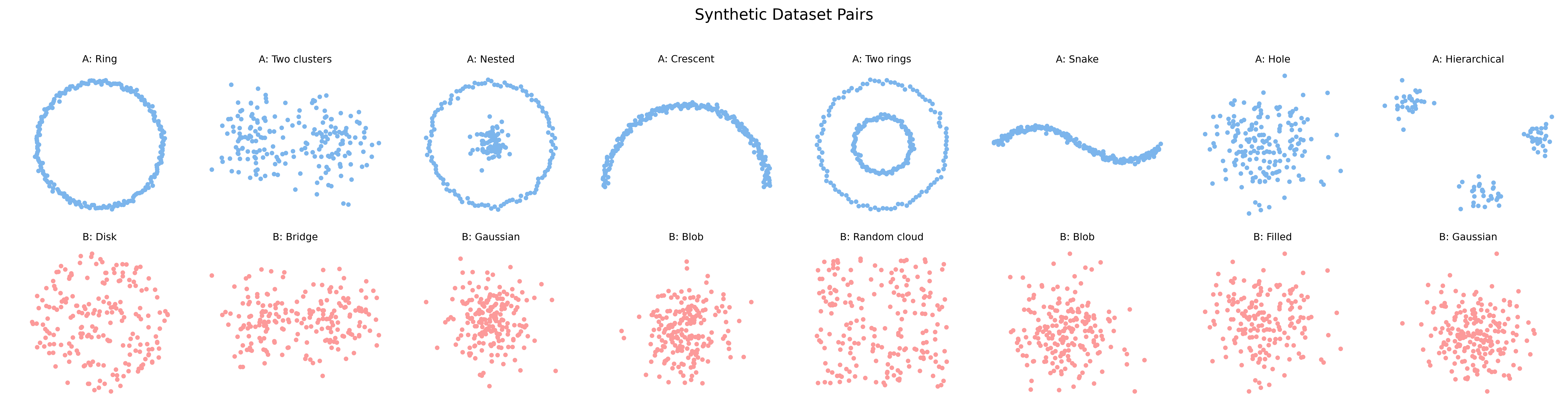}
    \includegraphics[width=\textwidth]{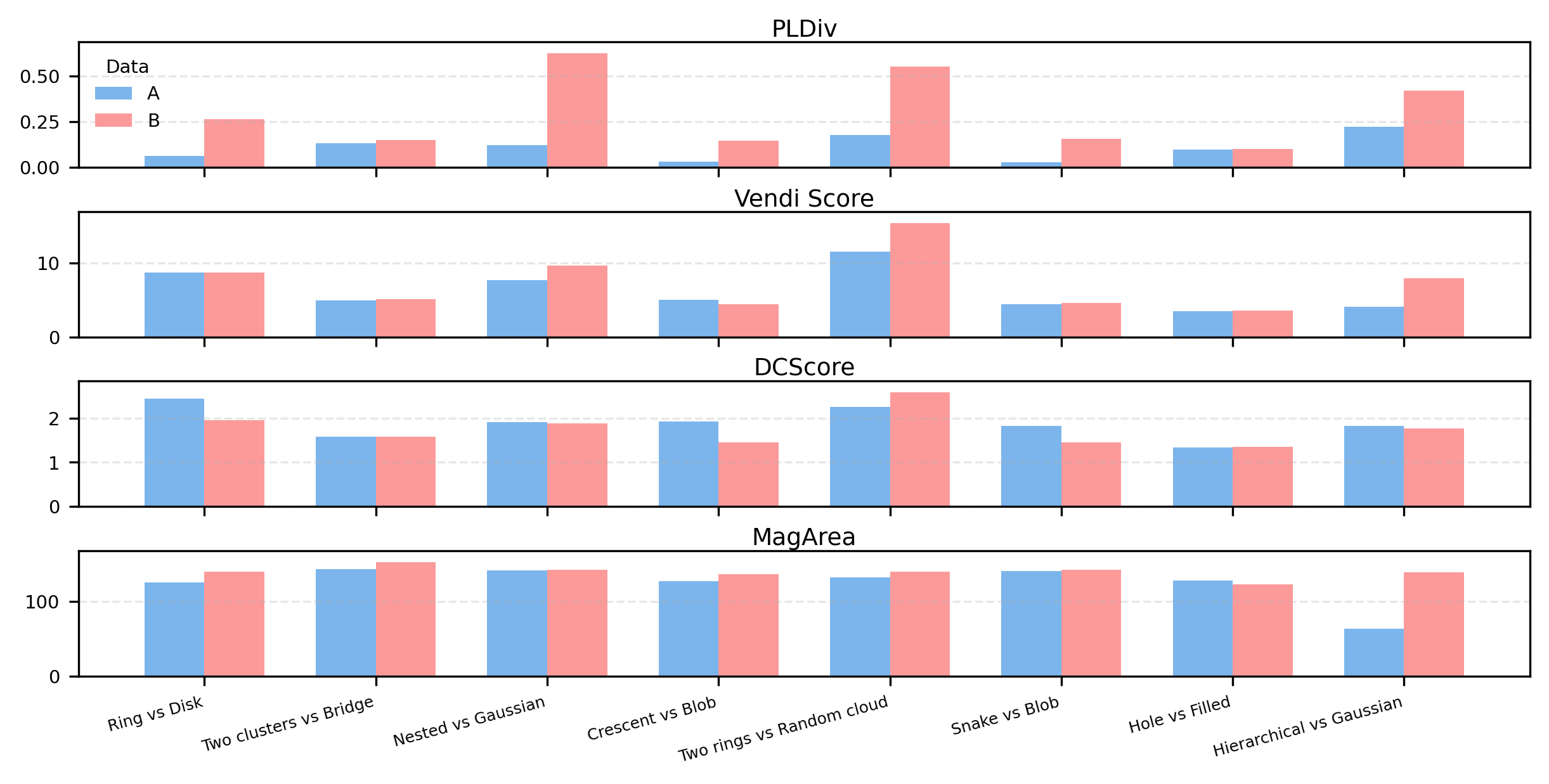}
    
    \caption{Synthetic dataset comparison. \textit{Upper:} eight dataset pairs (A vs.\ B), each with 200 points, generated to introduce or remove loops, bridges, or hierarchical clusters. \textit{Lower:} diversity scores across metrics. PLDiv yields sharper and more coherent distinctions that reflect the true geometric differences between datasets, while Vendi Score, DCScore, and MagArea respond mainly to overall spread and fail to capture these structural changes in most cases.}
    \label{fig:pairs}
\end{figure*}


To demonstrate PLDiv as an intrinsic geometry-aware diversity metric, we simulated eight pairs of two-dimensional point clouds (A, B), each containing about 200 points generated from parameterized geometric functions described in Appendix Table~\ref{tab:8datasets}. Each pair modifies one specific geometric property by adding or removing loops, bridges, curvature, or hierarchical clusters, while maintaining a comparable overall spatial scale. Scenario B is designed to be more diverse than Scenario A per pair. These controlled scenarios allow a direct comparison of how different metrics respond to structural variation rather than random dispersion.

We computed PLDiv, Vendi Score, DCScore, and MagArea on Euclidean distance matrices for each dataset. A metric is considered \emph{consistent} if it assigns a higher diversity value to the configuration exhibiting richer geometric organization. PLDiv meets this criterion across all eight cases, with Vendi Score and MagArea in seven and DCScore in only three. Moreover, PLDiv produces sharper and directionally coherent contrasts between paired clouds. For instance, \textit{Ring vs Disk} and \textit{Nested vs Gaussian} exhibit strong PLDiv separation that quantitatively reflects the presence or loss of loops, whereas the other metrics change only slightly. The difference arises from what each measure encodes: Vendi Score and DCScore emphasize global similarity spectra or density separation, and MagArea summarizes scale magnitude but not connectivity. PLDiv, by integrating the persistence of topological features across filtrations, captures geometrically meaningful and visually intuitive differences, as illustrated in Fig.~\ref{fig:pairs}.

\subsection{Characterizing Geometry with Curvature}

As a fundamental property in geometry, curvature quantifies the extent to which a manifold deviates from being flat, thereby governing the behavior of distances within that space.
Curvature inherently relates to diversity \citep{limbeck2024metric}: On positively curved spaces, such as spheres, data points concentrate and the variety of configurations is reduced; while on negatively curved spaces, such as hyperbolic disks, distances spread apart more quickly, creating a greater range of possible arrangements. Being able to recover curvature from point clouds is a principled and theoretically solid approach to validate whether a diversity measure is geometry-aware, rather than relying solely on pairwise dissimilarities. This is important because modern representation learning often places data in non-Euclidean spaces, such as spherical or hyperbolic embeddings, where curvature plays a key role in structuring similarity. A diversity measure sensitive to curvature better represents the data manifold’s geometry.


\begin{table}[h]
\centering
\caption{PLDiv estimates curvature}
\label{tab:curvature}
\scalebox{0.9}{%
\begin{tabular}{ll}
\toprule
Method & MSE ($\downarrow$) \\
\midrule
SVR(Vendi Score, L1 kernel) & 0.229 $\pm$ 0.042 \\
SVR(Vendi Score, RBF kernel) & 0.053 $\pm$ 0.004 \\
SVR(DCScore, L1 kernel) & 0.134 $\pm$ 0.019 \\
SVR(DCScore, RBF kernel) & 0.052 $\pm$ 0.004 \\
SVR(MAGAREA, Euclidean) & 0.120 $\pm$ 0.010 \\
\midrule
\textbf{SVR(PLDiv)} & \textbf{0.039} $\pm$ \textbf{0.001} \\
\textbf{SVR(Sparse PLDiv)} & \textbf{0.040} $\pm$ \textbf{0.001} \\
\bottomrule
\end{tabular}%
}
\end{table}

\begin{figure*}[!ht]
    \centering
    \includegraphics[width=1.0\linewidth]{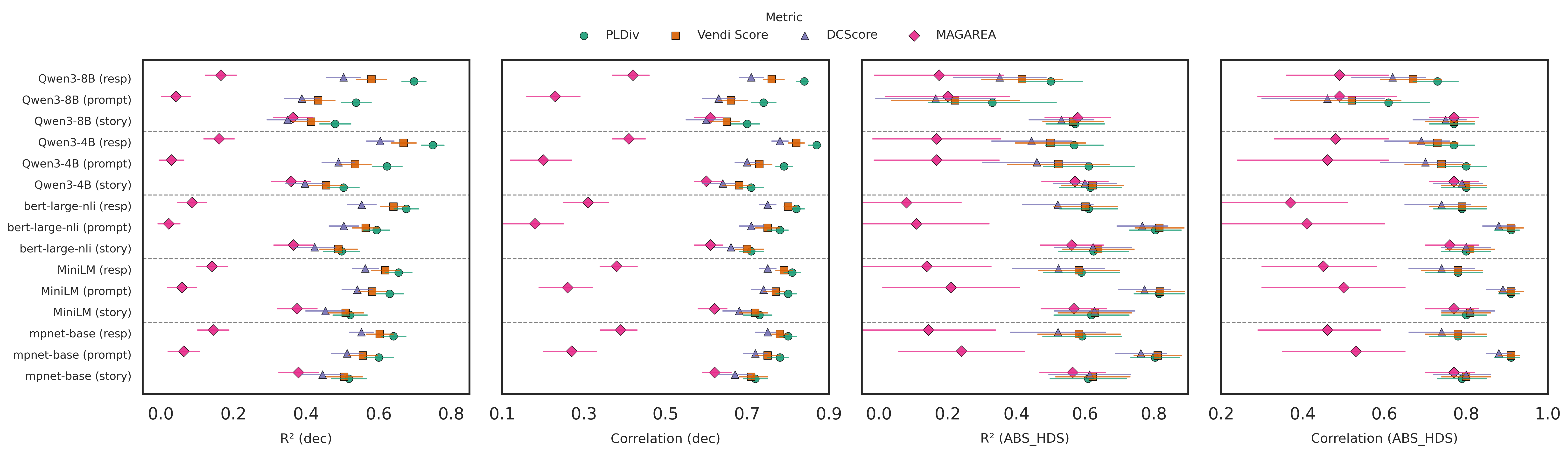}
    \caption{Demonstration that PLDiv achieves superior performance over alternative diversity metrics in predicting ground-truth diversity across tasks and embedding models. Points with different shapes denote different metric correlation scores, with error bars indicating standard deviations across 5 repeated cross-validation trials. Experiments with \textit{ABS-HDS} exhibit larger error bars due to its smaller sample size.}
    \label{fig:text_emb}
\end{figure*}

To this end, we compare PLDiv against several established metrics, including Vendi Score, DCScore, and MAGAREA on the dataset \citep{turkes2022effectiveness}, by computing similarity scores from the data and using these scores as features to regress the curvature labels. We employ an SVR (support vector regression) model with an RBF kernel and perform 5-fold cross-validation. For Vendi Score and DCScore, we consider both L1 distance and RBF as similarity functions, whereas MAGAREA uses the default Euclidean distance. 
Table \ref{tab:curvature} indicates that the performance of other metrics, such as Vendi Score and DCScore, is highly dependent on the choice of similarity functions, and PLDiv is the strongest predictor for capturing data geometric structure. The Sparse PLDiv uses the sparse Rips filtration to reduce computation efforts (see Section 5.5).


\subsection{Semantic Diversity in Text Embeddings}
We investigate the utility of PLDiv as a measure of semantic diversity encoded in text embeddings. We use the dataset from  \cite{tevet-berant-2021-evaluating}, which contains 1,000 sets of 10 sentences generated from unique prompts across three distinct tasks: story completion (story), dialogue response generation (resp), and three-word prompt completion (prompt). 
For each prompt, 10 candidate outputs were generated by varying the softmax temperature $dec$, yielding a dataset of 1,000 prompts, each associated with 10 output sentences. Subsequently, human evaluators annotated a subset of 200 prompts, with 10 responses per prompt, to obtain the mean human evaluation score (\textit{ABS-HDS}), forming the human dataset. $Dec$ demonstrates the trade-off between quality and diversity in text generation, as lower temperatures increase fidelity by discouraging low-probability tokens, but at the cost of diversity in sampling. 
\textit{ABS-HDS} serves as the ground truth reflecting how humans perceive text diversity. Accordingly, we use linear regression with 5-fold cross-validation to analyze the relationship between response diversity measurements and temperature settings (as a proxy for diversity in the $dec$ dataset) or the human diversity scores (in the \textit{ABS-HDS} dataset), assessed using $R^2$ and MSE. In addition, we compute Pearson’s correlation and perform 1,000 bootstrap iterations to derive confidence intervals.
Each response set is embedded using five models: ``bert-large-nli-stsb-mean-tokens", ``all-MiniLM-L12-v2", and ``all-mpnet-base-v2", ``Qwen3-Embedding-4B", and ``Qwen3-Embedding-8B".

Fig. \ref{fig:text_emb} visualizes the $R^2$ and correlation results across all tasks and embedding models. PLDiv consistently outperforms all other metrics across tasks and embedding models in temperature-based evaluations. It also demonstrates superior performance in dialogue response generation across all models, as well as in evaluations on two recent embedding models (Qwen3-4B and Qwen3-8B) for all tasks assessed by human judgments. Moreover, PLDiv performs comparably to the Vendi Score in both story completion tasks and prompt tasks in human evaluations, while outperforming DCScore and MagArea. Detailed MSE results and performance analyses under different distance matrix settings are provided in Appendix \ref{appedix:text}. Overall, these results demonstrate that PLDiv effectively captures the semantic diversity encoded in text embeddings.

\begin{figure}[!h]
    \centering
        \includegraphics[width=0.23\textwidth]{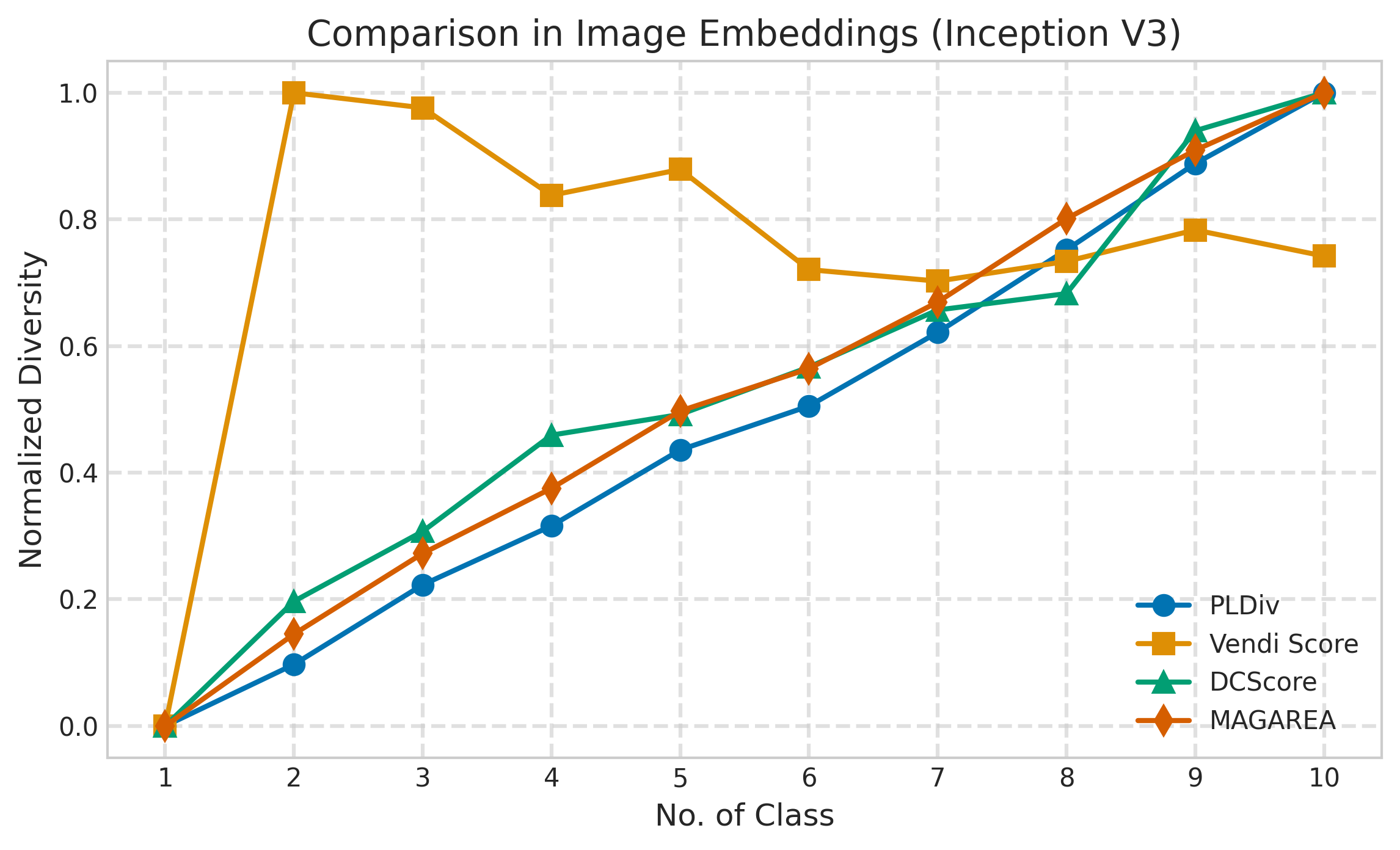}
        \includegraphics[width=0.23\textwidth]{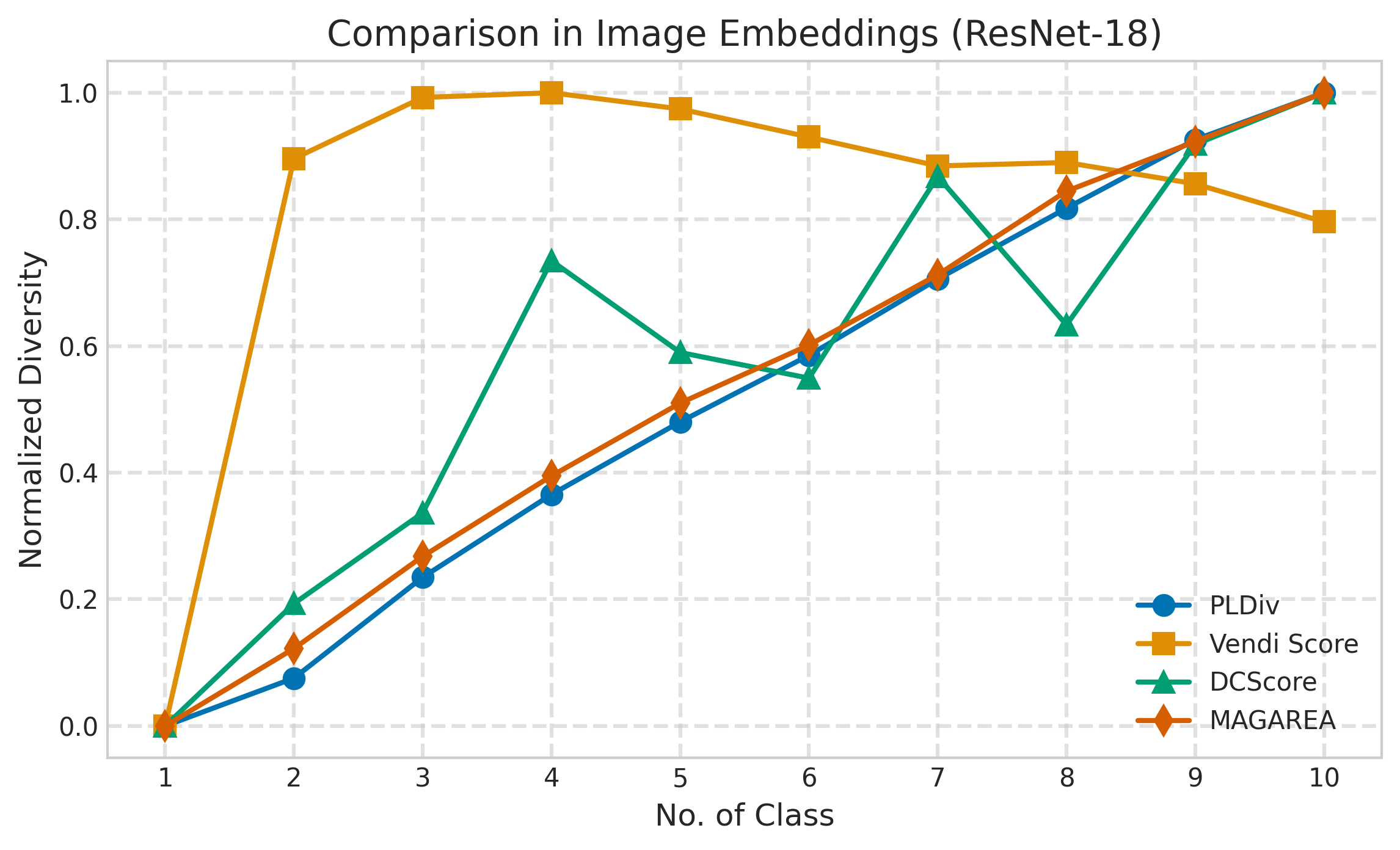}
 \caption{PLDiv shows a near-perfect correlation with the amount of the class involved in the dataset and remains consistent across different embedding models. MAGAREA performs next best, followed by DCScore, which exhibits some fluctuations in performance. Vendi Score, however, fails to capture the underlying patterns in the data.}
 \label{fig:image}
\end{figure}

\subsection{Diversity Evaluation for Image Embeddings}
To assess PLDiv's efficacy for image dataset evaluation, we tested it on Colored MNIST \citep{deng2012mnist}. Following the methodology of \citet{ospanov2024towards}, the number of labels served as the ground truth for diversity, where a higher label count signifies a more diverse set. Comparisons are conducted against Vendi Score, Magnitude, and DCScore, using two embedding models: Inception V3 and ResNet-18. Starting with a single class, we iteratively add one class at a time based on the previous data until all 10 classes are included. To facilitate a direct comparison, each metric was subsequently normalized to the [0, 1] interval (Min–max). This linear transformation preserves the underlying trends and the correlation of each score against the number of classes present in the evaluation.





In Fig. \ref{fig:image}, both PLDiv and MAGAREA exhibit a consistent and reliable correlation with the number of classes, aligning closely with the diagonal representing perfect correlation. PLDiv, however, offers faster computation and higher correlation. DCScore follows, showing comparable performance with one embedding model but greater variance with the other. In contrast, Vendi Score tends to decrease as the number of classes and the amount of data increase. This indicates that PLDiv's geometry-aware properties make it particularly well-suited for vision tasks, where embeddings often encode the geometric structure of images.

\subsection{Computation Complexity} 

\begin{table}[h]
\centering
\caption{Computation time (s) of diversity metrics from distance/similarity matrices on ImageNet-1K using ResNet-50 embeddings for sample sizes from 5k to 40k. Missing values for MAGAREA are due to its prohibitive runtime.}
\label{tab:time}
\setlength{\tabcolsep}{3pt}
\renewcommand{\arraystretch}{1.05}
\resizebox{\columnwidth}{!}{%
  \begin{tabular}{lccccc}
    \toprule
    \multirow{2}{*}{\textbf{Method}} & \multicolumn{5}{c}{\textbf{Sample size}} \\
    \cmidrule(lr){2-6}
    & 5k & 10k & 20k & 30k & 40k \\
    \midrule
    Vendi Score    & $1.60_{\pm0.83}$   & $10.82_{\pm2.73}$  & $183.80_{\pm12.88}$ & $746.51_{\pm30.74}$  & $1786.11_{\pm184.64}$ \\
    DCScore        & $0.03_{\pm0.02}$   & $0.13_{\pm0.01}$   & $0.46_{\pm0.01}$    & $1.00_{\pm0.01}$     & $1.80_{\pm0.05}$ \\
    MAGAREA        & $164.91_{\pm29.55}$& $716.14_{\pm31.23}$& --                  & --                   & -- \\
    \midrule
    \textbf{PLDiv} & $5.43_{\pm0.02}$   & $24.33_{\pm0.09}$  & $105.62_{\pm0.35}$  & $236.23_{\pm0.76}$   & $462.75_{\pm0.56}$ \\
    \textbf{\makecell[l]{Sparse PLDiv\\($\epsilon=0.95$)}}
                   & $3.97_{\pm0.03}$   & $16.80_{\pm0.37}$  & $68.55_{\pm2.21}$   & $147.48_{\pm6.50}$   & $273.86_{\pm14.35}$ \\
    \textbf{\makecell[l]{Sparse PLDiv\\($\epsilon=10$)}}
                   & $2.61_{\pm0.00}$   & $9.87_{\pm0.05}$   & $33.74_{\pm0.01}$   & $68.15_{\pm0.76}$    & $115.54_{\pm0.24}$ \\
    \bottomrule
  \end{tabular}%
}
\end{table}

In this section, we analyze the computational cost of our proposed metric compared with existing approaches. 
When the input is a point cloud $\mathcal{X} \in \mathbb{R}^{n \times d}$, computing all pairwise distances requires $\mathcal{O}(n^2 d)$ time, whereas utilizing a precomputed distance matrix sets the baseline at $\mathcal{O}(n^2)$. While standard persistent homology and PLDiv computation scale quadratically with $n$ due to the number of edges, their effective cost can be substantially reduced via sparsification. Specifically, the sparse Rips filtration \citep{cavanna2015geometric} 
utilizes a tolerance parameter $\epsilon$ to construct a $(1+\epsilon)$-approximation of the metric space. This method prunes the graph to a linear size $\mathcal{O}(C(\epsilon)n)$; since $C(\epsilon)$ scales inversely with $\epsilon$, larger tolerances significantly accelerate computation with negligible accuracy loss (see Table \ref{tab:sparse}). We then compute PLDiv using the Minimum Spanning Tree (MST) of the sparse Rips graph, a strategy that reduces the standard $\mathcal{O}(n^2)$ time and memory complexity of dense methods to near-linear time and linear $\mathcal{O}(n)$ memory \citep{cormen2022introduction}. Finally, PLDiv can be computed via a closed-form expression in $\mathcal{O}(N_d)$ time, outperforming the Vendi Score and MAGAREA on large-scale benchmarks (see Table \ref{tab:time}). 

Although the objective of PLDiv is to provide a precise and reliable data diversity measurement to supplement the existing diversity metrics, we demonstrate its practicality at scale. Given its superior performance in multiple domains, we consider the overall advantage of PLDiv quite evident compared to alternative metrics.

\begin{table}[h]
\centering
\caption{Sparse PLDiv values demonstrating reliable computation.}
\label{tab:sparse}
\setlength{\tabcolsep}{4pt}
\renewcommand{\arraystretch}{1.05}
\resizebox{0.7\columnwidth}{!}{%
\footnotesize	
\begin{tabular}{lccccc}
\toprule
\multirow{2}{*}{\textbf{Method}} & \multicolumn{5}{c}{\textbf{Sample size}} \\
\cmidrule(lr){2-6}
& 5k & 10k & 20k & 30k & 40k \\
\midrule
\textbf{PLDiv}
& $46.51$ & $78.01$ & $133.55$ & $184.93$ & $232.89$ \\
\textbf{\makecell[l]{Sp.\ PLDiv\\($\epsilon=0.95$)}}
& $46.52$ & $78.03$ & $133.58$ & $184.92$ & $232.89$ \\
\textbf{\makecell[l]{Sp.\ PLDiv\\($\epsilon=10$)}}
& $47.32$ & $79.70$ & $136.86$ & $190.23$ & $240.04$ \\
\bottomrule
\end{tabular}%
}
\end{table}

\section{Conclusion}

Understanding data diversity requires moving beyond traditional notions of variation or entropy to account for the intricate geometric and topological structures inherent in complex datasets. We propose a geometry-aware data diversity measure based on persistence landscapes, a tool from topological data analysis that provides a stable and expressive representation of hidden structural patterns. Our metric, PLDiv, offers a richer and more nuanced quantification of diversity. 
Through extensive experiments across multiple domains and modalities, we demonstrate  PLDiv's ability to characterize structural properties in data clouds (e.g., synthetic and curvature data) and in vector embeddings (e.g., text and image data). 
These results suggest that PLDiv provides a principled foundation for analyzing geometric diversity, with potential applications in dataset construction, augmentation, and model evaluation. Looking forward, integrating topological perspectives into automated dataset design and generative modeling could fundamentally reshape how diversity is understood, measured, and leveraged in artificial intelligence.

\section*{Impact Statement}
This paper presents work the goal of which is to advance the field of Machine Learning. There are many potential societal
consequences of our work, none which we feel must be specifically highlighted here.

\bibliography{refs}
\bibliographystyle{icml2026}

\newpage
\appendix
\onecolumn

\section{Additional Iterature Review}
\subsection{Diversity Measurement}


Evaluating diversity has long been a challenge in machine learning and generative modeling, partly because it is not always formalized under a single definition but manifests across different dimensions. For example, holistic evaluations of language models highlight variation in task coverage, domain shifts, linguistic and dialectal richness, input perturbations, and social context, all of which directly connect to the broader notion of data diversity \citep{liang2022holistic}.

Some works emphasize that inducing or controlling diversity can be as important as measuring it. Behavioral frameworks such as CheckList \citep{ribeiro2020beyond} systematically probe models through templating, lexical substitutions, and perturbations, showing that diverse inputs are essential for revealing hidden model failures, even though diversity itself is not explicitly quantified.

Diversity is not always treated only as an evaluation objective, but also as a design principle at the training level. For instance, Du and Black \citep{du2019boosting} mitigate mode collapse in dialogue generation by iteratively boosting models to promote semantic and lexical variation. Although effective in practice, these approaches underscore the need for principled evaluation frameworks that can verify whether training-time interventions truly enhance diversity across settings.

To address semantic variation more directly, semantic diversity methods examine conceptual distinctions between outputs. Stasaski and Hearst \citep{stasaski2022semantic} use Natural Language Inference models to identify entailment, contradiction, and neutrality among generated texts, treating contradiction as a marker of diversity and entailment as redundancy. Although intuitive and fine-grained, this relational approach is inherently limited to pairwise comparisons and does not capture global structural diversity across datasets.

A large class of methods focuses on surface-level variation, particularly in text. N-gram–based metrics such as distinct-n \citep{song2024scaling}, self-BLEU \citep{shu2019generating}, and ROUGE-L \citep{wang2022self,padmakumar2023does} capture token-level dispersion across samples \citep{yu2017seqgan}. Similarly, the Data Quality Index (DQI) \citep{mishra2020dqi} aggregates vocabulary richness, entropy, and syntactic variation to assess dataset quality. While easy to compute, these approaches provide only a narrow view of diversity, often missing deeper semantic or structural patterns.

\subsection{Persistent Homology in Metric Space}

The formal algebraic foundations were established by \cite{zomorodian2004computing}, who introduced persistence modules, provided algorithms for computing persistence, and proved the barcode decomposition theorem as a complete invariant over fields. This work grounded PH in computability and algebraic classification, laying the basis for its adoption across domains \citep{zhao2019learning,hiraoka2016hierarchical,pun2022persistent}. However, these foundational contributions primarily emphasize topology extraction and stability, without directly connecting persistence to data-level diversity or representational richness.

Beyond its theoretical foundations, TDA and persistent homology have shown practical utility across diverse domains. In neuroscience, PH captures vascular structures linked to disease \citep{bendich2016persistent}; in materials science, it characterizes microstructures and force chains in amorphous solids \citep{hiraoka2016hierarchical}; and in biology and chemistry, it reveals topological signatures of protein folding, molecular stability, and binding sites \citep{xia2015multidimensional,kovacev2016using,gameiro2015topological}. These examples highlight PH’s ability to extract robust, multi-scale features from high-dimensional and noisy data.

PH has also been applied to both temporal and spatial systems. Persistence landscapes have been used to track transitions in dynamical systems and classify time-series data \citep{gidea2018topological,umeda2017time}, while in astrophysics, PH captures the multiscale filamentary structure of the cosmic web from cosmological simulations \citep{aragon2010multiscale}. Collectively, these applications highlight PH’s versatility as a modality-agnostic framework for extracting global, nonlinear structure that often remains inaccessible to conventional statistical or machine learning methods.

\section{Description of Diversity Scores in Comparisons}



Vendi Score (VS) \citep{dan2023vendi}, derived from a set of samples and their pairwise similarity functions, quantifies the similarities among the data in a dataset. 
Mathematically, VS is given by the exponential of the Shannon entropy, which is obtained from the eigenvalues of the scaled similarity matrix \( X^\top X \):
\[
VS 
= \exp \left( - \sum_{i=1}^{n} \lambda_i \log \lambda_i \right)
\]
where \( \lambda_i \) are the eigenvalues of scaled \( X^\top X \).


\citet{limbeck2024metric} introduces several \emph{magnitude-based} diversity
measures that leverage the notion of the effective size of a metric space across scales. The core idea
is to compute the \emph{magnitude function}, $\text{Mag}_X(t)$, which tracks how the effective
number of points in a space changes as pairwise distances are rescaled. To summarise this behaviour,
the authors propose two derived metrics: the area under the magnitude function (MAGAREA) as a
reference-free measure of intrinsic diversity, and the difference between magnitude functions
(MAGDIFF) as a reference-based measure:
\[
\text{MAGAREA} = \int_{t_0}^{t_{\text{cut}}} \text{Mag}_X(t)\, dt, 
\quad
\text{MAGDIFF} = \int_{t_0}^{t_{\text{cut}}} \big(\text{Mag}_X(t) - \text{Mag}_Y(t)\big)\, dt,
\]
where $\text{Mag}_X(t)$ is the magnitude function of $X$ at scale $t$ and $t_{\text{cut}}$ denotes
the convergence scale used for evaluation. These measures provide robust multi-scale summaries of diversity and have been shown to detect phenomena such as curvature, mode collapse, and mode dropping in text, image, and graph representations.

\citet{zhu2025measuring} proposes \textbf{DCScore}, which departs from entropy or scale-based approaches by reframing diversity measurement as a \textit{classification problem}. Instead of relying on eigenvalue decomposition or scale-sensitive geometric measures, DCScore evaluates how well each individual sample in a dataset can be distinguished from all others. Specifically, each sample is treated as its own class, and pairwise similarities are converted into classification probabilities through a softmax function. The last score is then defined as the trace of the resulting probability matrix:
\[
\text{DCScore}(D) = \mathrm{tr}(P) = \sum_{i=1}^n P[i,i], 
\quad P[i,j] = \frac{\exp\left(\tfrac{K[i,j]}{\tau}\right)}{\sum_{k=1}^n \exp\left(\tfrac{K[i,k]}{\tau}\right)},
\]
where \(K[i,j]\) denotes the similarity between samples \(i\) and \(j\), and \(\tau\) is a temperature parameter that controls the classification sharpness. This formulation is principled and efficient, emphasizing sample separability without considering the geometric or topological structure of the dataset, which can also be important for characterizing diversity.

\section{Mathematical Proofs}
\label{appex:b}

\subsection{PLDiv Closed Form }
Let $\mathcal D=\{(b_i,d_i)\}_{i=1}^m$ be a finite multiset of persistence birth--death pairs and let $\lambda_{(b_i,d_i)}:\mathbb R\to[0,\infty)$ denote the usual persistence ``tent'' function associated to the interval $(b_i,d_i)$. Let $\{\lambda_k(t)\}_{k\ge1}$ be the persistence landscape functions obtained by ordering the values $\{\lambda_{(b_i,d_i)}(t)\}_{i=1}^m$ at each fixed $t$ in nonincreasing order (with $\lambda_k(t)=0$ for all $k>m$). Then

\[
\mathrm{PLDiv}(\mathcal{X})
= \sum_{k=1}^\infty \int_{\mathbb{R}} \lambda_k(t)\,dt
= \sum_{i=1}^m \int_{\mathbb{R}} \lambda_{(b_i,d_i)}(t)\,dt
= \frac{1}{4}\sum_{i=1}^m (d_i-b_i)^2.
\]

\begin{proof}
By definition $\lambda_k(t)$ are the order statistics (at each fixed $t$) of the family $\{\lambda_{(b_i,d_i)}(t)\}_{i=1}^m$. For any finite collection of nonnegative functions $f_i(t)$,
\[
\sum_{k=1}^\infty \text{$k$-th largest of $\{f_i(t)\}$} \;=\; \sum_{i=1}^m f_i(t),
\]

Applying this pointwise gives
\[
\sum_{k=1}^\infty \lambda_k(t) = \sum_{i=1}^m \lambda_{(b_i,d_i)}(t).
\]

Each $\lambda_{(b_i,d_i)}$ is continuous with compact support $[b_i,d_i]$, hence measurable and integrable. By Tonelli's theorem \citep{tao2011introduction},
\[
\sum_{k=1}^\infty \int_{\mathbb R} \lambda_k(t)\, dt
= \int_{\mathbb R} \sum_{k=1}^\infty \lambda_k(t)\, dt
= \int_{\mathbb R} \sum_{i=1}^m \lambda_{(b_i,d_i)}(t)\, dt
= \sum_{i=1}^m \int_{\mathbb R} \lambda_{(b_i,d_i)}(t)\, dt.
\]

Finally, each tent function supported on the interval $[b_i,d_i]$ is a symmetric isosceles triangle of base length $d_i-b_i$ and height $(d_i-b_i)/2$, hence its area is
\[
\int_{\mathbb{R}} \lambda_{(b_i,d_i)}(t)\,dt \;=\; \tfrac12\cdot (d_i-b_i)\cdot \tfrac{d_i-b_i}{2} \;=\; \frac{(d_i-b_i)^2}{4},
\]
Summing over $i=1,\dots,m$ gives the final identity
\[
\sum_{i=1}^{m} \int_{\mathbb{R}} \lambda_{(b_i,d_i)}(t)\, dt
= \frac{1}{4} \sum_{i=1}^{m} (d_i - b_i)^2.
\]

\end{proof}

\subsection{Axiomatic Properties of Diversity}

A diversity measure derived from Persistence Landscapes (PLs) is defined as a summary statistic of the persistence lifetimes generated from a dataset's Vietoris-Rips filtration. We prove that such a measure satisfies the key principles of effective size, monotonicity, the twin property, and symmetry.

\begin{itemize}
    \item \textbf{Effective size.} 
    For a fixed number of points, $\mathrm{PLDiv}(\mathcal{X})$ increases when data points are well-separated and decreases as they cluster, reaching a maximum when all points are distinct and a minimum when all are identical.

 \begin{proof}   
 
\textit{Minimum $\mathrm{PLDiv}$:}
The minimum value of $\mathrm{PLDiv}$ is achieved when all points in the cloud $\mathcal{X}$ are identical. Let all $n$ points in the cloud be the same, so
$x_{1} = x_{2} = \dots = x_{n}.$
The distance between any two points is zero:
\[
d(x_{i}, x_{j}) = 0 \quad \text{for all } i,j.
\]

Every point is born at $\varepsilon=0$ and immediately merges with every other point at $\varepsilon=0$, all persistence lifetimes are zero. That is,
\[
b_{i} = 0, \quad d_{i} = 0 \quad \text{for all features}.
\]

Therefore, 
\[
\min \mathrm{PLDiv}(\mathcal{X}) = \frac{1}{4}\sum_{i} (d_{i} - b_{i})^{2} = \frac{1}{4}\sum_{i} (0 - 0)^{2} = 0.
\]


\textit{Maximum $\mathrm{PLDiv}$:} The maximum value of $\mathrm{PLDiv}$ is achieved when the points are ``well-separated.''
Let $\mathcal{X} = \{x_1, \dots, x_n\}$ be a point cloud in a metric space $(\mathcal{M}, d)$ such that all points are distinct and equidistant:
\[
d(x_i, x_j) = c > 0 \quad \text{for all } i \neq j.
\]
Then, the $H_0$ persistence lifetimes are all equal to $c$, except for the last surviving component. 
Let $c = \max_{i \neq j} d(x_i, x_j)$. In the Vietoris--Rips filtration, at $\varepsilon = 0$, each point forms a separate connected component. Thus, there are $n$ components born at $b_i = 0$. For $0 < \varepsilon < c$, no edges appear because all pairwise distances are $c$. Hence, no components merge in this interval. At $\varepsilon = c$, all pairwise edges appear simultaneously, and the $n$ components merge into a single connected component. Thus, $n-1$ components die at $d_i = c$, while the last component persists indefinitely. 



By Proposition 3.2, the corresponding $\mathrm{PLDiv}$ is
\[
\max \mathrm{PLDiv}(\mathcal{X}) = \frac{n-1}{4}\,c^2.
\]
\end{proof}

\item \textbf{Monotonicity}


    Fix $n$ and let $\mathcal{X}$ be a point cloud in a metric space. If all pairwise distances in $\mathcal{X}$ are scaled by a factor $\alpha>1$ (i.e. replace the metric $d(\cdot,\cdot)$ by $\alpha d(\cdot,\cdot)$), then
\[
\mathrm{PLDiv}(\alpha \mathcal{X}) \;\begin{cases}
\le \alpha^2 \,\mathrm{PLDiv}(\mathcal{X}), & \alpha > 1, \\[6pt]
\ge \alpha^2 \,\mathrm{PLDiv}(\mathcal{X}), & 0 < \alpha < 1.
\end{cases}
\]

 \begin{proof}   
    Fix $n$ and let $\mathcal{X}$ be a point cloud in a metric space. If all pairwise distances in $\mathcal{X}$ are scaled by a factor $\alpha>1$ (i.e. replace the metric $d(\cdot,\cdot)$ by $\alpha d(\cdot,\cdot)$), then every lifetime $d_i-b_i$ is multiplied by $\alpha$. By Proposition 3.2,
\[
\mathrm{PLDiv}(\alpha\mathcal{X})=\frac14\sum_i(\alpha(d_i-b_i))^2=\alpha^2\cdot\frac14\sum_i(d_i-b_i)^2=\alpha^2\mathrm{PLDiv}(\mathcal{X}).
\]
Hence, spreading the same set of points apart (uniform dilation) strictly increases PLDiv (for $\alpha>1$). More generally, moving points so as to increase lifetimes of the dominant features increases PLDiv; conversely, clustering points tends to shorten lifetimes and reduce PLDiv.
\end{proof}

\item \textbf{Twin property.}
Adding an exact duplicate of a point does not change $\mathrm{PLDiv}(\mathcal{X})$. 
    Let $\mathcal{X}$ be a dataset and let $x_i \in \mathcal{X}$. For the set $\mathcal{X}' = \mathcal{X} \cup \{x_n\}$ where $x_n = x_i$, the diversity is unchanged:
    \[
    \mathrm{PLDiv}(\mathcal{X}') = \mathrm{PLDiv}(\mathcal{X}).
    \]

\begin{proof}
A duplicate point at exactly the same coordinates is at zero distance from its twin. In the usual filtrations built from pairwise distances (e.g., Vietoris--Rips), the duplicate component is born at radius $0$ and immediately merges with its twin also at radius $0$. Hence the corresponding birth--death pair is $(0,0)$ and has lifetime $0$, contributing $(d-b)^2/4=0$ to the PLDiv sum. All other birth--death pairs are unchanged as well. Therefore $\mathrm{PLDiv}$ is unchanged.
\end{proof}


\item \textbf{Symmetry.}
PLDiv is invariant to the ordering of data points (permutation invariance). 
    Since persistent homology depends only on the metric structure of $\mathcal{X}$ and $\mathrm{PLDiv}(\mathcal{X})$ is computed from the multiset of intervals $\{(b_i,d_i)\}$, relabeling or reordering points does not affect the value of the score. 
    Let $\mathcal{X}=(x_1,\dots,x_n)$ be an ordered sequence of points and let $\pi$ be any permutation of $\{1,\dots,n\}$. For the permuted sequence $\mathcal{X}_\pi=(x_{\pi(1)},\dots,x_{\pi(n)})$, we have
    \[
    \mathrm{PLDiv}(\mathcal{X}_\pi) = \mathrm{PLDiv}(\mathcal{X}).
    \]

\begin{proof}
    The PH pipeline begins with the pairwise distance matrix $D$, where $D_{ij}=d(x_i,x_j)$. Let $\mathcal{X}_\pi$ be the reordered dataset. The distance matrix $D_\pi$ for the permuted data has entries $(D_\pi)_{ij}=d(x_{\pi(i)},x_{\pi(j)})$. 
    Importantly, the set of all unique pairwise distances
    \[
    \{d(x_i,x_j)\}_{1 \leq i < j \leq n}
    \]
    is unchanged for both $\mathcal{X}$ and $\mathcal{X}_\pi$. The construction of the Vietoris--Rips filtration depends only on these distances. Hence, the persistence diagrams and lifetimes $\{l_i\}$ are identical. Therefore, any diversity measure computed from these lifetimes is invariant under permutation of the data and $\mathrm{PLDiv}$ is symmetry.

\end{proof}

\end{itemize}

\section{Detailed Experiment Descriptions}
\subsection{Synthetic Toy Examples}
\label{appendix:toy}

Our toy example in Figure~\ref{fig:pipeline} utilizes the examples from \citet{limbeck2024metric}. Specifically, we simulated four synthetic datasets with varying diversity levels. D1 (Poisson Process): 200 points uniformly sampled in the square $[0,2]^2$, representing a spatially random distribution. D2 (Hawkes Process): a clustered dataset generated via a self-exciting point process with base intensity $\lambda=91$ and excitation parameter $\alpha=0.6$. D3 (Two Gaussians): 200 samples drawn from two Gaussian clusters centered at $(0.5,\,0.5)$ and $(1.5,\,1.5)$ with covariance $0.02I$. D4 (One Gaussian): 200 samples drawn from a single Gaussian centered at $(0.5,\,0.5)$ with the same covariance. These datasets progressively transition from highly diverse and dispersed (D1) to concentrated and homogeneous (D4). Table \ref{tab:toy} represents diversity scores calculated by four metrics. (Vendi Score and DCScore are based on RBF kernel)

\begin{table}[ht]
\centering
\caption{Performance comparison of subset selection}
\label{tab:toy}
\begin{adjustbox}{max width=\textwidth}
\begin{tabular}{c|cccc}
\toprule
\textbf{Task} & \textbf{PLDiv ($\uparrow$)} & \textbf{Vendi Score (rbf) ($\uparrow$)} & \textbf{DCScore ($\uparrow$)} & \textbf{MagArea ($\uparrow$)} \\
\midrule
D1 &  0.53 & 136.98 & 2.67 & 141.23 \\
D2 & 0.49 & 79.96 & 2.63 & 108.83 \\
D3 & 0.26 & 40.40 & 2.48 & 81.93 \\
D4 & 0.05 & 23.66 & 2.32 & 58.53 \\
\bottomrule
\end{tabular}
\end{adjustbox}
\end{table}

\subsection{Imbalanced Long-tail data}
\label{appendix:tail}
\begin{figure}[h!]
    \centering
    \includegraphics[width=1.0\linewidth]{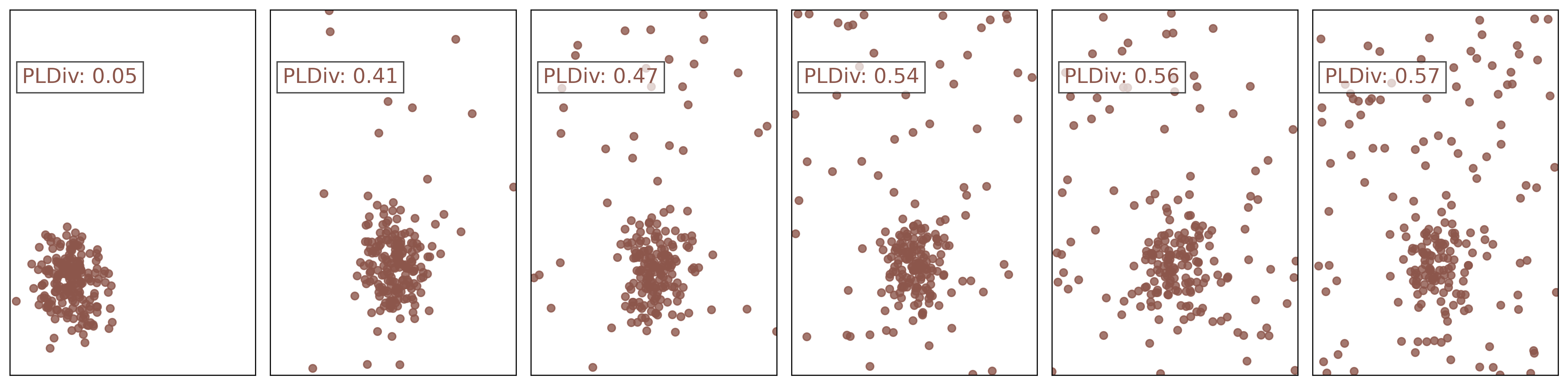}
    \caption{PLDiv can reliably predict diversity in imbalanced data, where diversity increases monotonically.}
    \label{fig:tail}
\end{figure}

To explore how PLDiv performs on imbalanced data, we generated a series of small long-tail datasets. First, we utilized D4 in synthetic toy examples, which form a single cluster with 200 data points. To simulate long-tail effects, outlier points were added uniformly within a square region in varying amounts of 20, 40, 60, 80, and 100 samples, while keeping the cluster size at 200 - $n_{\text{outliers}}$. Each variant thus exhibits increasing imbalance between the dense Gaussian core and sparse tail regions. Figure \ref{fig:tail} demonstrates that PLDiv effectively handles the imbalanced dataset.

\subsection{Implementation of Synthetic Data Clouds}
\label{app:datasets}

We created eight pairs of synthetic scenarios, each containing about 200 points generated from parameterized geometric functions. Each pair modifies one specific geometric property by adding or removing loops, bridges, curvature, or hierarchical clustering, while maintaining a comparable overall spatial scale. Table~\ref{tab:8datasets} summarizes the data generation process for the eight synthetic point-cloud pairs used in Sec.~\ref{sec:synthetic}. Each cloud contains approximately 200 points produced by explicit geometric or probabilistic functions (e.g., rings, Gaussian mixtures, sinusoidal manifolds). These datasets complement Table~\ref{tab:div_10cases}, which reports diversity metric values across the same scenarios.


\begin{table}[ht]
\centering
\caption{Synthetic dataset pairs used for geometry-aware diversity evaluation. Each cloud contains 200 points.}
\label{tab:8datasets}
\begin{tabular}{lll}
\hline
\textbf{Pair}  & \textbf{A (less varied geometry)}  & \textbf{B (more varied geometry)}\\
\hline
Disk vs Ring & Points on noisy circular rim (loop) & Uniform points in filled disk \\
Bridge vs Two Clusters & Same blobs plus short bridge (connectivity) & Two separated Gaussian blobs \\
Nested vs Gaussian & Inner Gaussian + outer ring (hierarchy) & Single Gaussian \\
Crescent vs Blob & Half-ring manifold (curvature) & Isotropic Gaussian cloud \\
Two Rings vs Random Cloud & Two concentric noisy rings (multi-loop) & Uniform on square [0, 2]\textsuperscript{2} \\
Snake vs Blob & Sinusoidal curve with noise (manifold) & Isotropic Gaussian \\
Hole vs Filled & Outer Gaussian with inner void (cavity) & Outer Gaussian + center points \\
Hierarchical vs Gaussian & Multi-level small clusters (multi-scale) & Single broad Gaussian \\
\hline
\end{tabular}
\end{table}

\begin{table}[ht]
\centering
\caption{Comparison of diversity metrics across synthetic dataset pairs.}
\begin{tabular}{llcccc}
\hline
\textbf{Scenario} & \textbf{Data} & \textbf{PLDiv} & \textbf{Vendi Score} & \textbf{DCScore} & \textbf{MagArea} \\
\hline
Ring vs Disk & A & 0.064 & 8.702 & 2.437 & 125.732 \\
Ring vs Disk & B & 0.262 & 8.746 & 1.957 & 140.620 \\
Two clusters vs Bridge & A & 0.134 & 4.915 & 1.578 & 143.599 \\
Two clusters vs Bridge & B & 0.150 & 5.132 & 1.585 & 153.364 \\
Nested vs Gaussian & A & 0.123 & 7.696 & 1.906 & 141.750 \\
Nested vs Gaussian & B & 0.623 & 9.641 & 1.878 & 142.509 \\
Crescent vs Blob & A & 0.030 & 5.025 & 1.919 & 127.702 \\
Crescent vs Blob & B & 0.147 & 4.469 & 1.450 & 136.976 \\
Two rings vs Random cloud & A & 0.176 & 11.569 & 2.257 & 132.447 \\
Two rings vs Random cloud & B & 0.551 & 15.436 & 2.583 & 140.364 \\
Snake vs Blob & A & 0.027 & 4.405 & 1.827 & 141.067 \\
Snake vs Blob & B & 0.156 & 4.589 & 1.455 & 142.696 \\
Hole vs Filled & A & 0.096 & 3.458 & 1.342 & 128.140 \\
Hole vs Filled & B & 0.101 & 3.559 & 1.352 & 122.926 \\
Hierarchical vs Gaussian & A & 0.222 & 4.048 & 1.824 & 63.258 \\
Hierarchical vs Gaussian & B & 0.420 & 7.972 & 1.768 & 139.101 \\
\hline
\end{tabular}
\label{tab:div_10cases}
\end{table}

\subsection{Implementation of Curvature Experiment}
\label{appendix: curvature}

In Section 5.2, we evaluate PLDiv along with alternative diversity metrics on the curvature dataset \citep{turkes2022effectiveness}. 
The dataset consists of two-dimensional point clouds sampled from smooth surfaces with varying degrees of curvature. Each sample represents a set of points $\{x_i\}_{i=1}^n \subset \mathbb{R}^d$ labeled by the curvature of the underlying manifold, either as discrete curvature classes or continuous curvature values, ranging from -2 to 2. The task is to predict this curvature from the sampled points, assessing how well diversity measures capture geometric information such as local bending and shape variation. This setup allows controlled evaluation of geometric sensitivity, robustness to noise, and invariance under isometric transformations. 

\begin{figure*}[h!]
    \centering
    \includegraphics[width=0.32\textwidth]{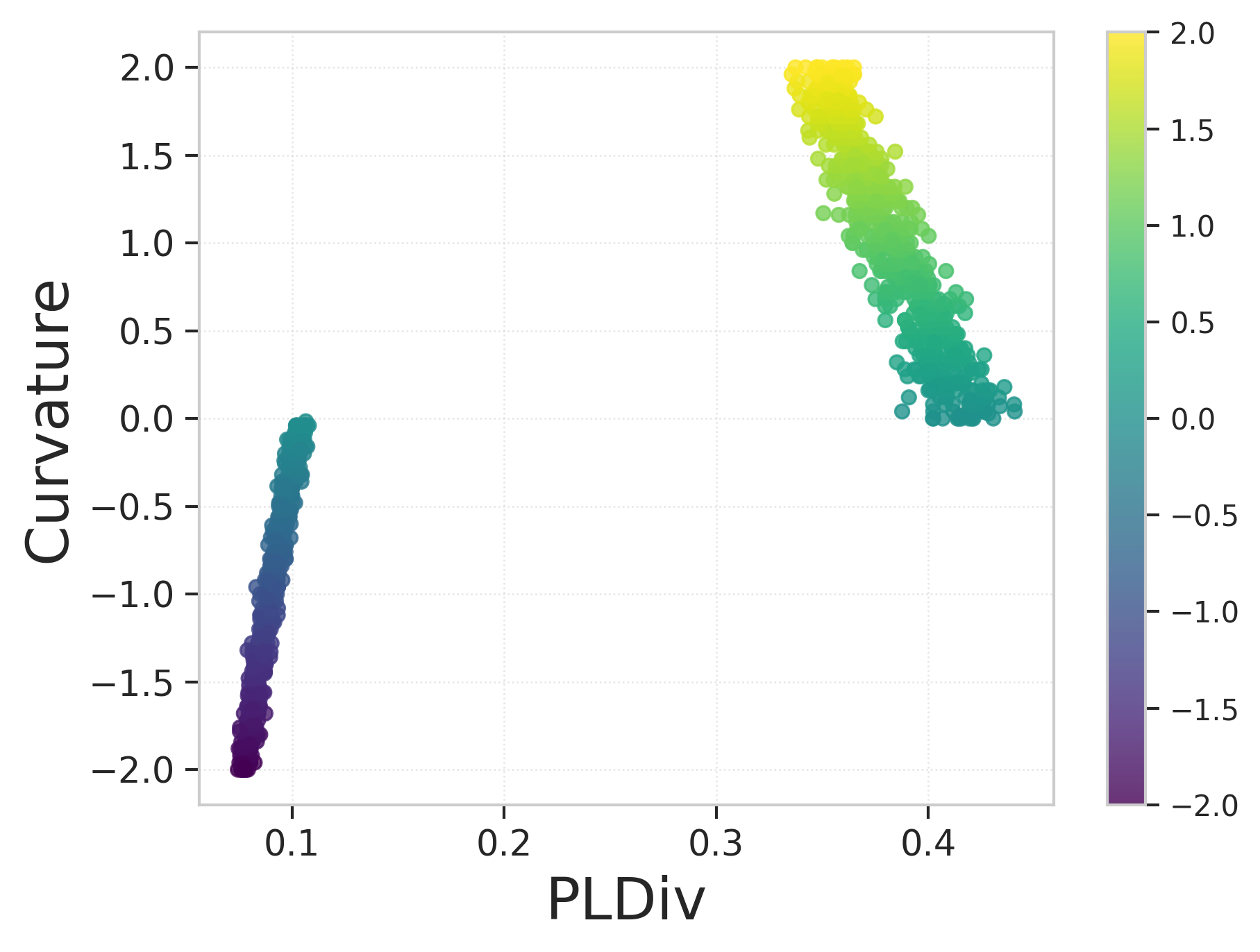}
    \includegraphics[width=0.32\textwidth]{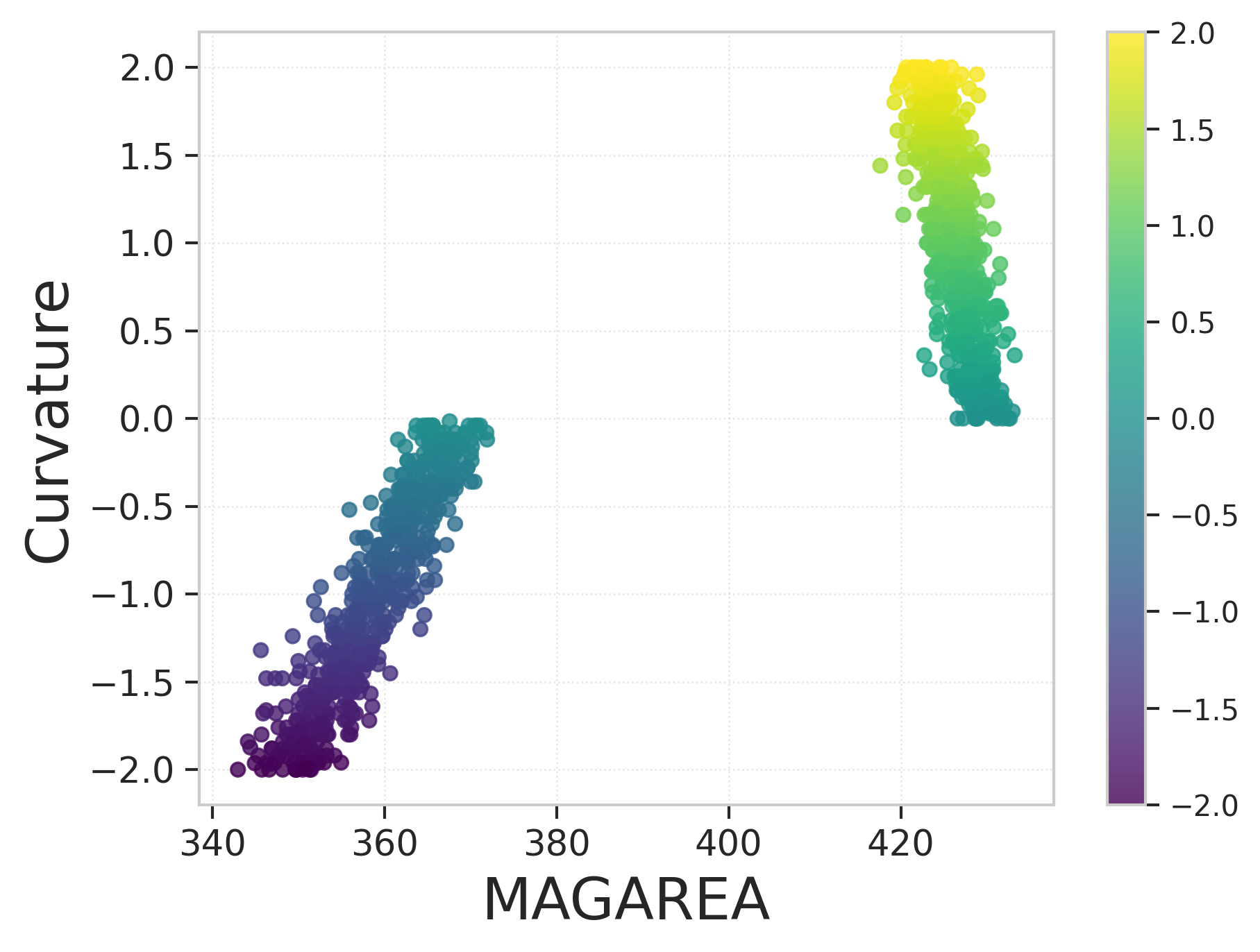}
    \includegraphics[width=0.32\textwidth]{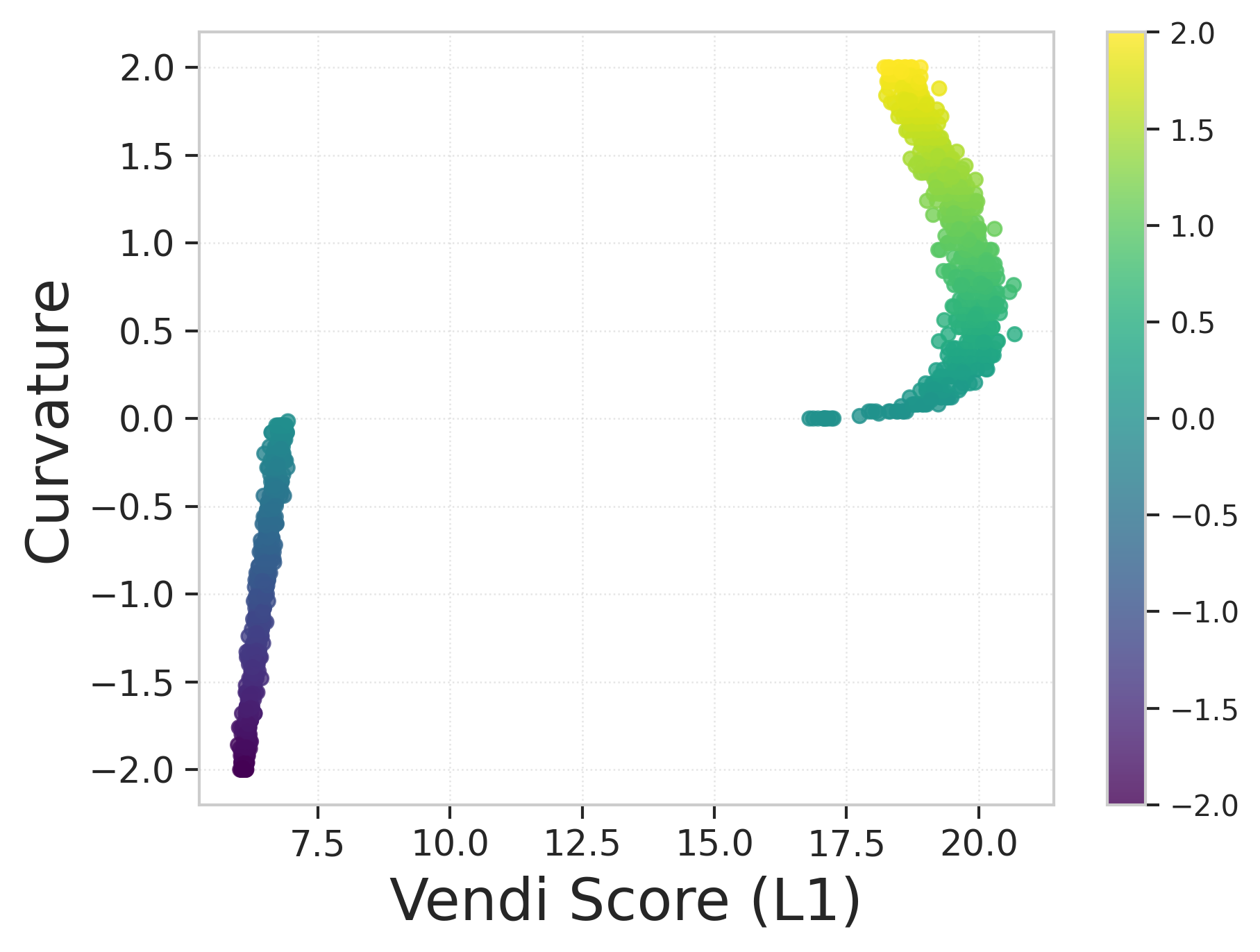}
    
    \vspace{2mm} 
    
    \includegraphics[width=0.32\textwidth]{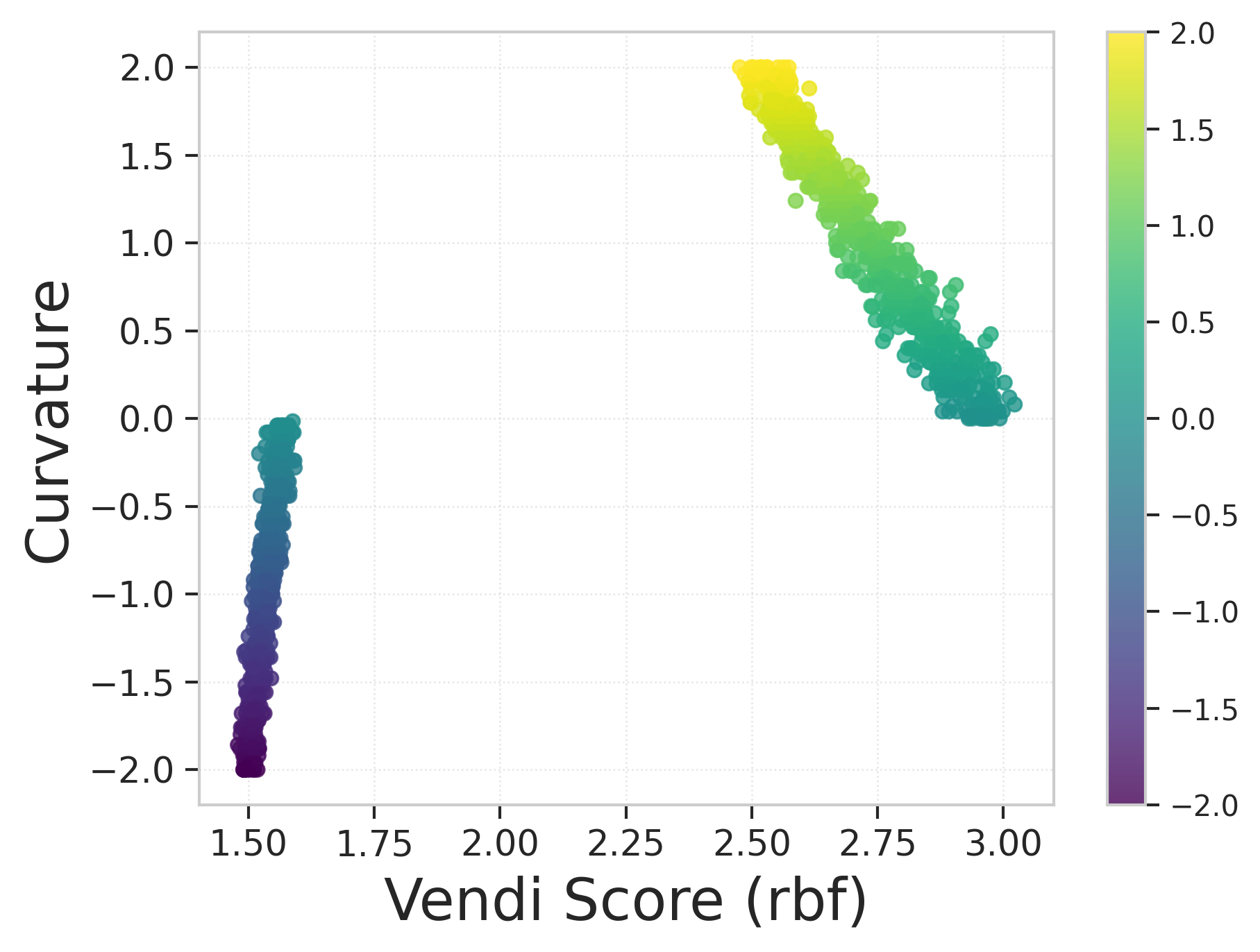}
    \includegraphics[width=0.32\textwidth]{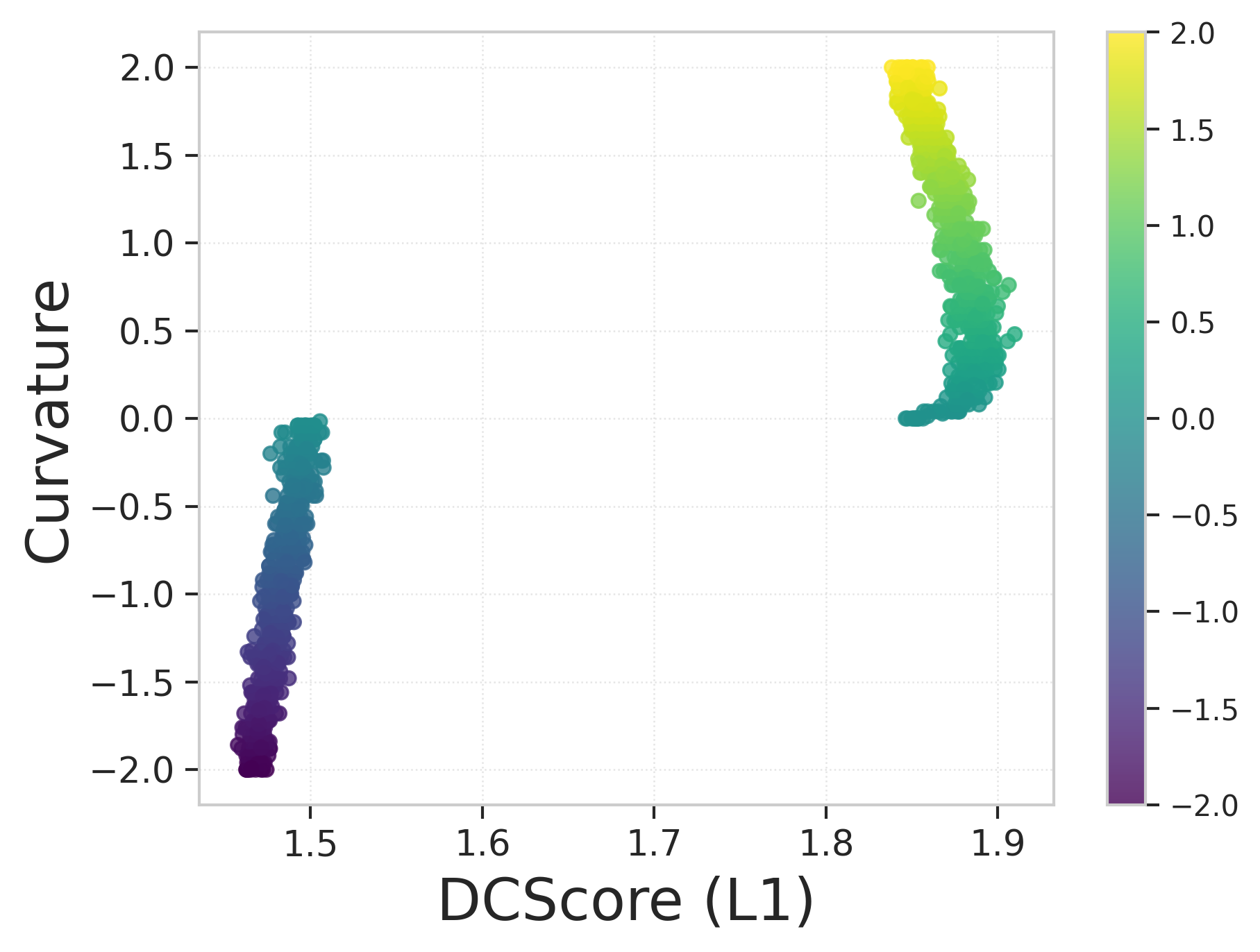}
    \includegraphics[width=0.32\textwidth]{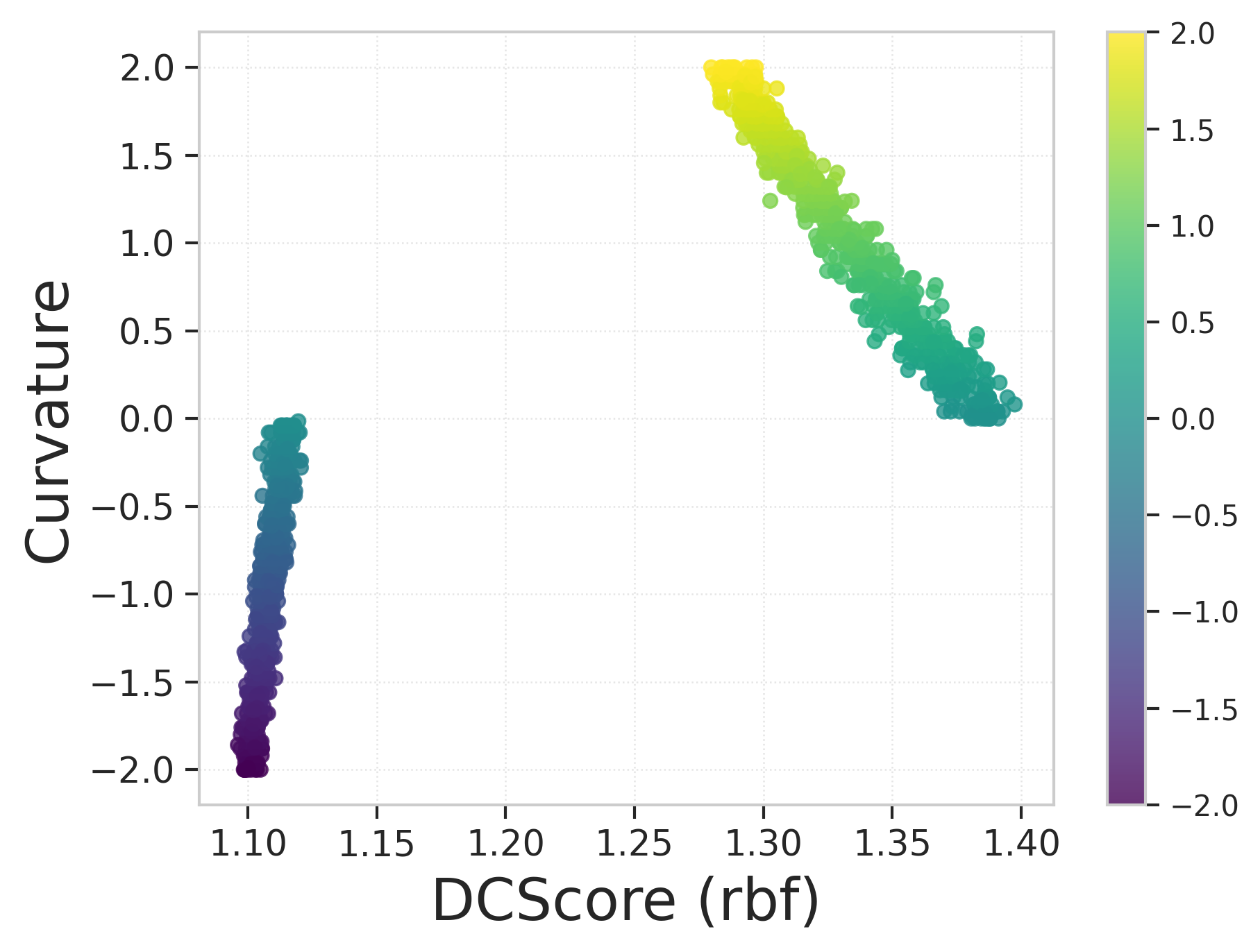}

    \caption{Visualizations of the diversity measures against the curvature labels show that PLDiv achieves the best separation between positive and negative curvatures, providing clear evidence of why it performs best in Section 5.2.}
    \label{fig:cur_all}
\end{figure*}

We employ a Support Vector Regression (SVR) model with a radial basis function (RBF) kernel, using the parameters C = 1.0 and $\epsilon = 0.1$. This configuration is applied to all metrics (PLDiv, Vendi Score, DCScore, and MagArea). MagArea uses Euclidean distance, while Vendi Score and DCScore are evaluated with both RBF and Laplacian kernels. In contrast, PLDiv takes the curvature data cloud as input and internally computes pairwise Euclidean distances. Table \ref{tab:curvature} and Figure \ref{fig:cur_all} demonstrate that PLDiv exhibits a truly geometry-aware property.

\subsection{Implementation of Text Embeddings}
\label{appedix:text}
We evaluate PLDiv as a metric of semantic diversity using the dataset from \citet{tevet-berant-2021-evaluating}, comprising 1,000 prompts from three tasks. Ten outputs per prompt were generated by varying the softmax temperature ($dec$), and a subset of 200 prompts was human-annotated to obtain mean diversity scores (\textit{ABS-HDS}). Text embedding models we used are listed below:

\begin{itemize} 
    \item all-MiniLM-L12-v2: general text embedding model, dimension 384
    \item all-mpnet-base-v2: general text embedding model, dimension 768 
    \item bert-large-nli-stsb-mean-tokens:  general text embedding model, dimension 1024
    \item Qwen3-Embedding-4B: advanced LLM-based embedding models, dimension 2560
    \item Qwen3-Embedding-8B: advanced LLM-based embedding models, dimension 4096 
\end{itemize}

\begin{figure}[htbp]
    \centering
    \includegraphics[width=1\linewidth]{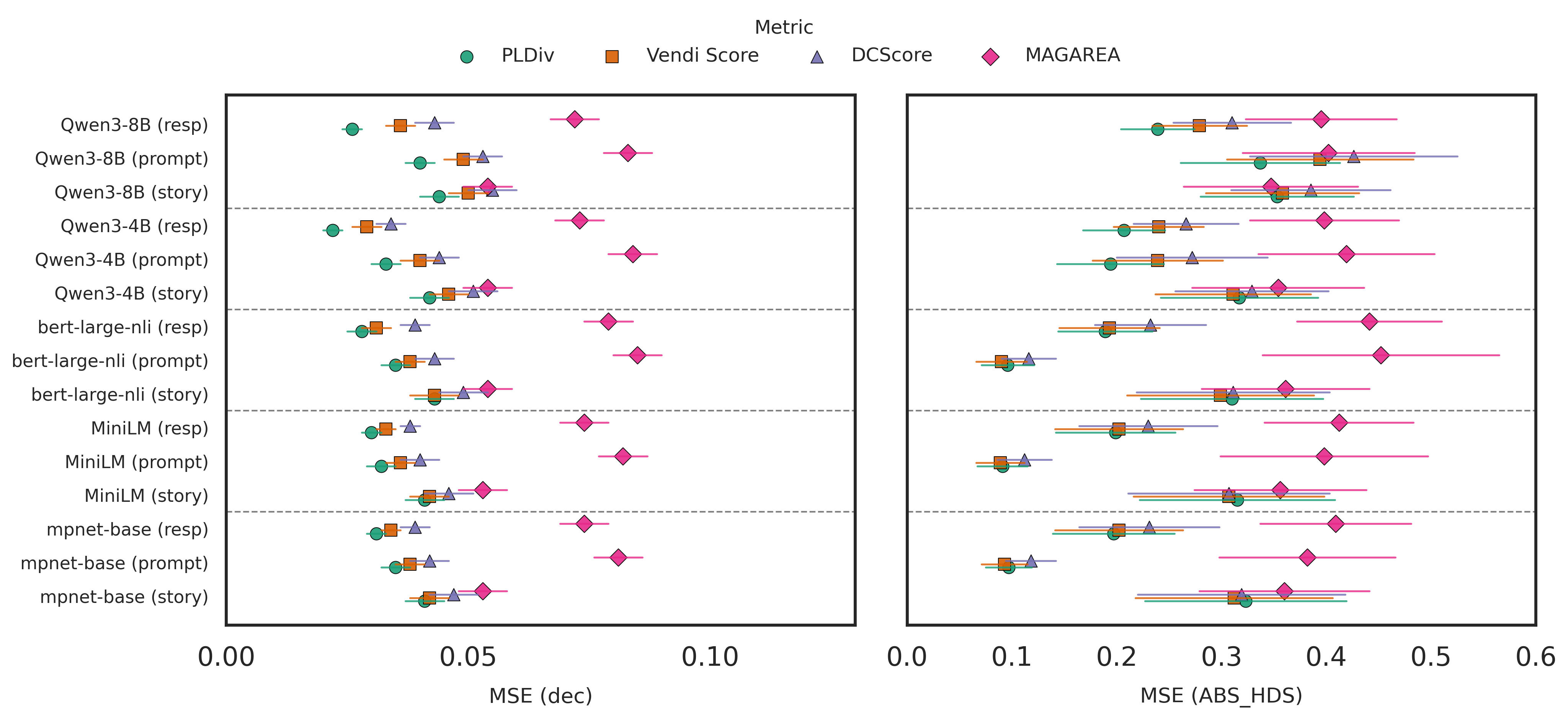}
    \caption{MSE for four metrics on both temperature $dec$ and human diversity score \textit{ABD-HDS}. PLDiv achieves the lowest MSE in the temperature ($dec$) tasks across all embedding models and remains among the lowest when evaluated on human-annotated scores.}
    \label{fig:text_mse}
\end{figure}

Figure \ref{fig:text_mse} represents Mean Squared Error (MSE) for linear regression that indicates the predictive capability for diversity metrics on softmax temperature $dec$ and mean human annotated diversity score (\textit{ABD-HDS}). PLDiv achieves the lowest MSE in the temperature ($dec$) tasks across all embedding models and remains among the lowest when evaluated on human-annotated scores. 

\begin{figure*}[t]
    \centering
    \includegraphics[width=0.65\textwidth]{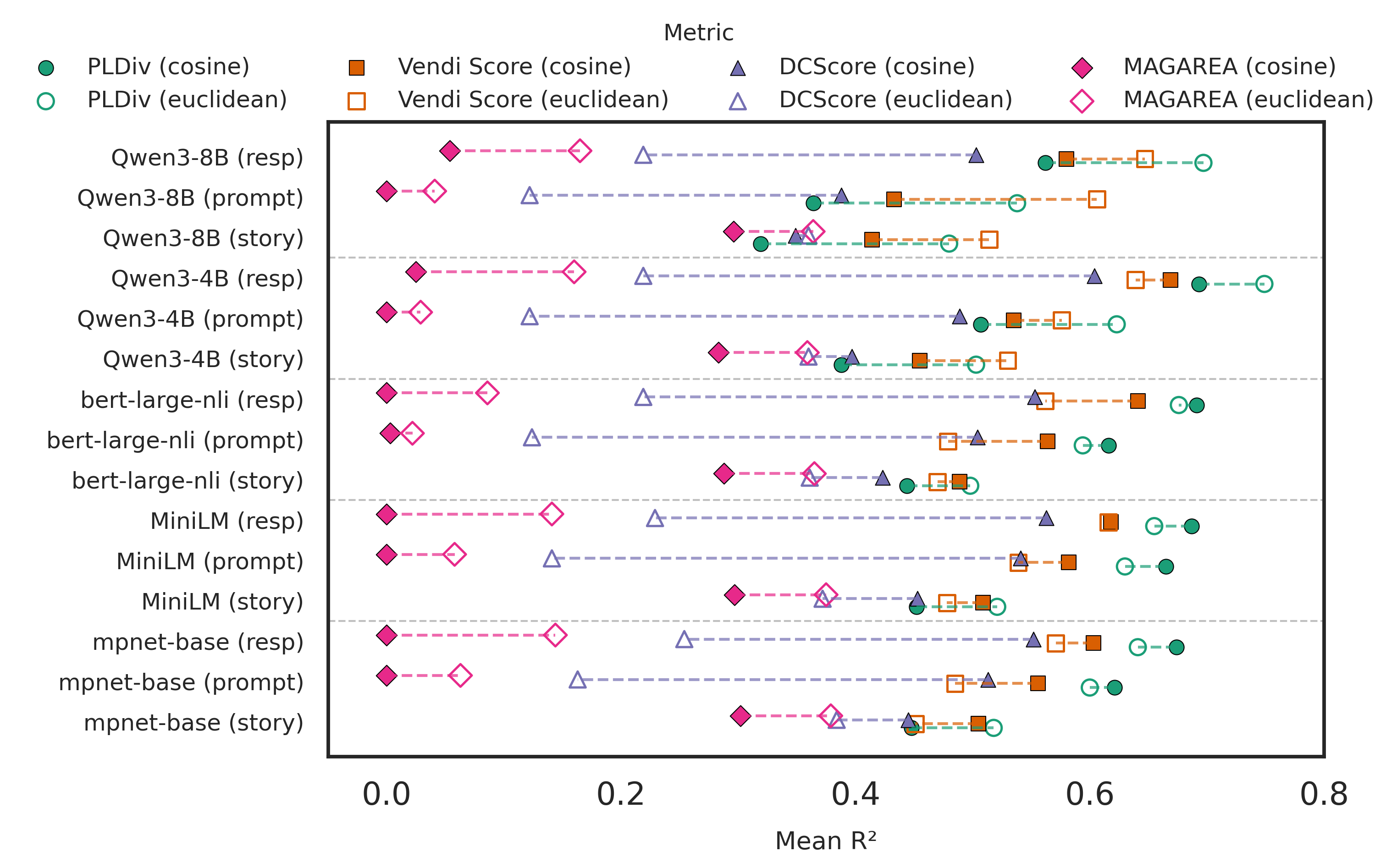}

    \vspace{2mm} 
    
    \includegraphics[width=0.65\textwidth]{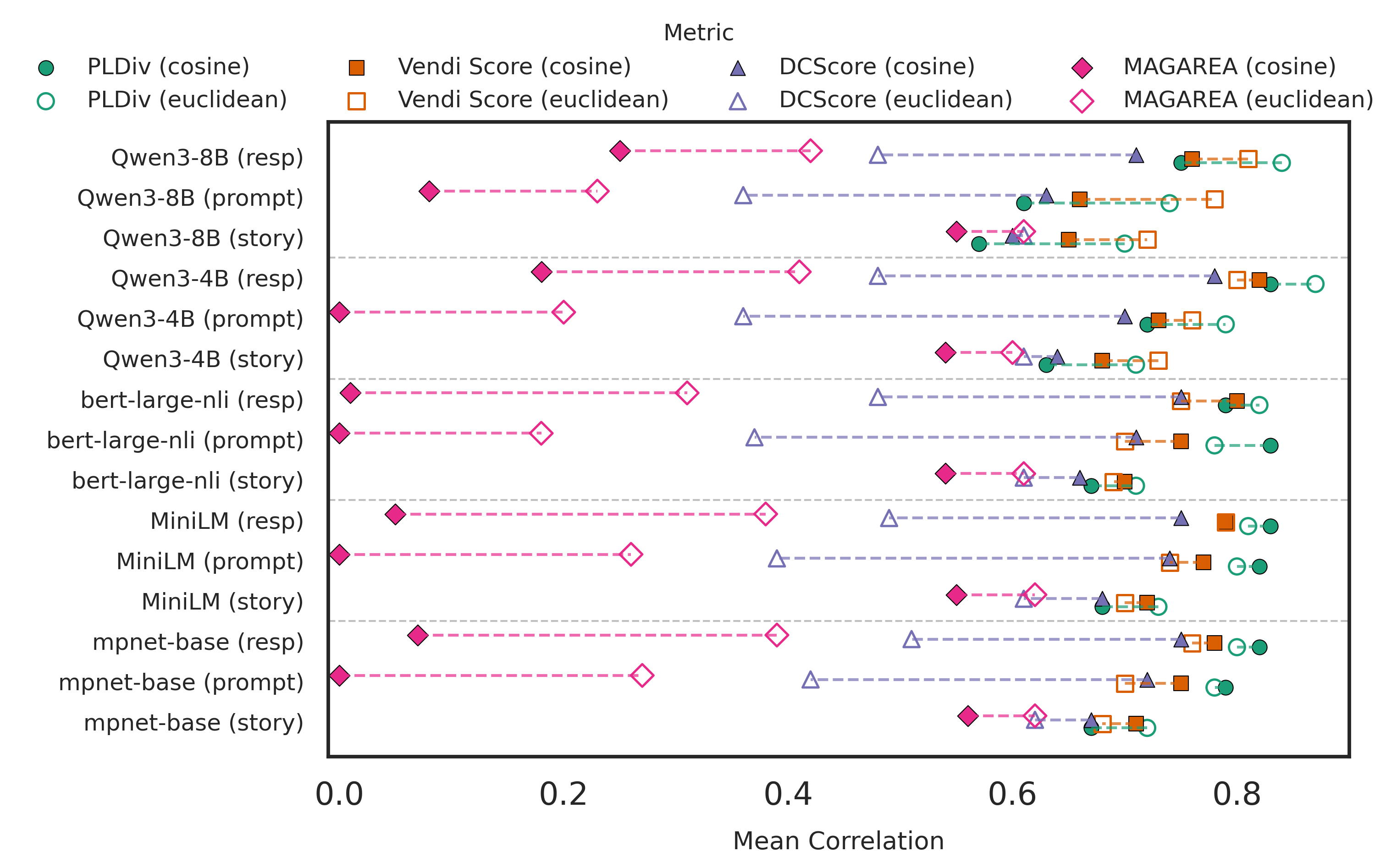}

    \vspace{2mm} 
    \includegraphics[width=0.65\textwidth]{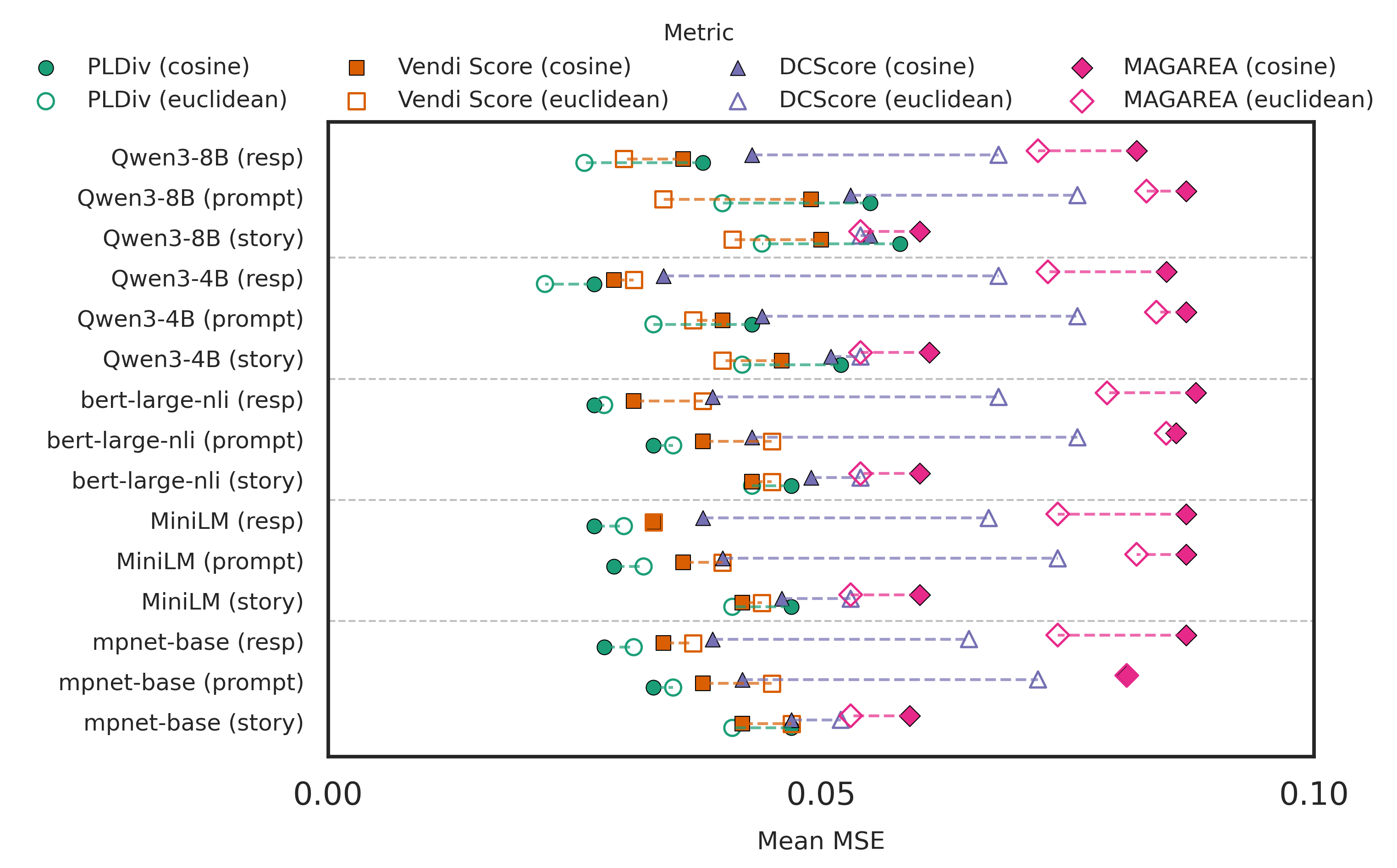}

    \caption{Diversity metric performance is evaluated across different distance/similarity matrices. For Vendi Score and DCScore, the Euclidean setting corresponds to the RBF kernel. PLDiv consistently and reliably outperforms other metrics across various embedding models and distance matrices.}
    \label{fig:cs_eu}
\end{figure*}

To explore the impact of the distance/similarity matrix, we applied both cosine distance/similarity and Euclidean distance/RBF kernel as inputs in this experiment ihe temperature (dec) tasks. Figure \ref{fig:cs_eu} demonstrates that PLDiv consistently and reliably outperforms other metrics across various embedding models and distance matrices. In contrast, switching from cosine similarity to the RBF kernel significantly degrades the performance of alternative metrics, particularly DCScore.

\begin{figure*}[h!]
    \centering
    \includegraphics[width=0.24\textwidth]{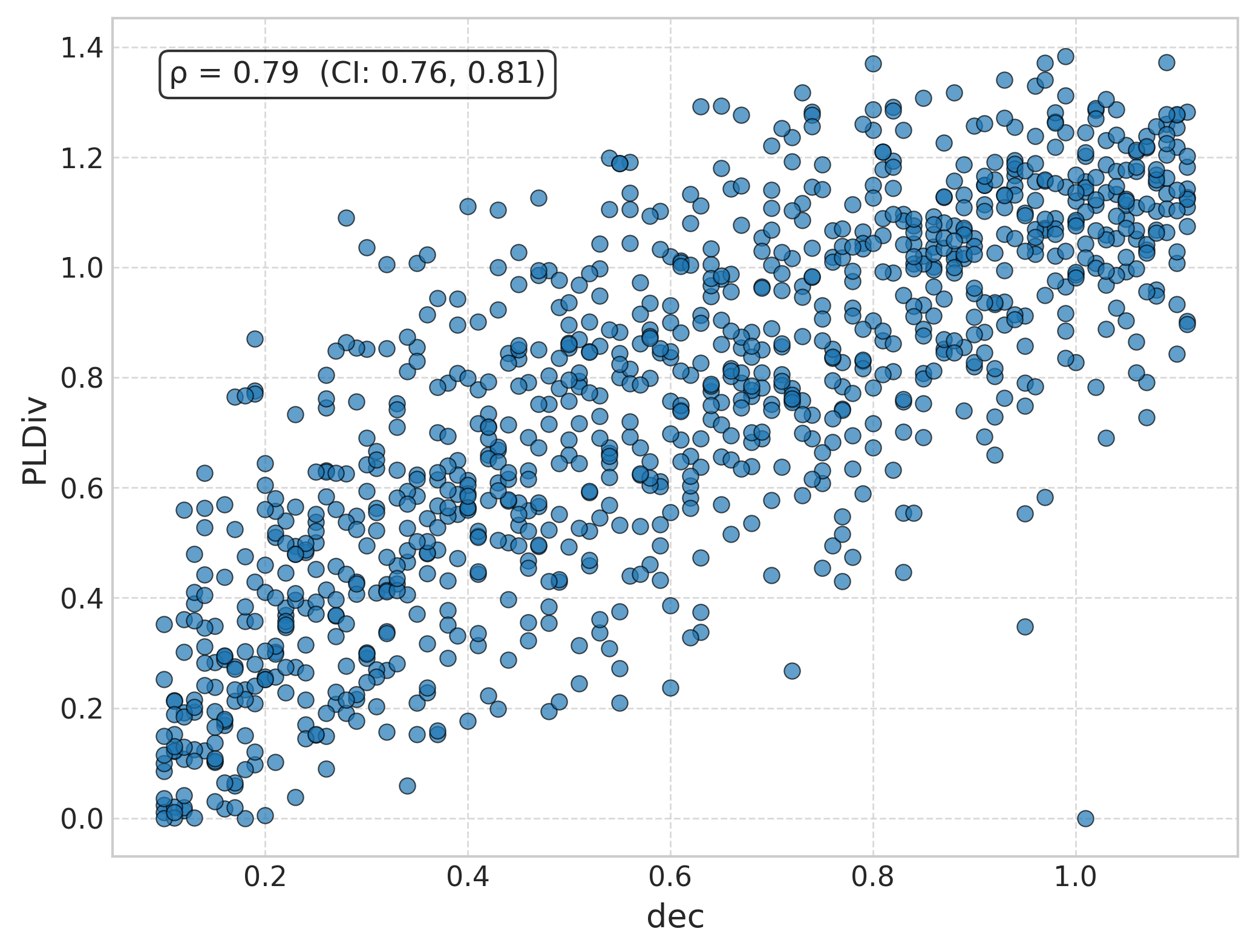}
    \includegraphics[width=0.24\textwidth]{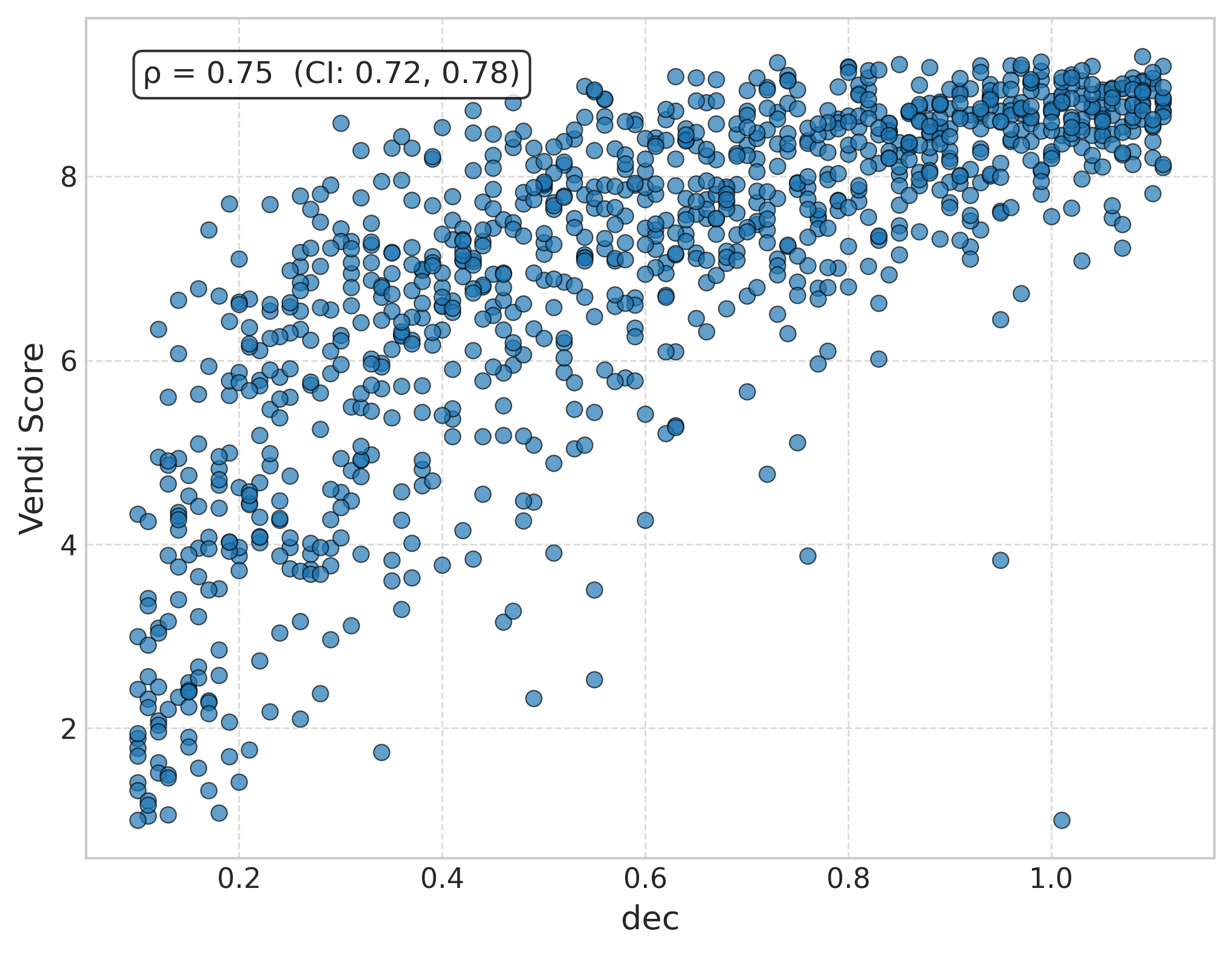}
    \includegraphics[width=0.24\textwidth]{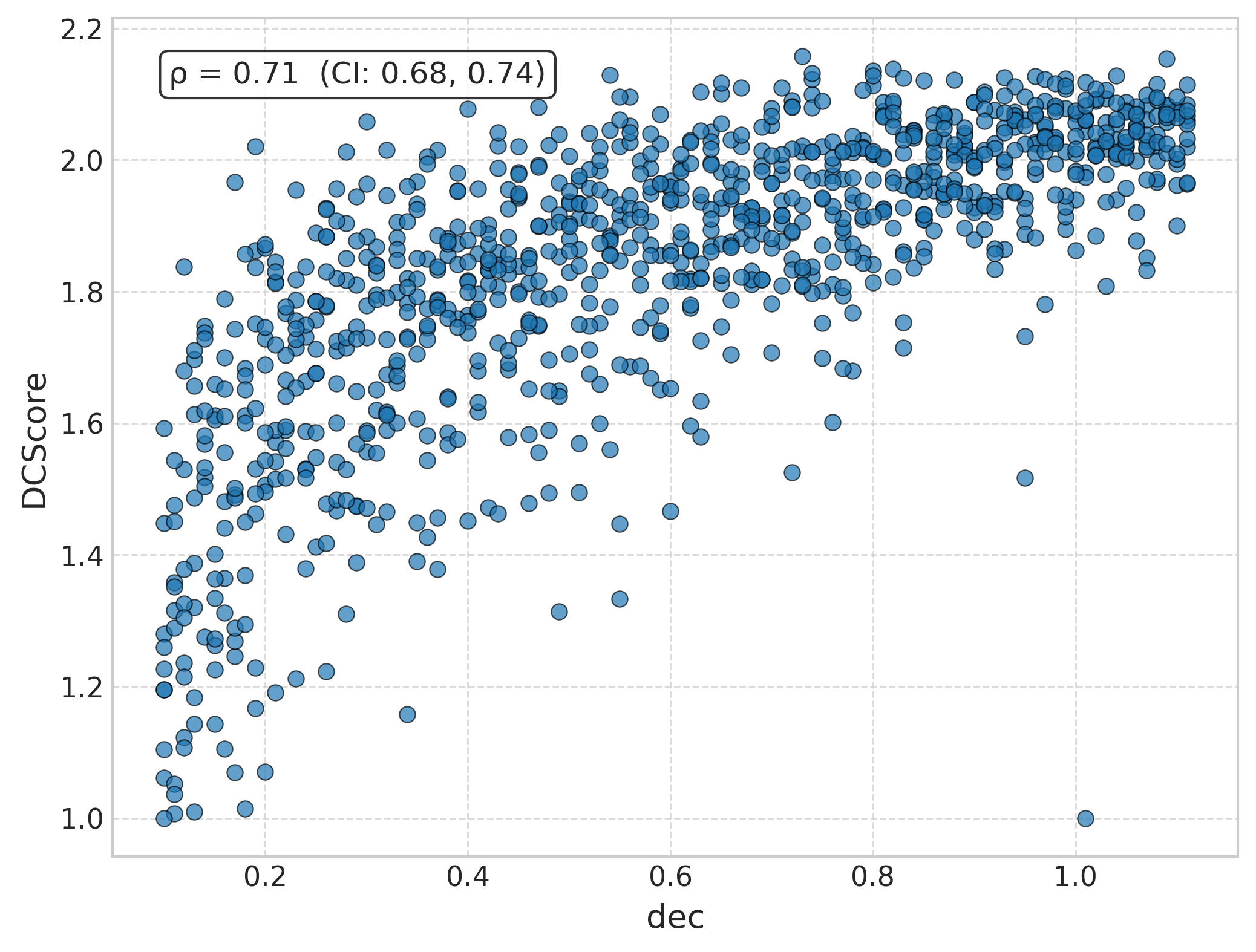}
    \includegraphics[width=0.24\textwidth]{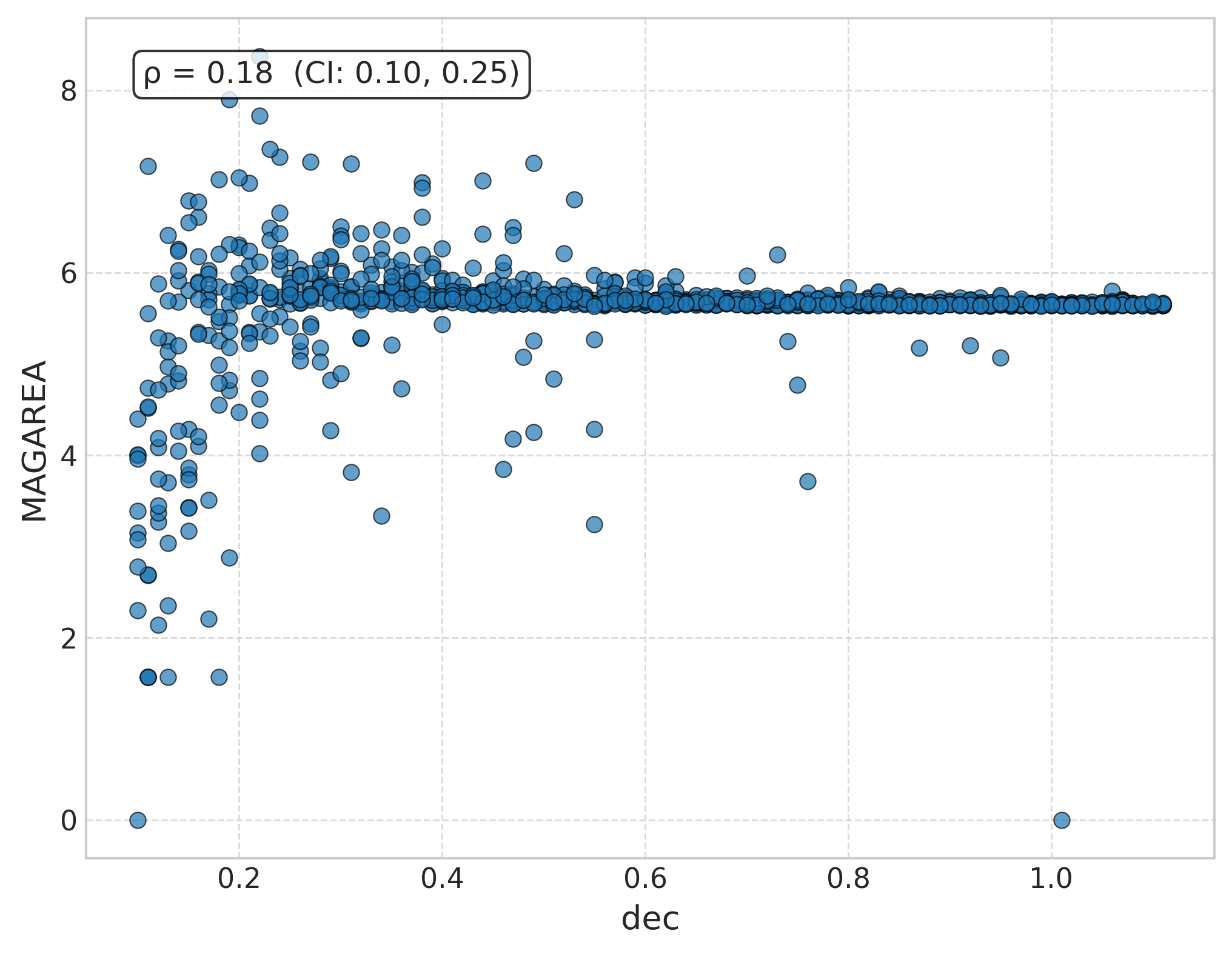}
    
    \vspace{2mm} 
    
    \includegraphics[width=0.24\textwidth]{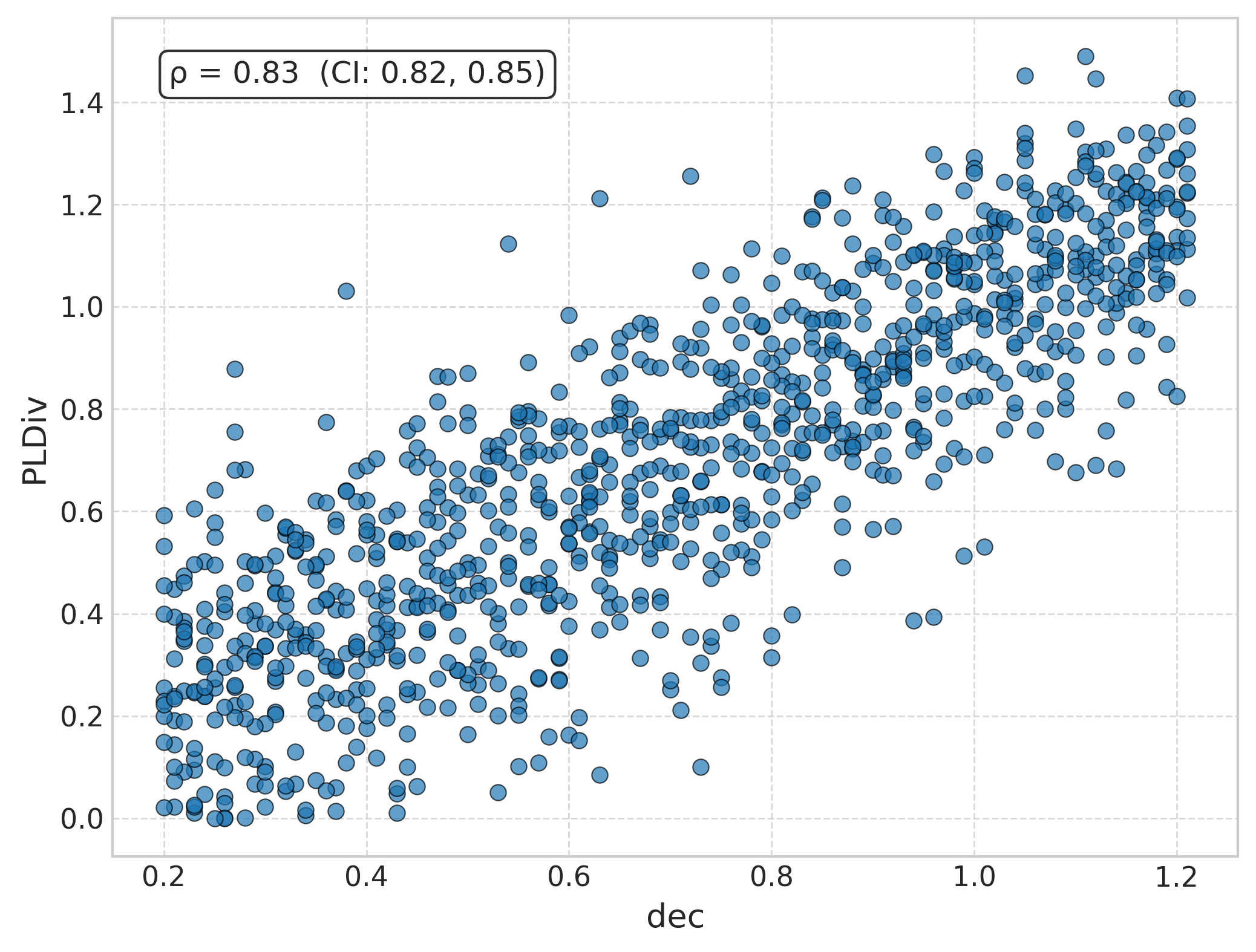}
    \includegraphics[width=0.24\textwidth]{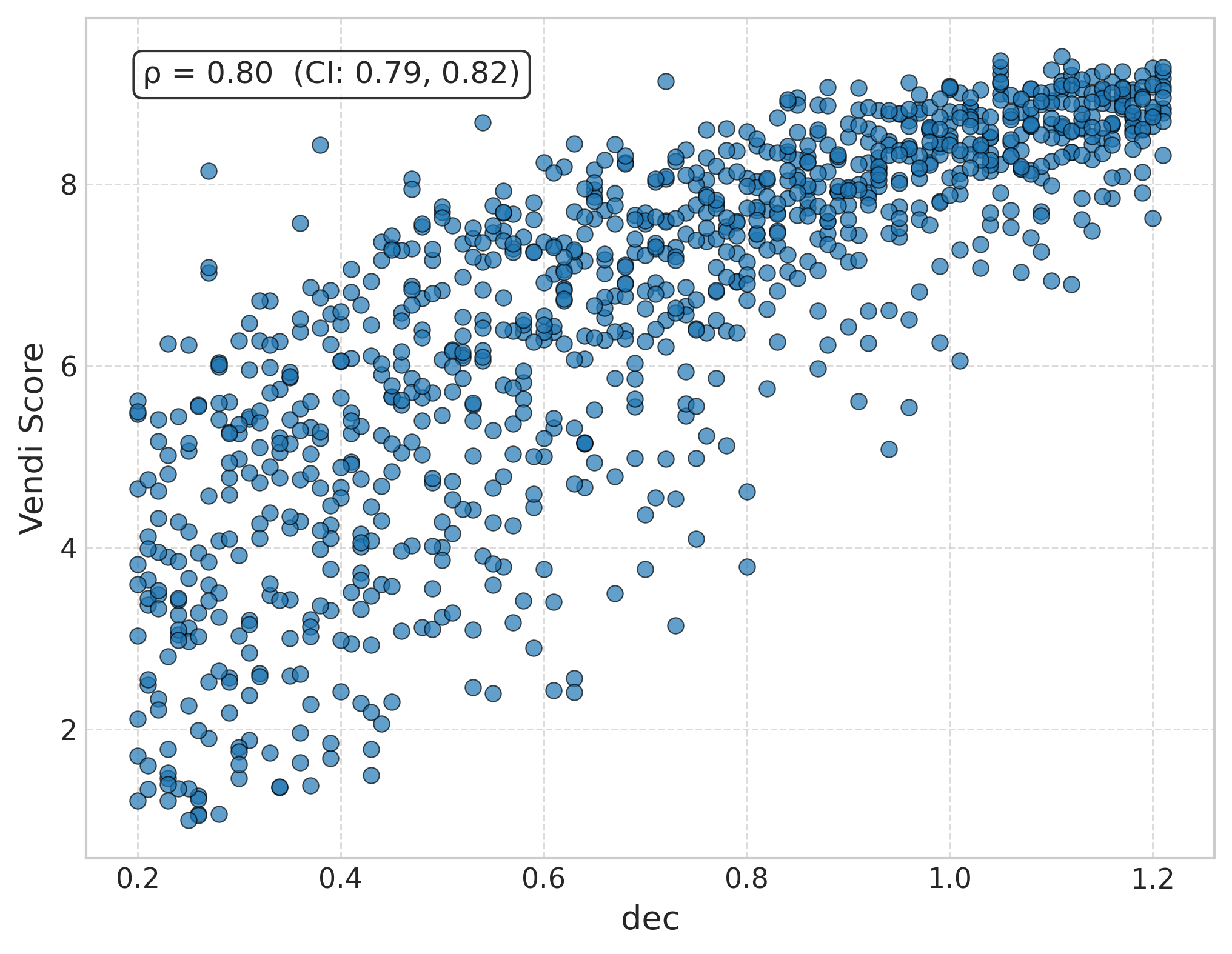}
    \includegraphics[width=0.24\textwidth]{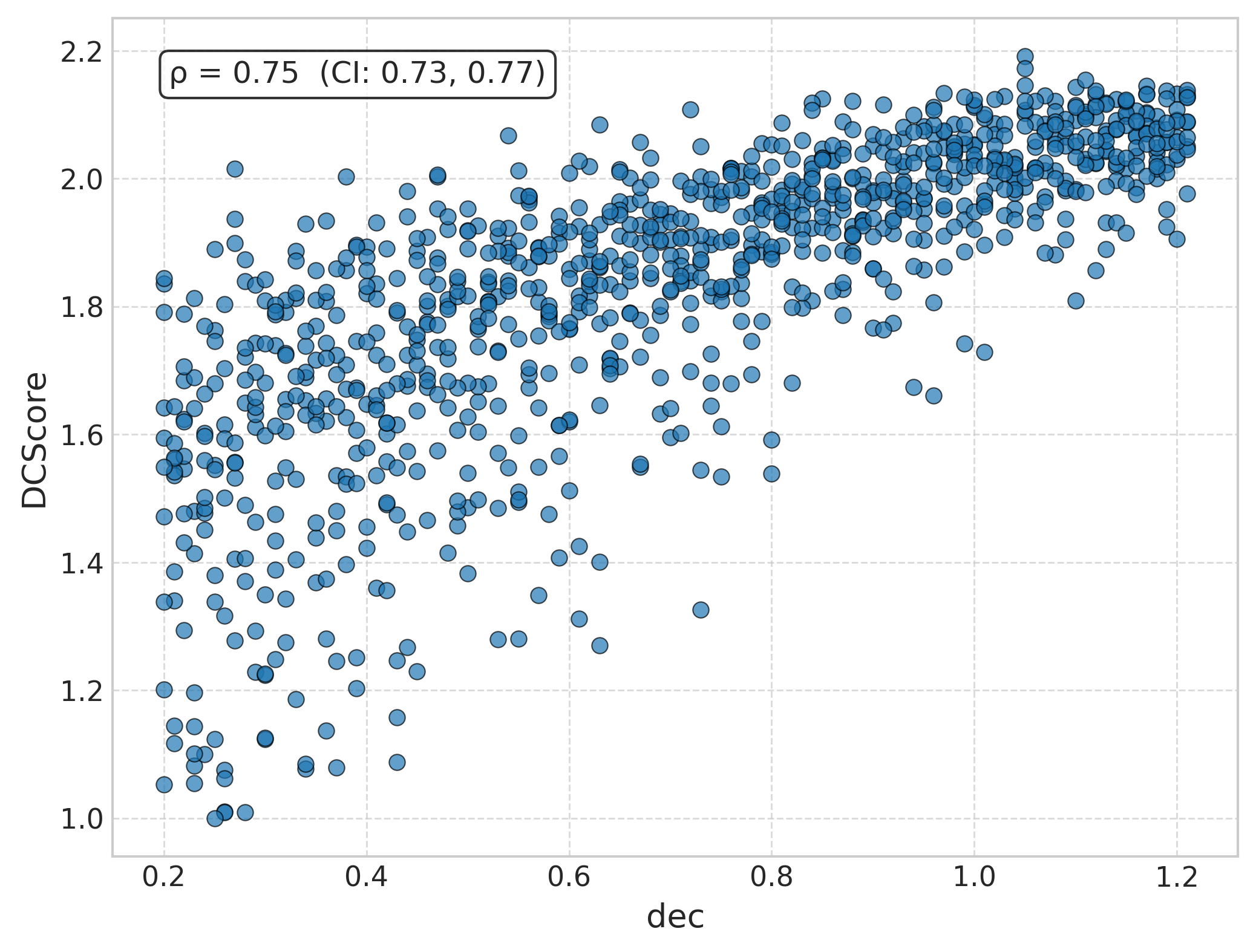}
    \includegraphics[width=0.24\textwidth]{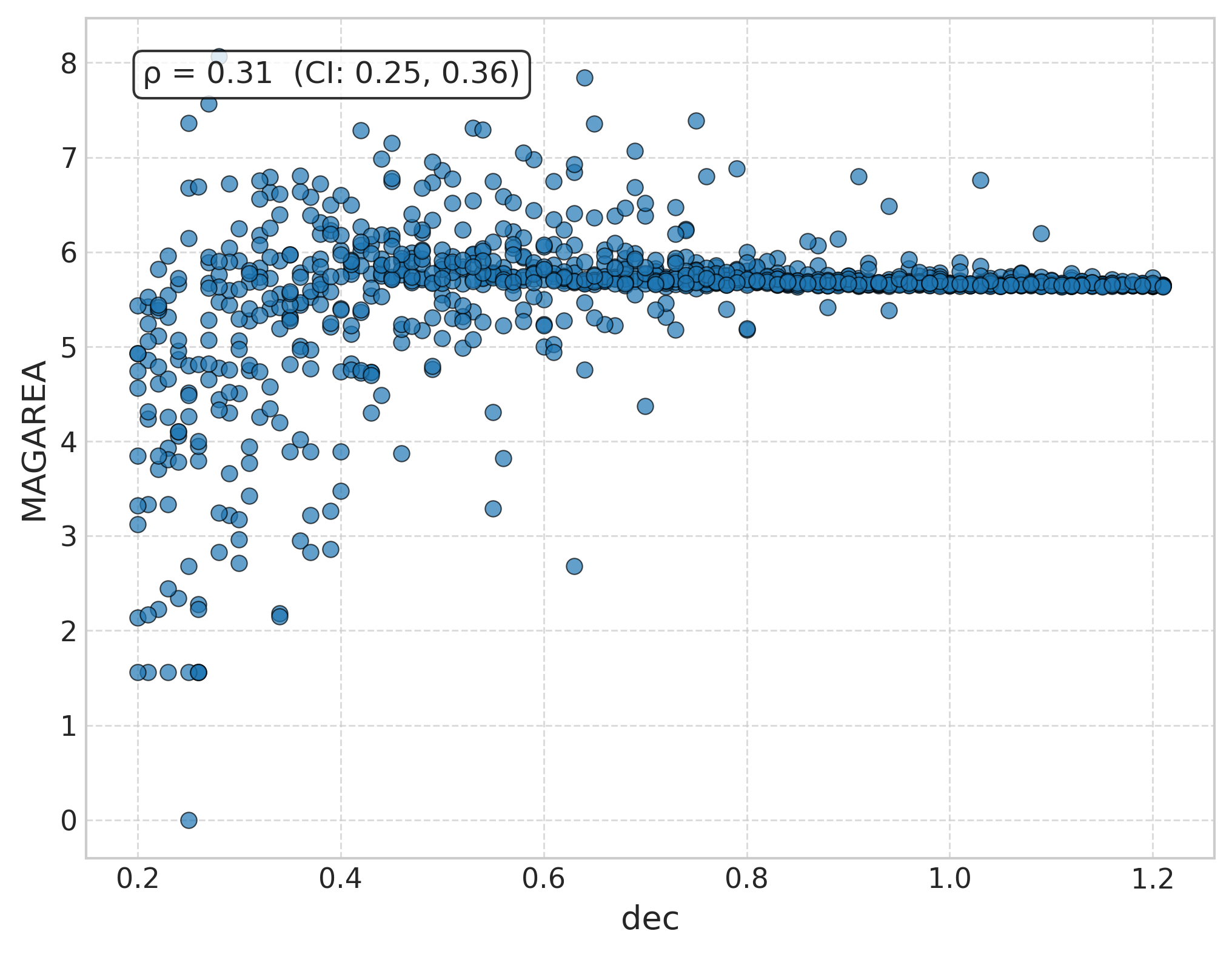}
    
    \vspace{2mm} 
    
    \includegraphics[width=0.24\textwidth]{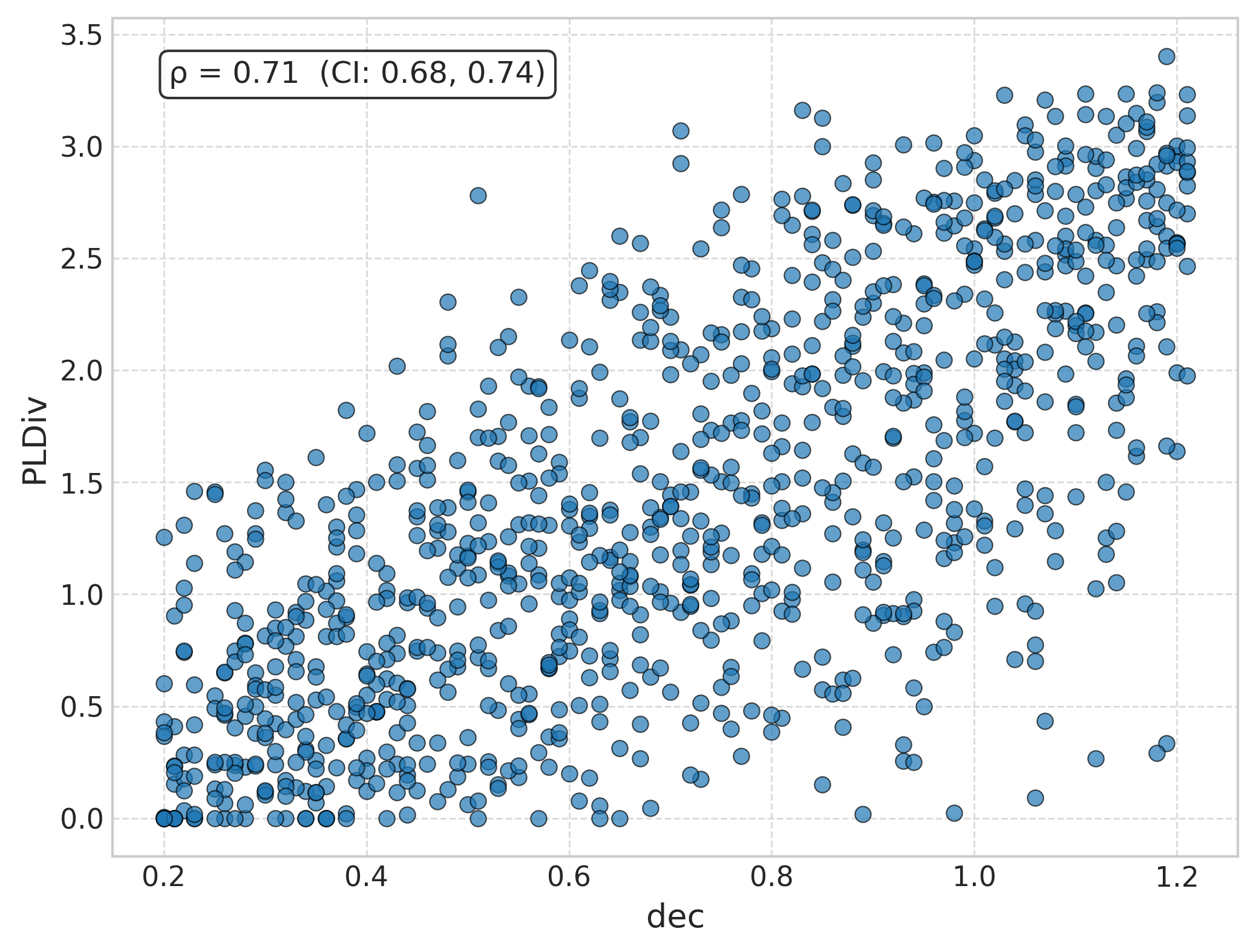}
    \includegraphics[width=0.24\textwidth]{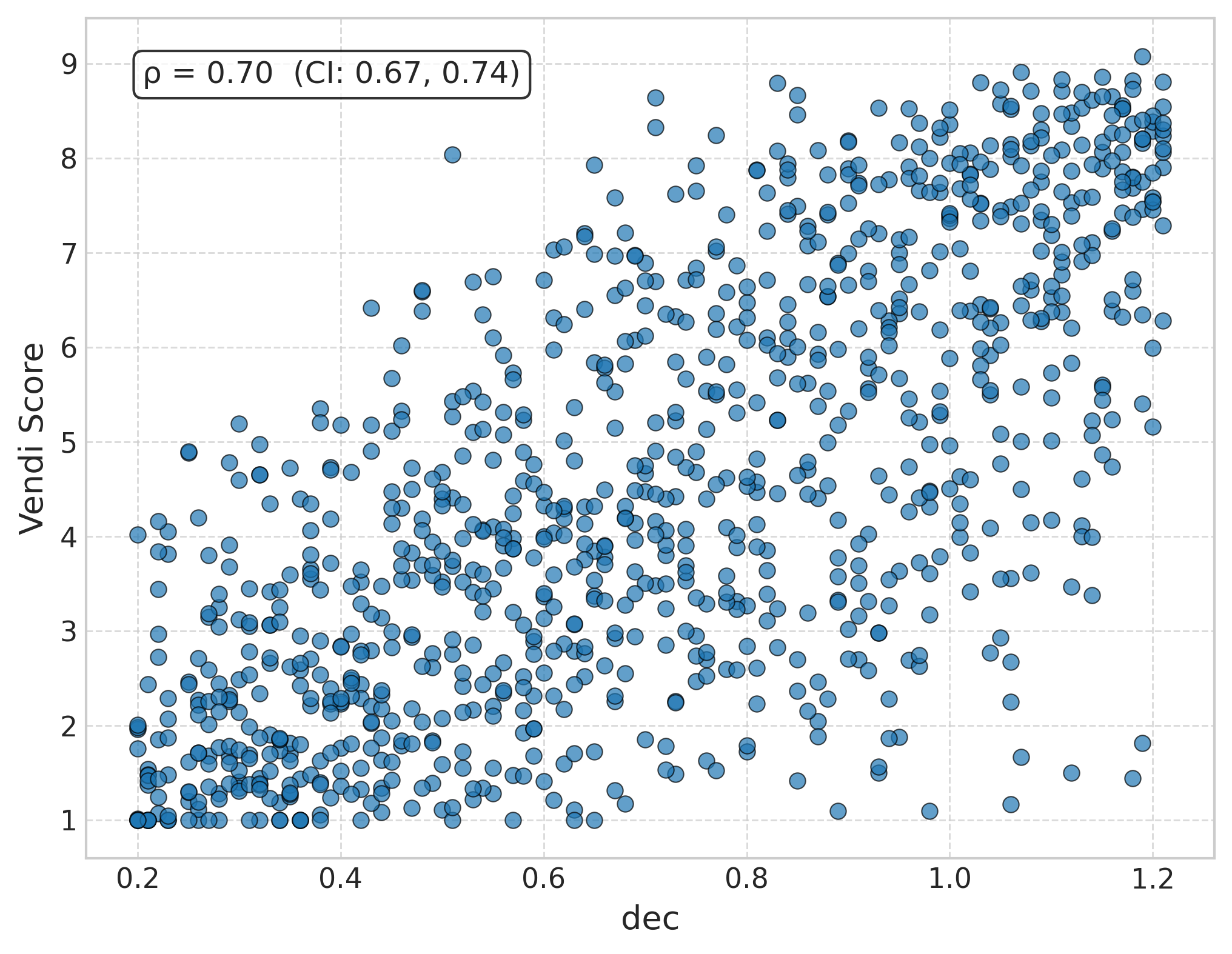}
    \includegraphics[width=0.24\textwidth]{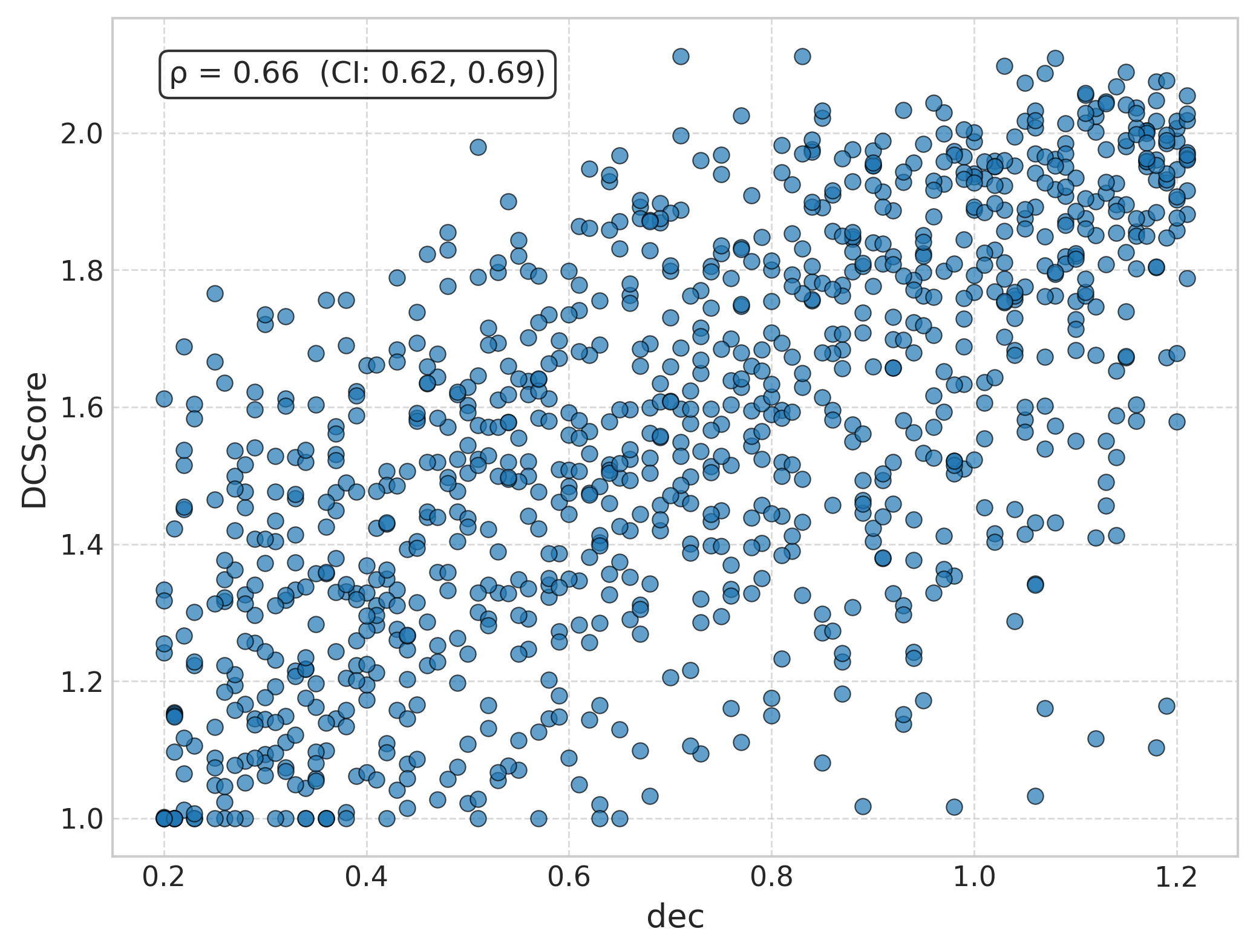}
    \includegraphics[width=0.24\textwidth]{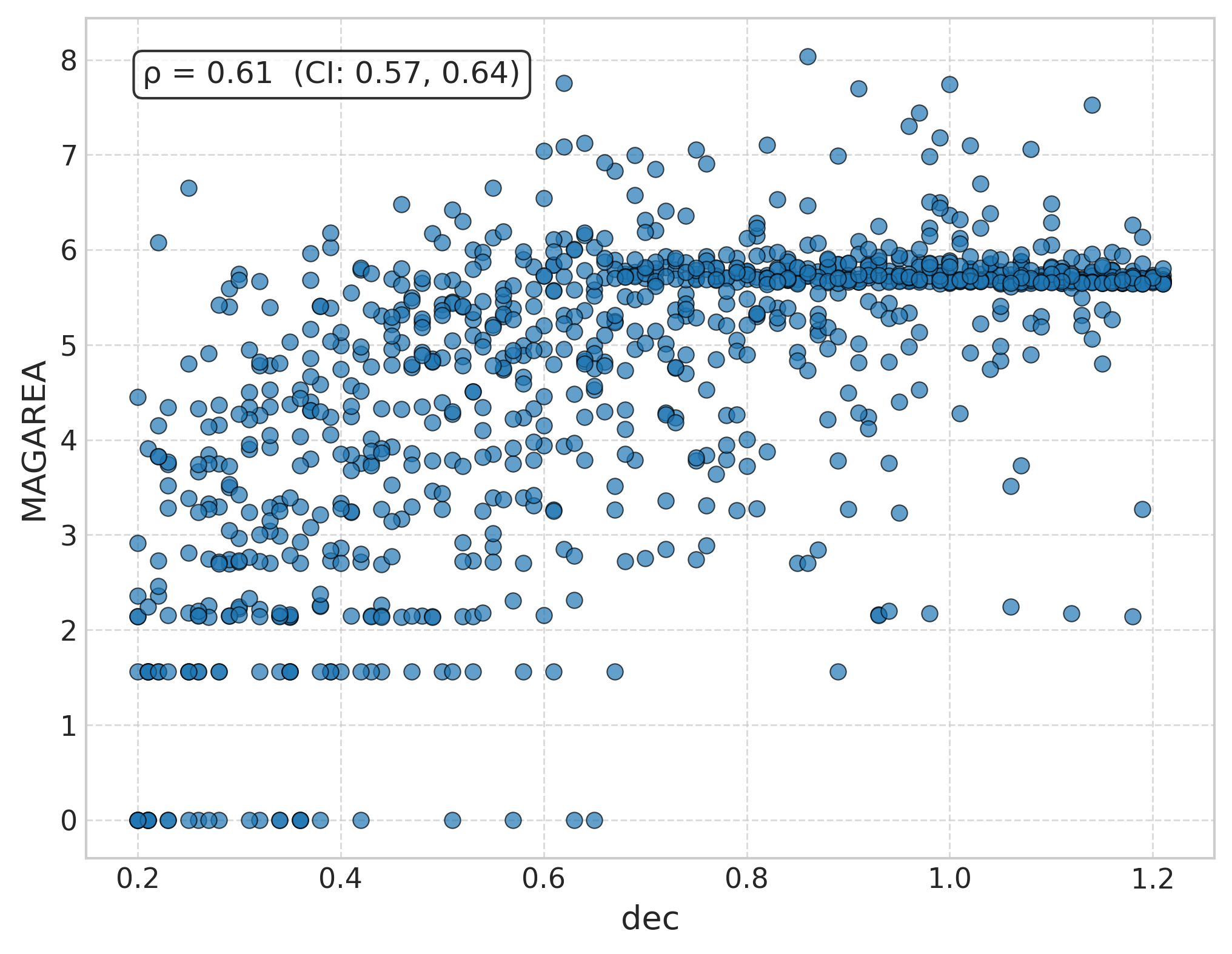}
    
    \caption{Correlation results for embeddings model: ``bert-large-nli-stsb-mean-tokens" across three tasks: Row 1 shows prompt, Row 2 shows response, and Row 3 shows story. Columns 1–4 represent the results for PLDiv, VS, DCS, and MagArea, respectively.}
    \label{fig:bert}
\end{figure*}

\begin{figure*}[h!]
    \centering
     \includegraphics[width=0.24\textwidth]{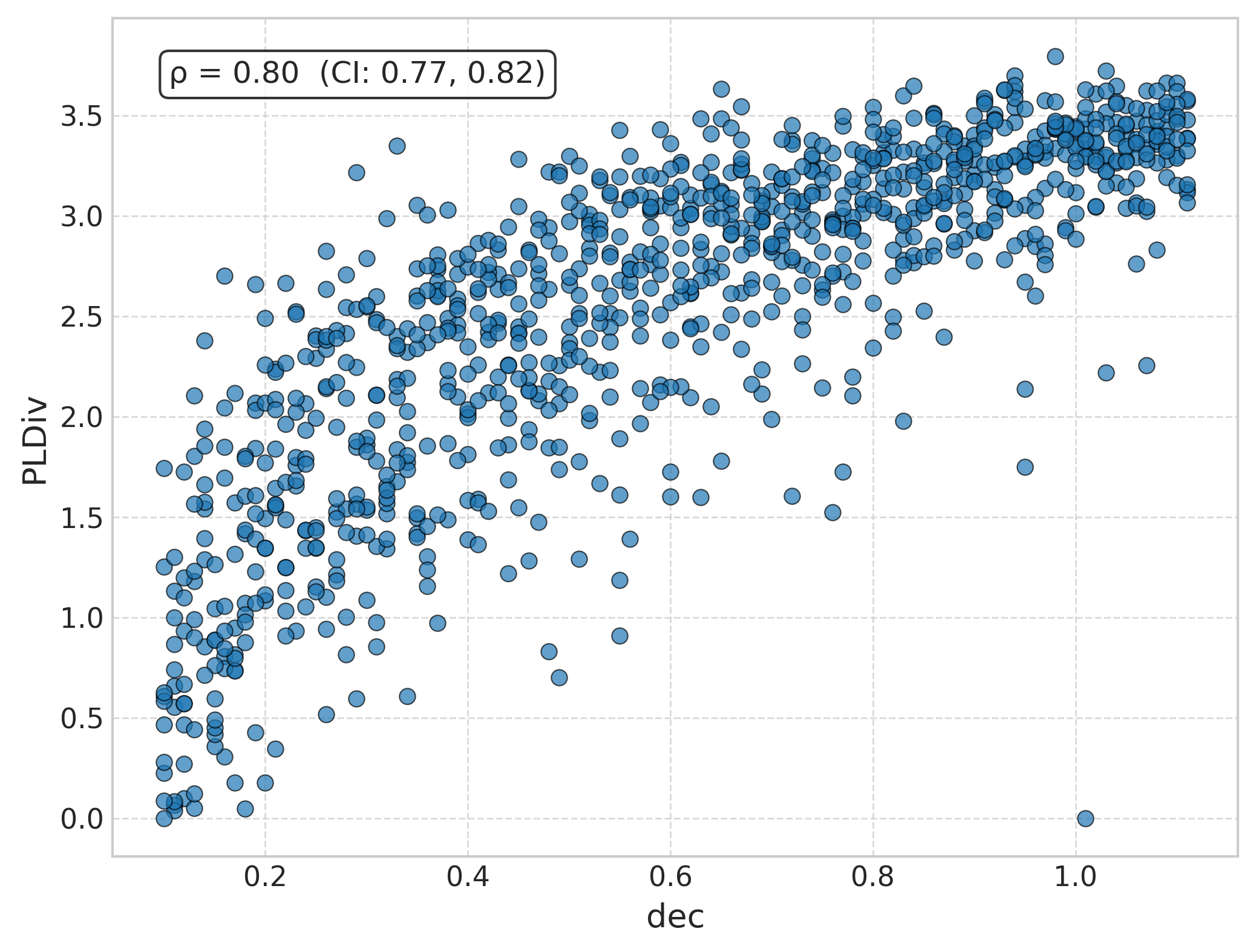}
    \includegraphics[width=0.24\textwidth]{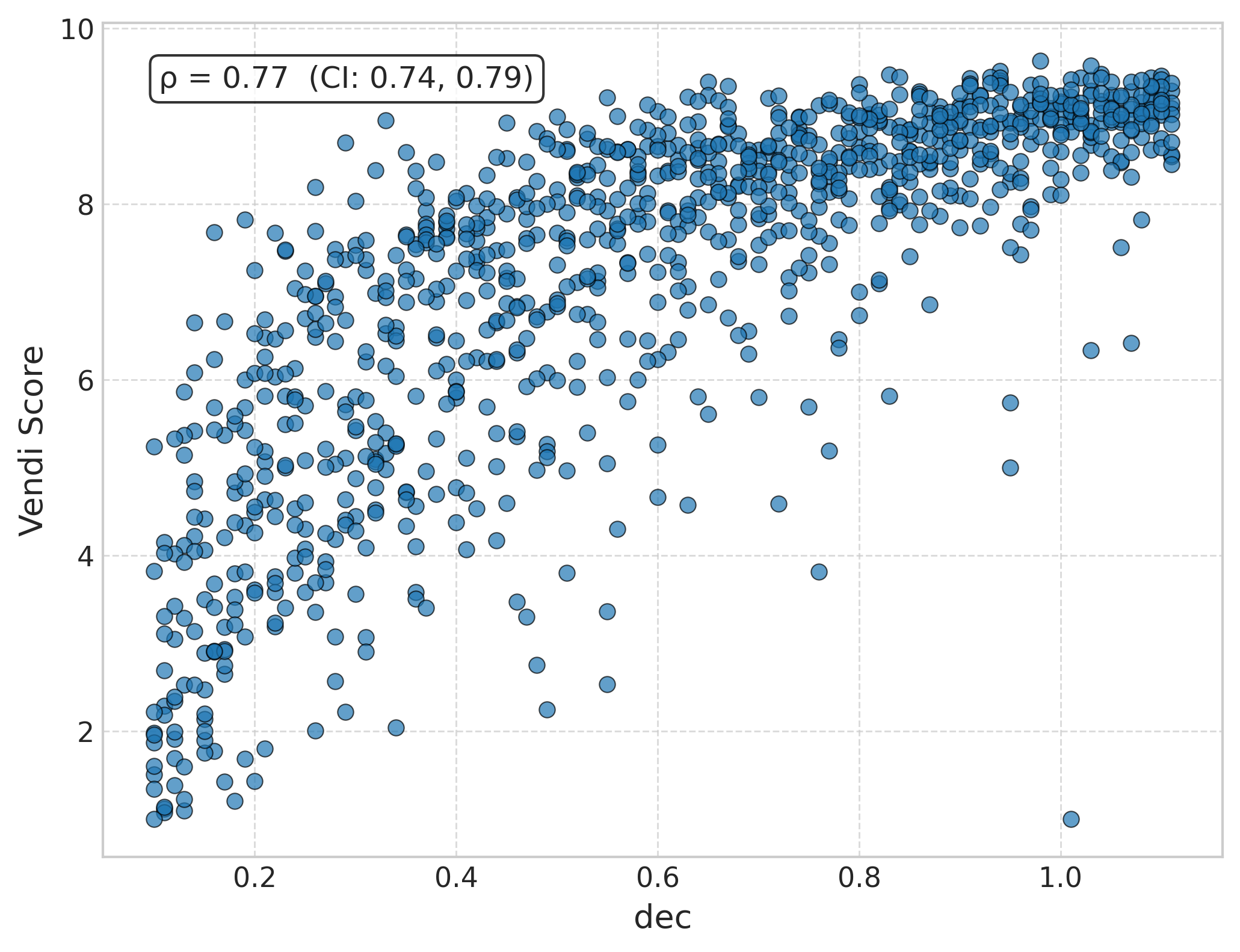}
    \includegraphics[width=0.24\textwidth]{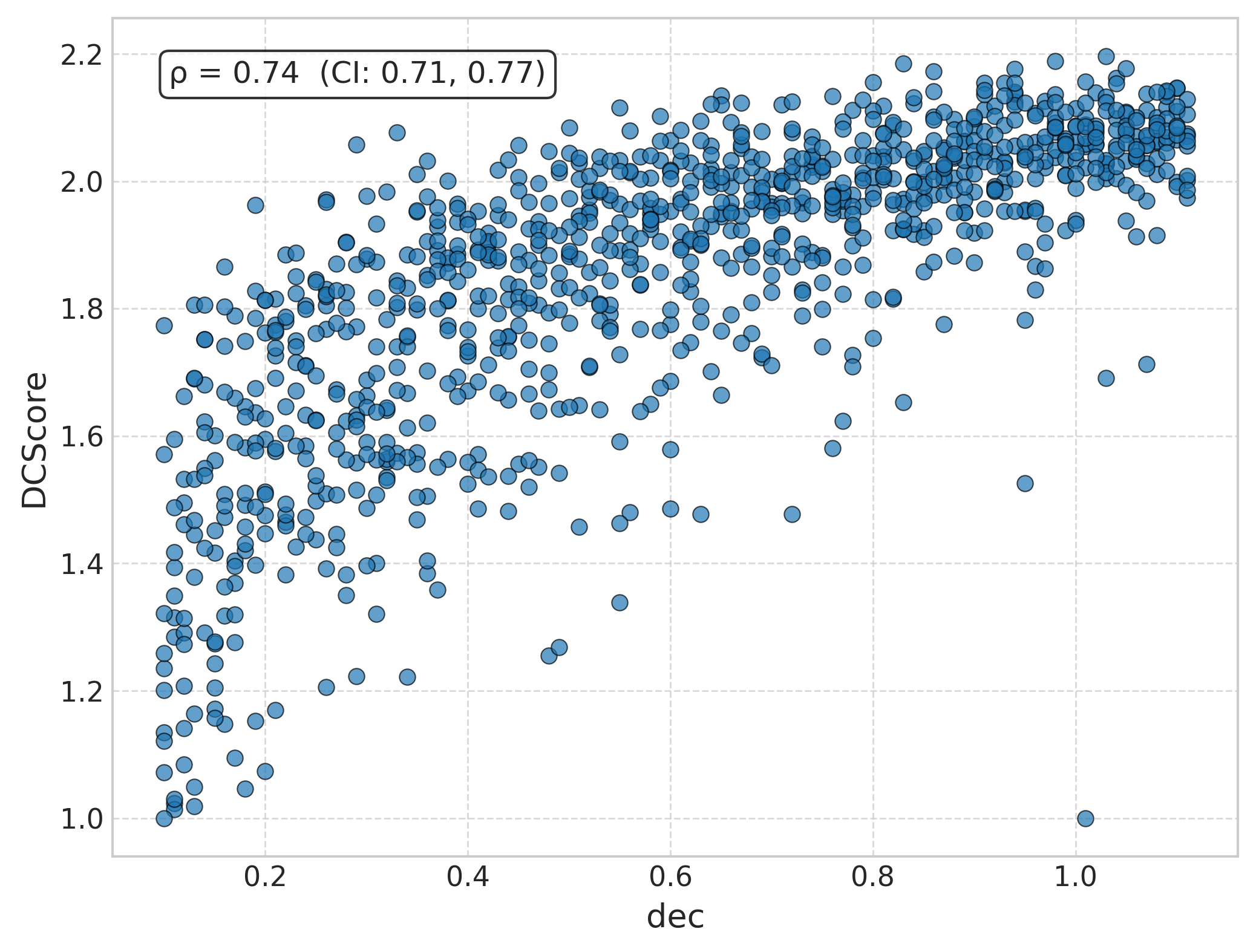}
    \includegraphics[width=0.24\textwidth]{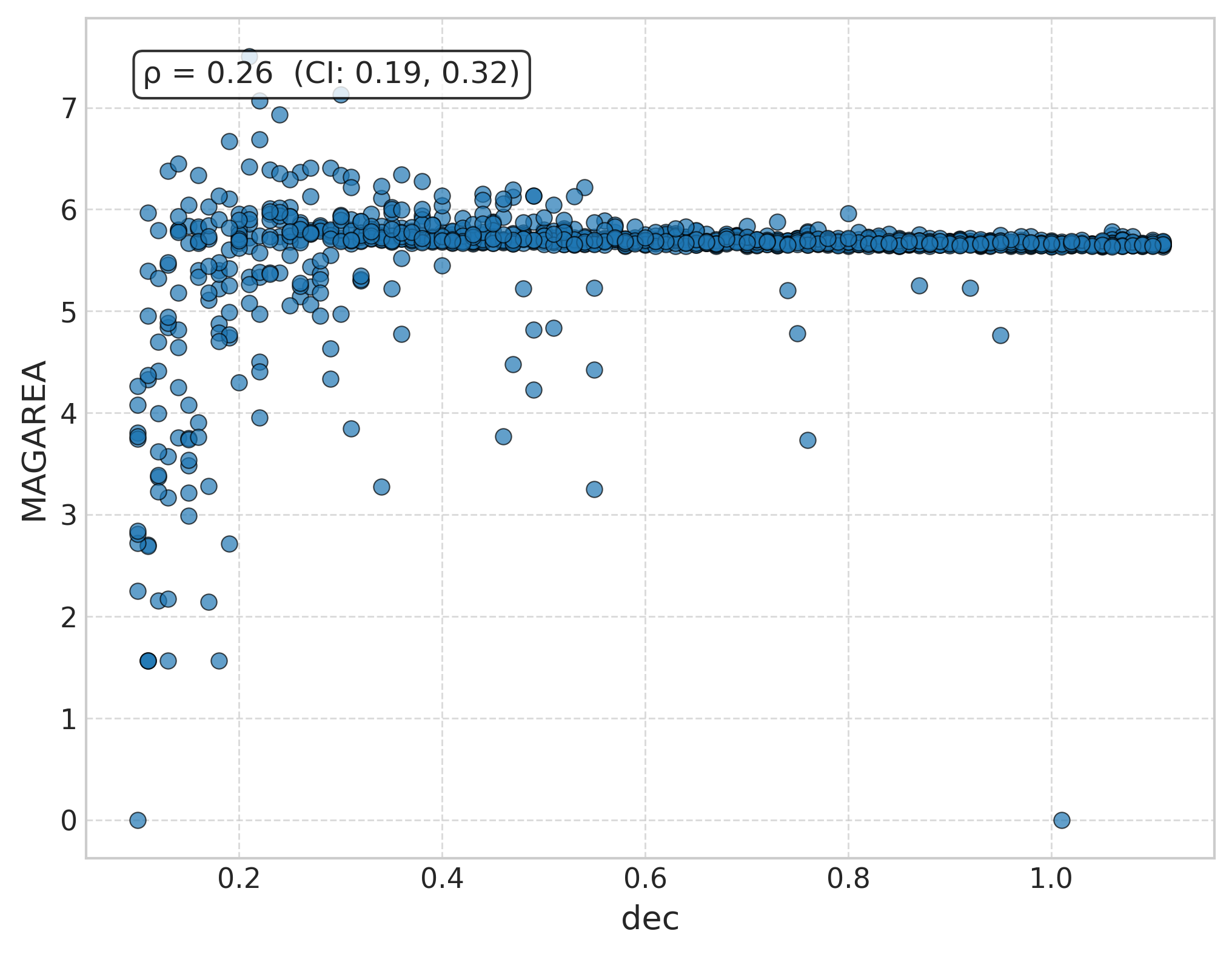}
    
    \vspace{2mm} 
    
    \includegraphics[width=0.24\textwidth]{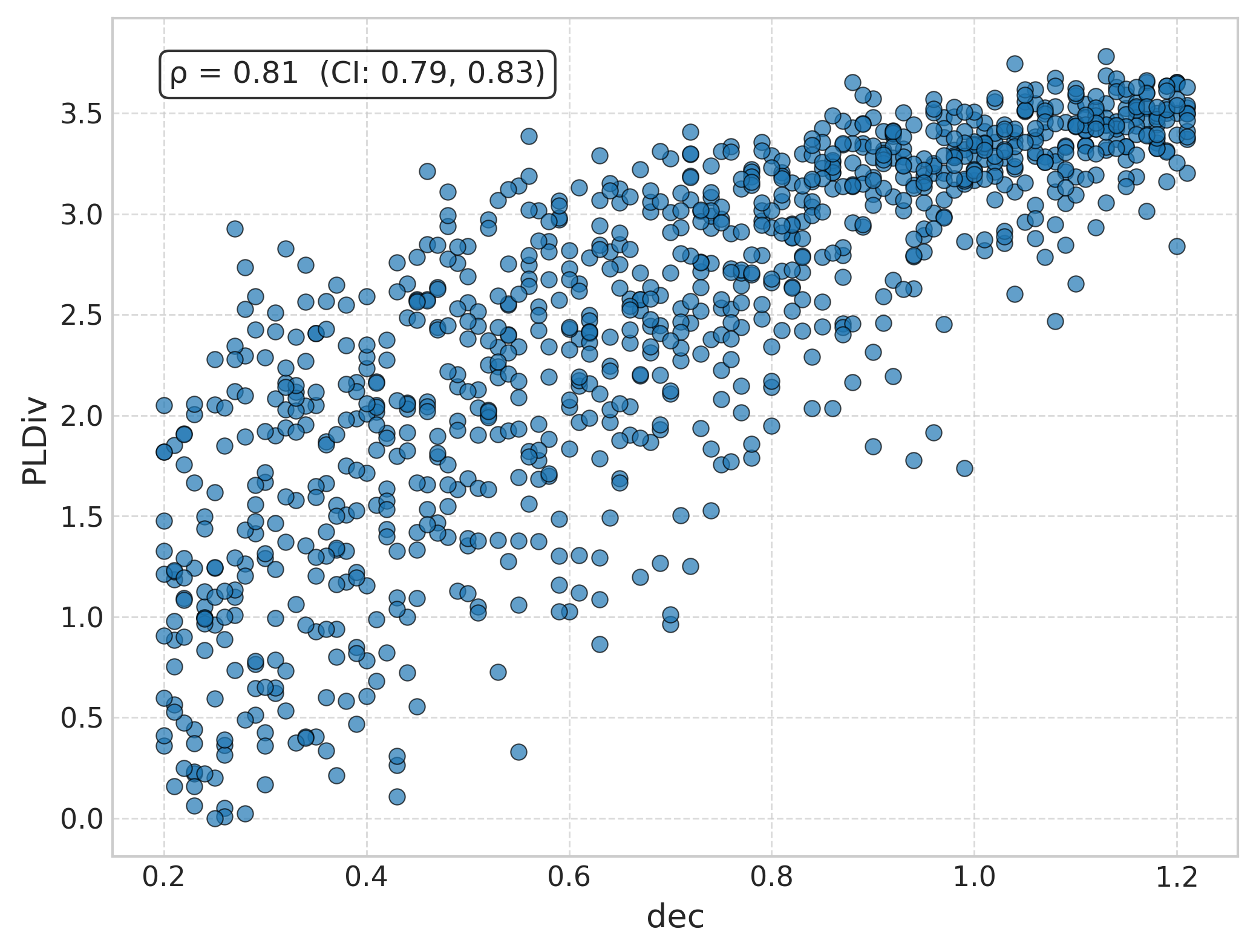}
    \includegraphics[width=0.24\textwidth]{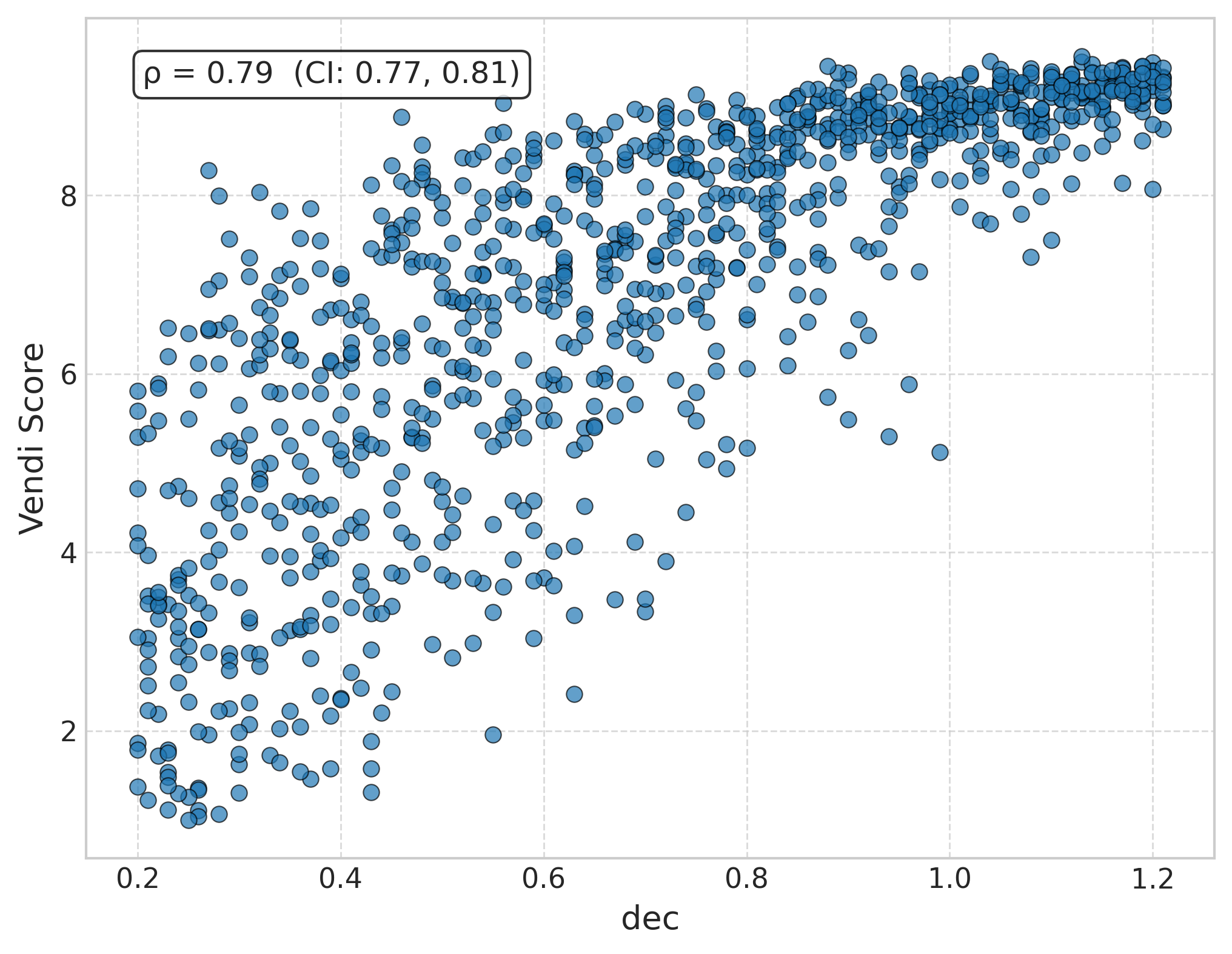}
    \includegraphics[width=0.24\textwidth]{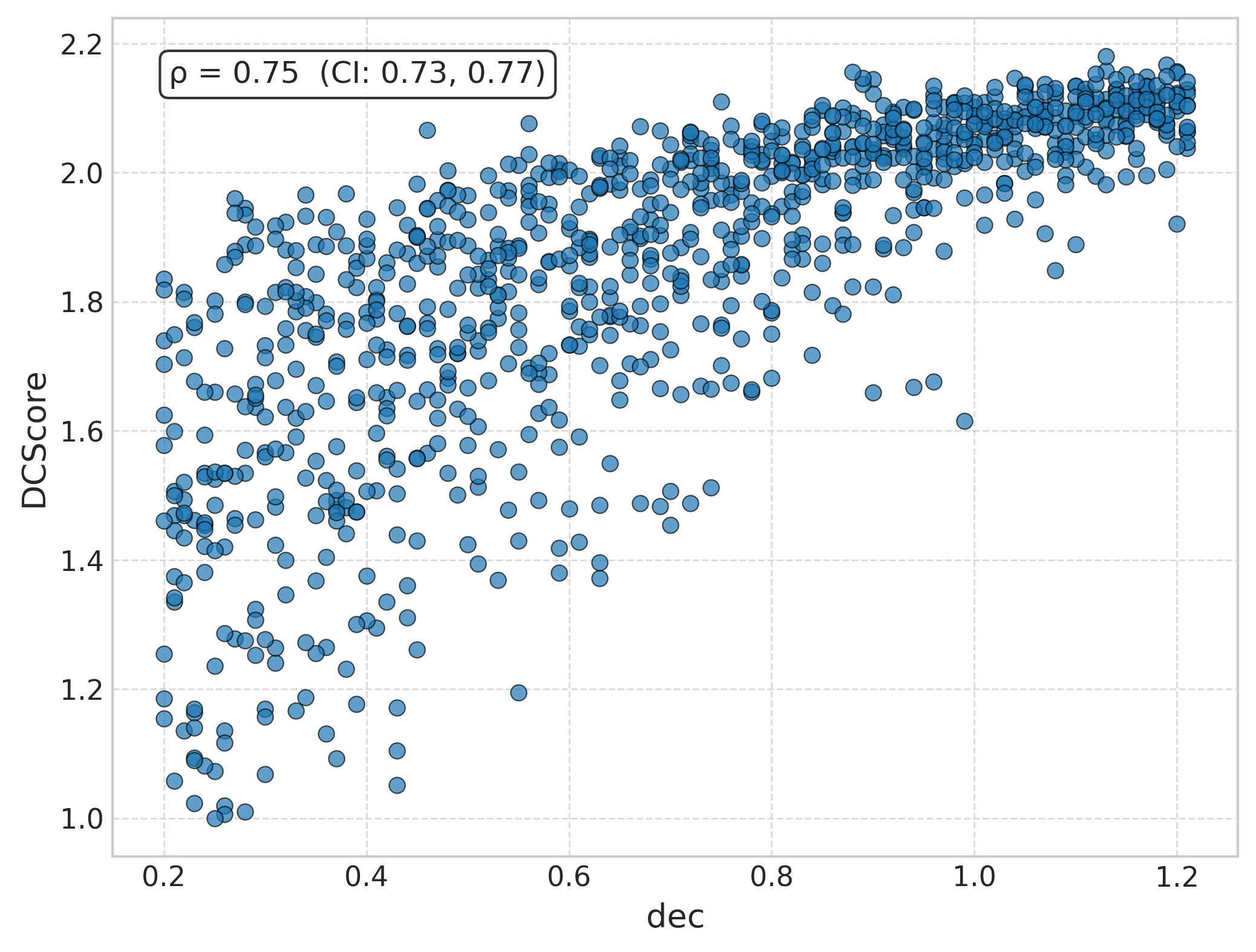}
    \includegraphics[width=0.24\textwidth]{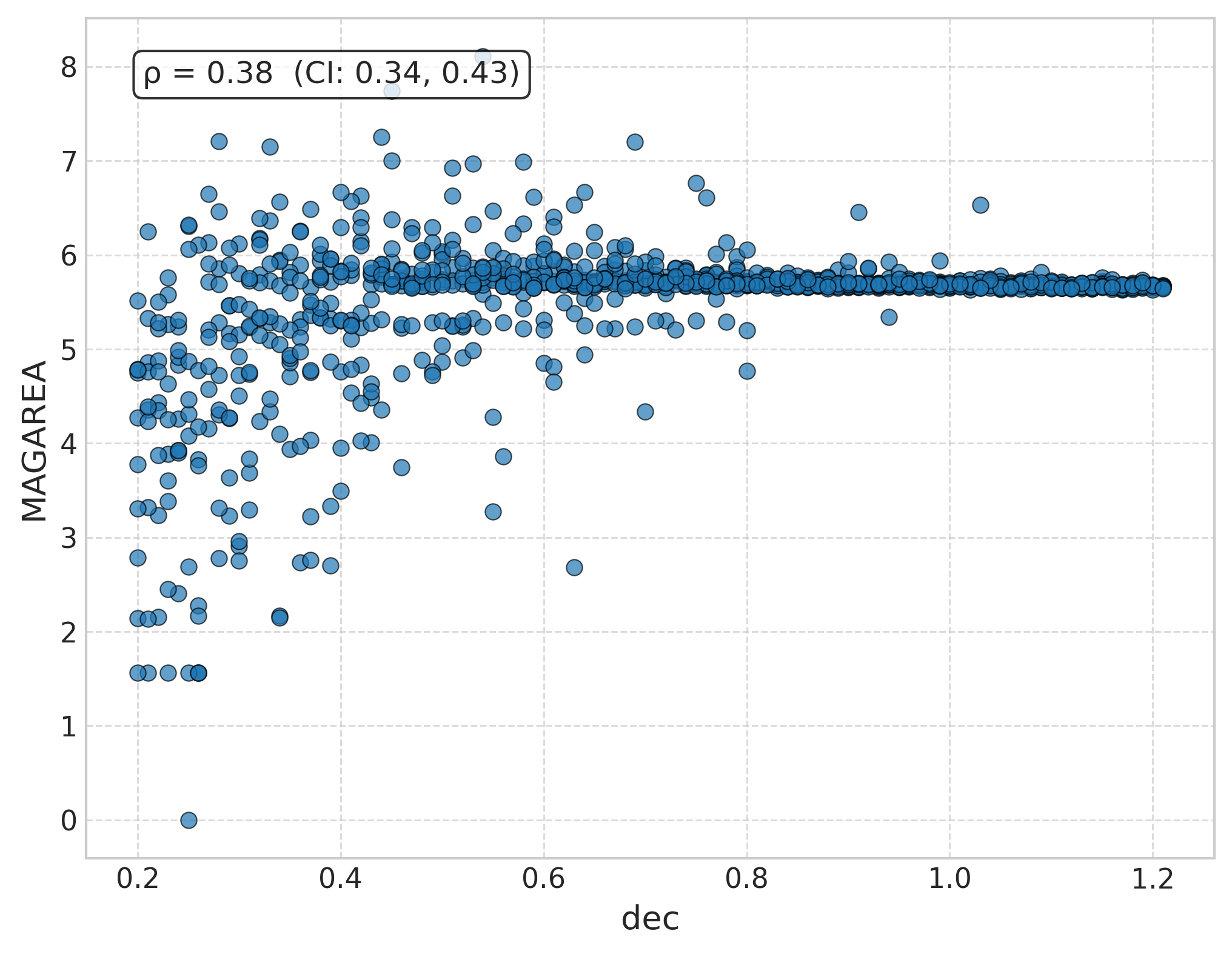}
    
    \vspace{2mm} 
    
    \includegraphics[width=0.24\textwidth]{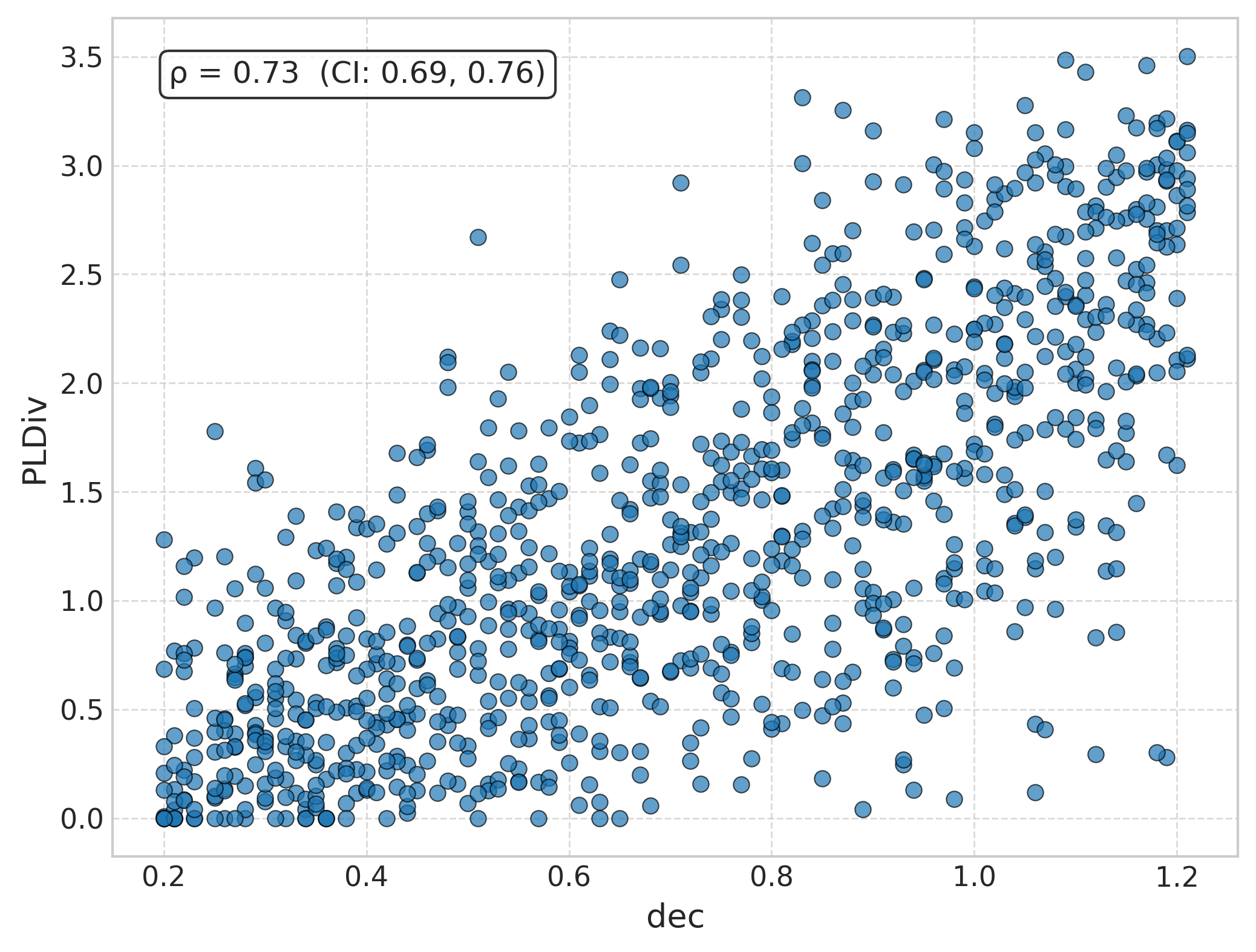}
    \includegraphics[width=0.24\textwidth]{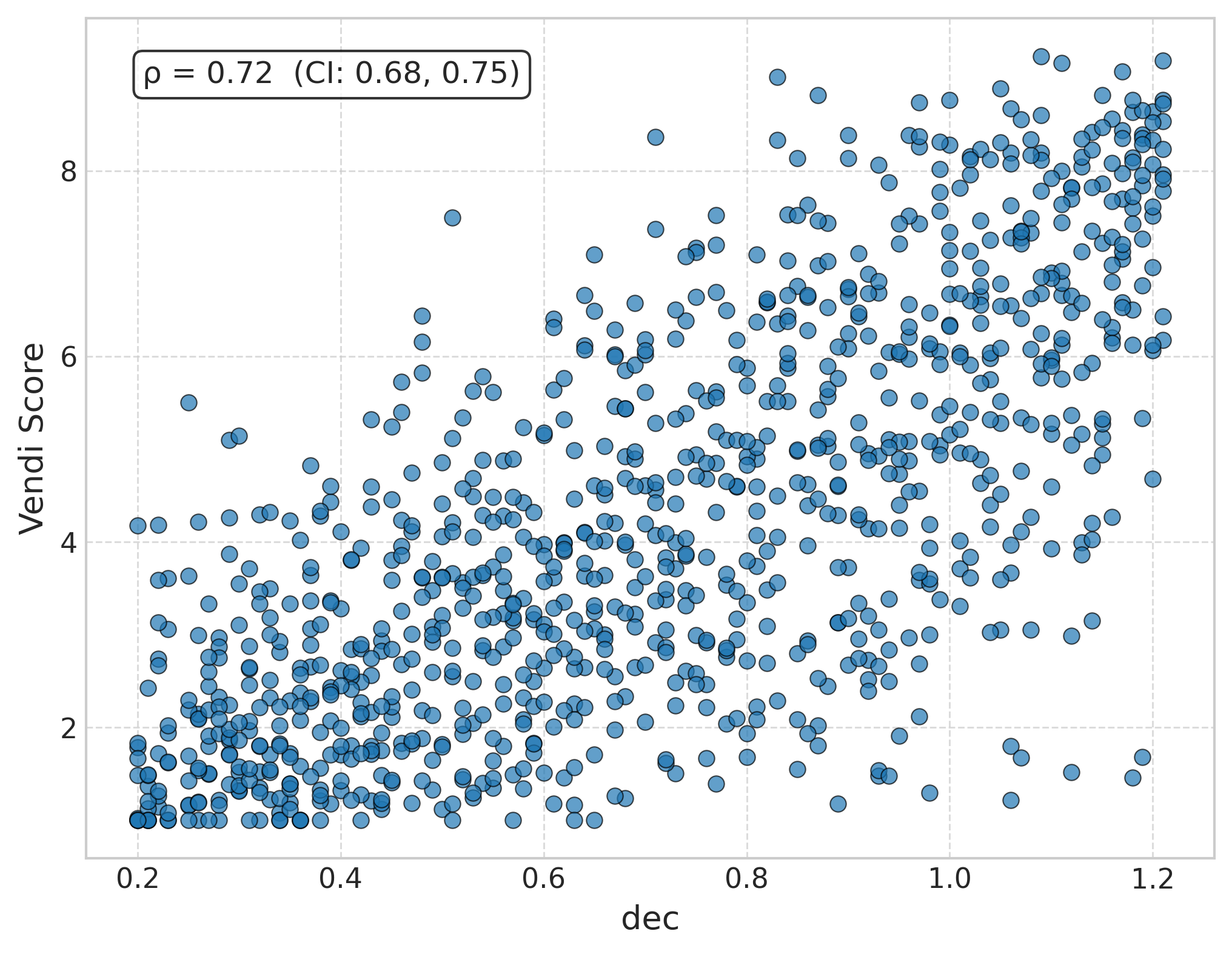}
    \includegraphics[width=0.24\textwidth]{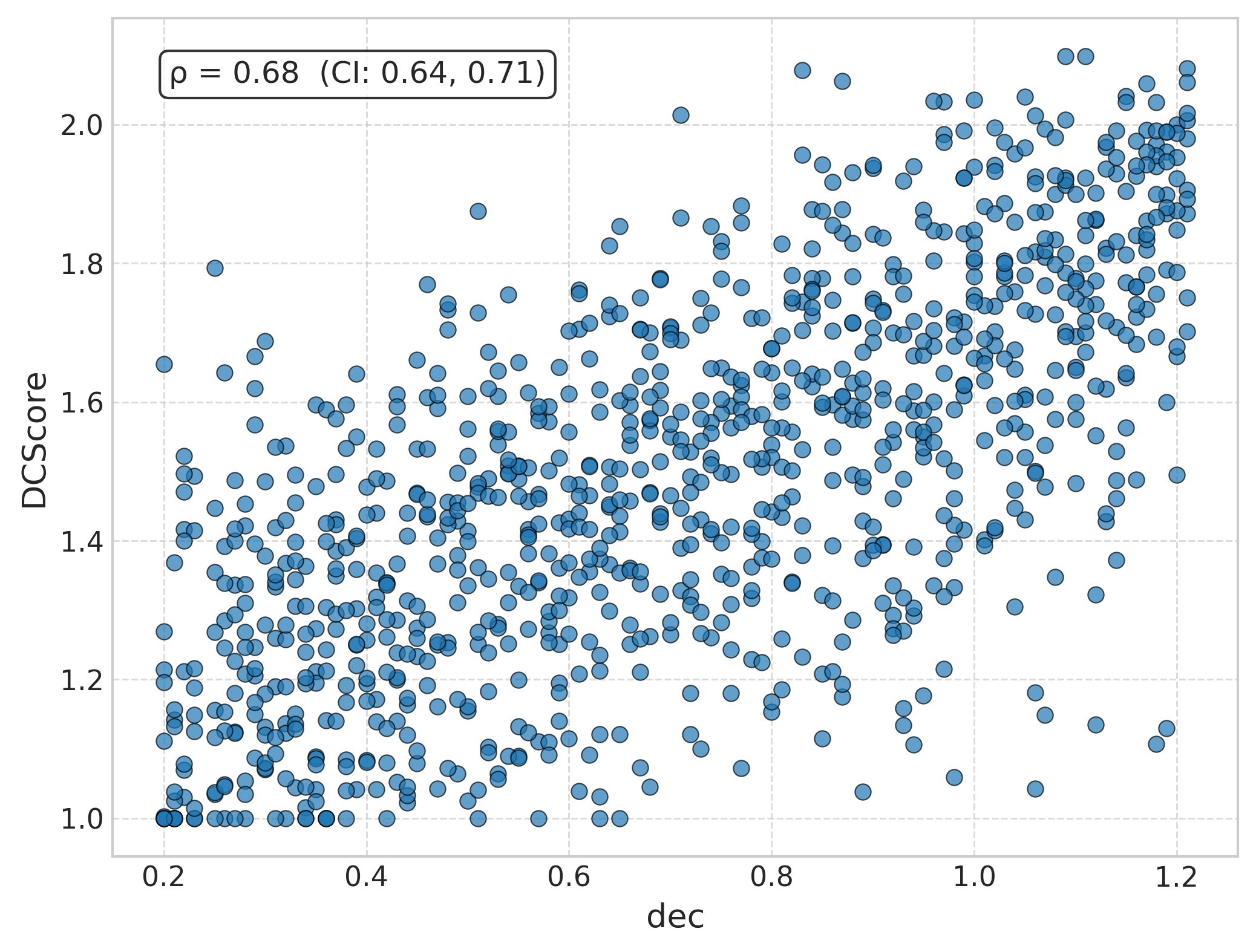}
    \includegraphics[width=0.24\textwidth]{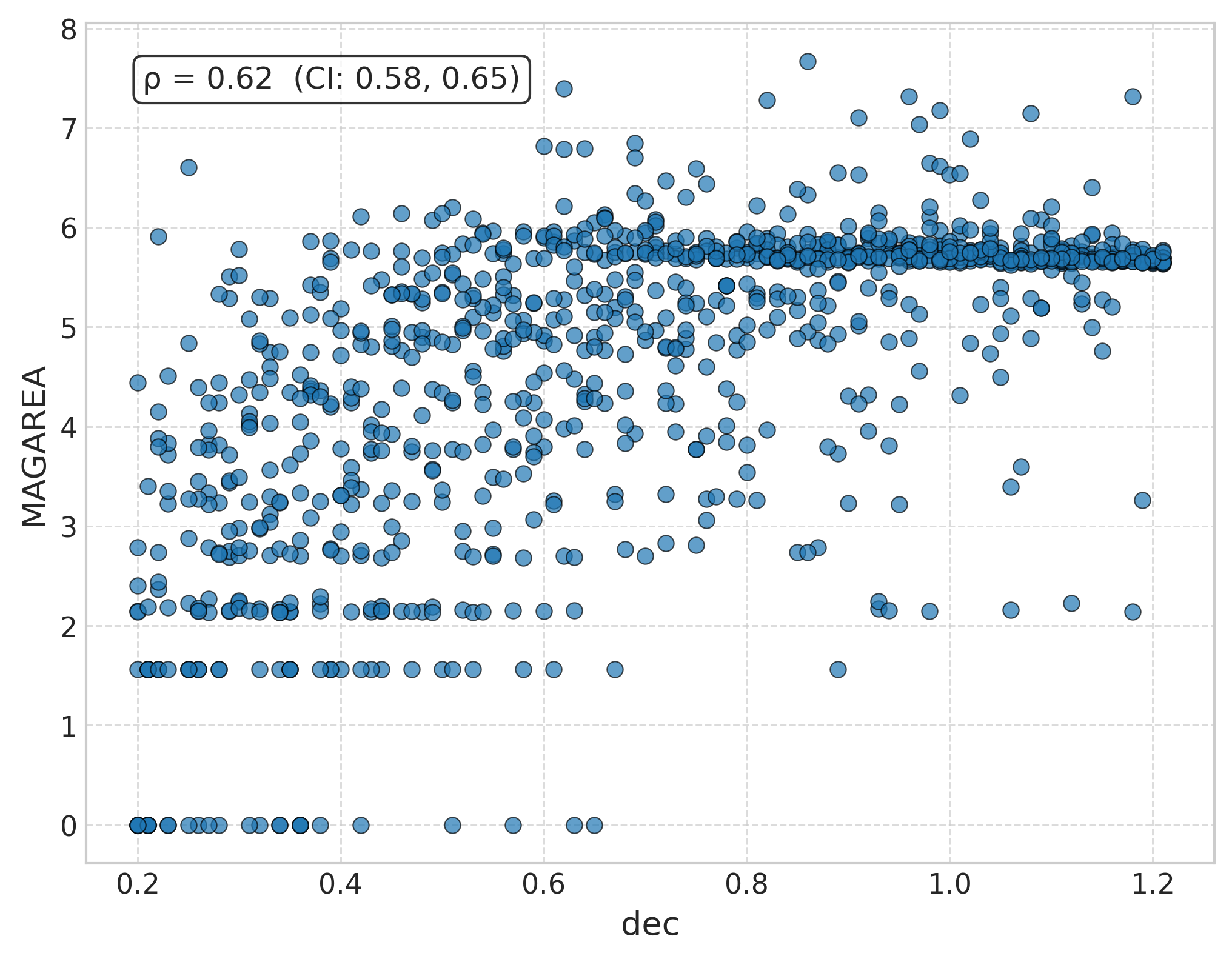}
    \caption{Correlation results for embeddings model: ``all-MiniLM-L12-v2" across three tasks: Row 1 shows prompt, Row 2 shows response, and Row 3 shows story. Columns 1–4 represent the results for PLDiv, VS, DCS, and MagArea, respectively.}
    \label{fig:minilm}
\end{figure*}

\begin{figure*}[h!]
    \centering
     \includegraphics[width=0.24\textwidth]{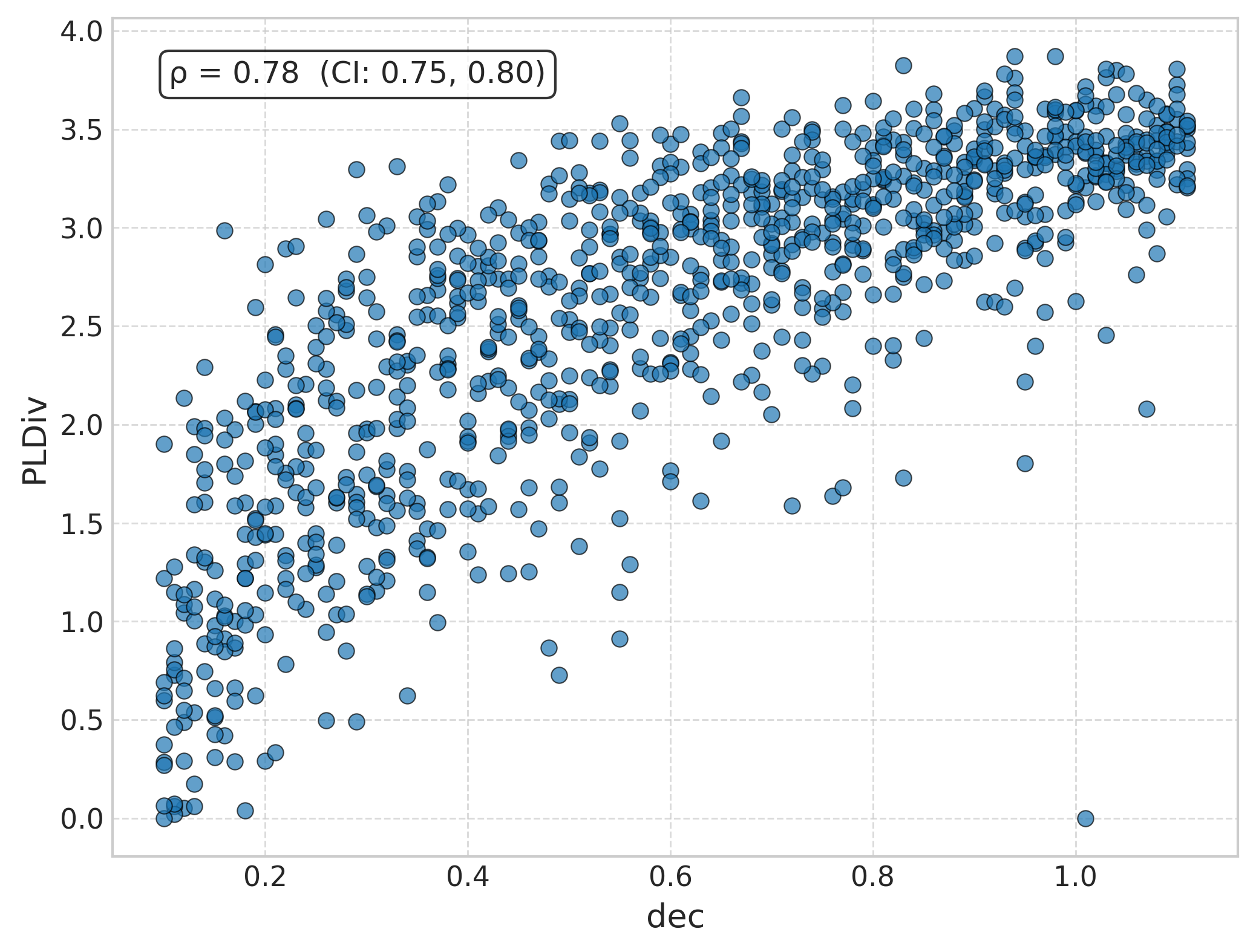}
    \includegraphics[width=0.24\textwidth]{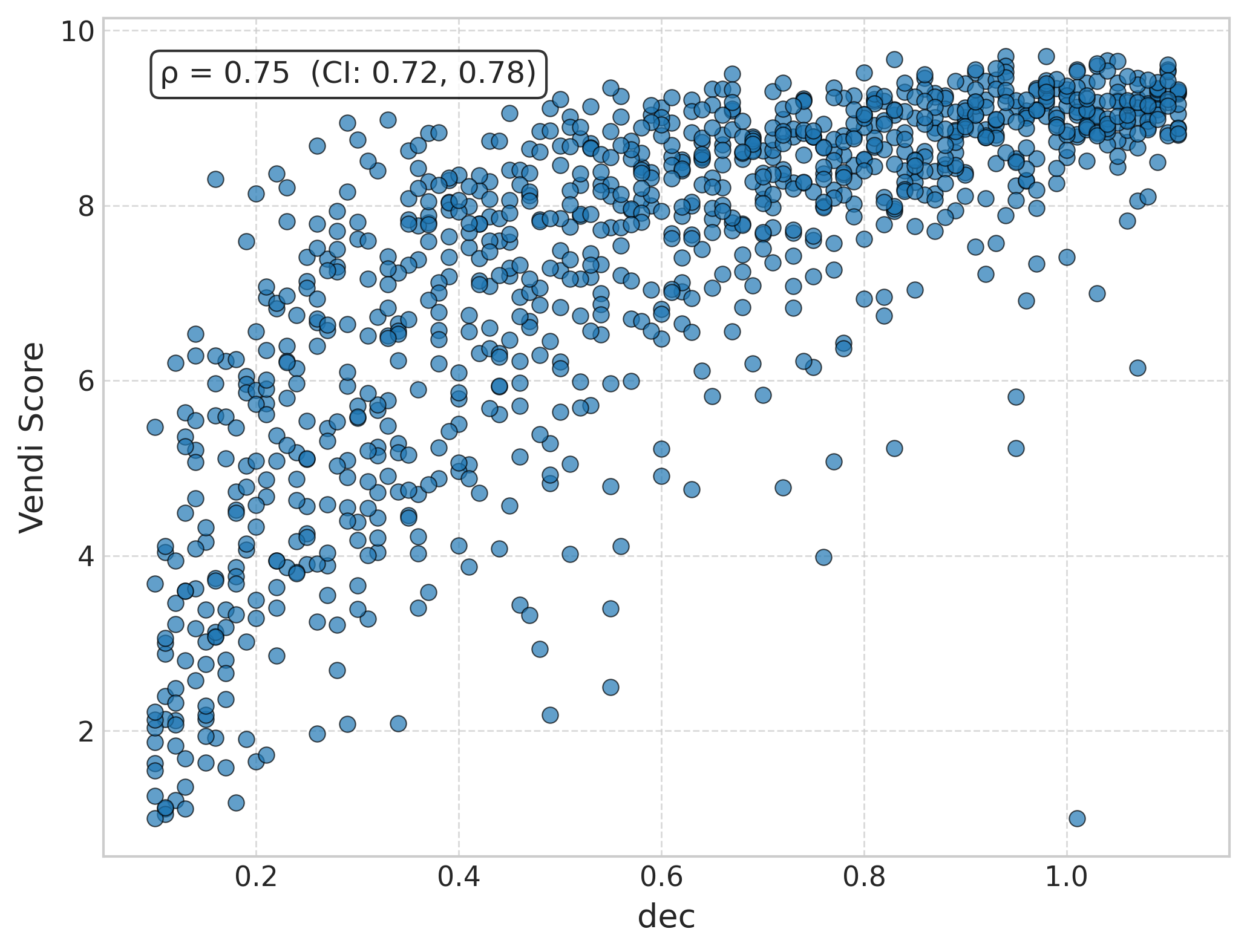}
    \includegraphics[width=0.24\textwidth]{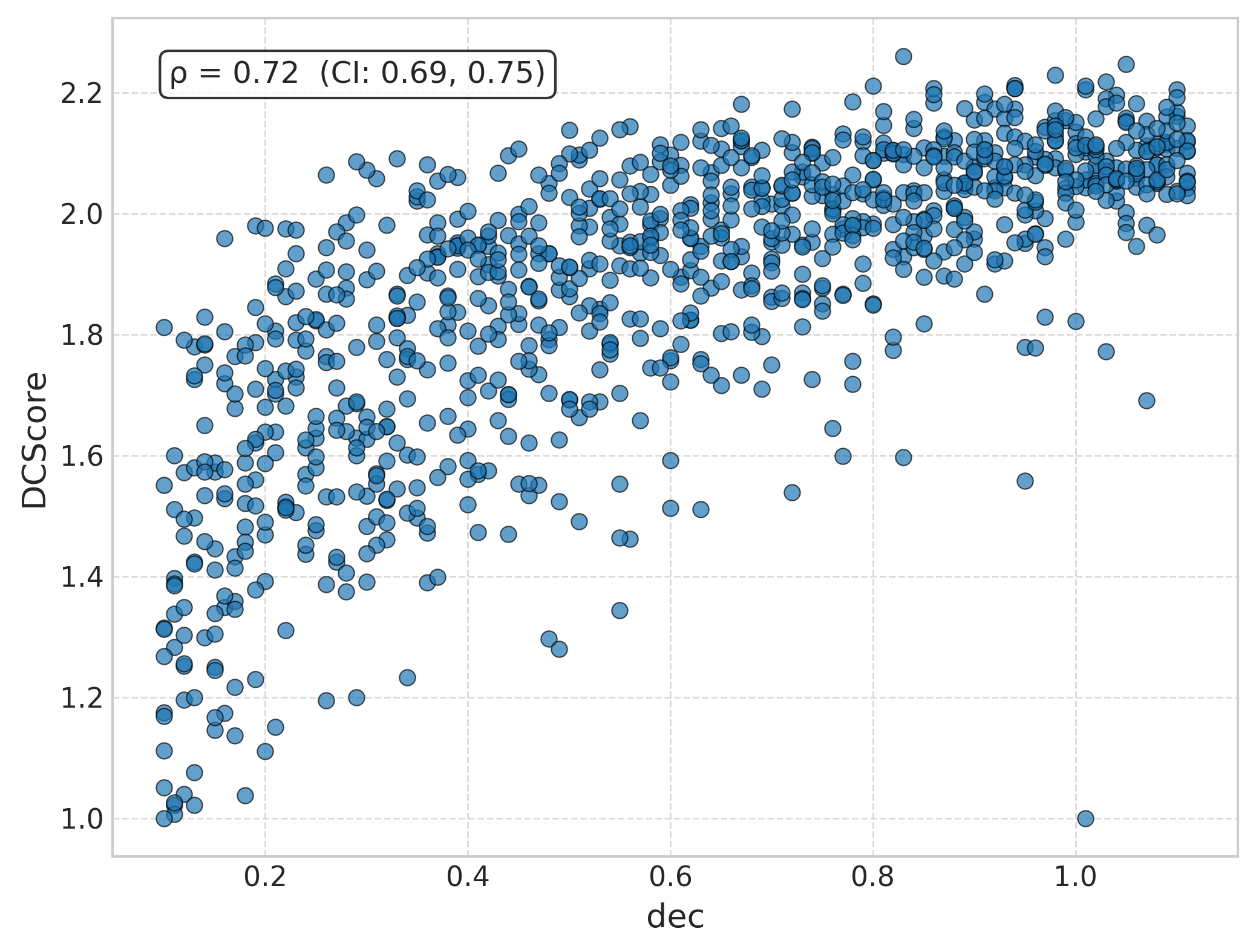}
    \includegraphics[width=0.24\textwidth]{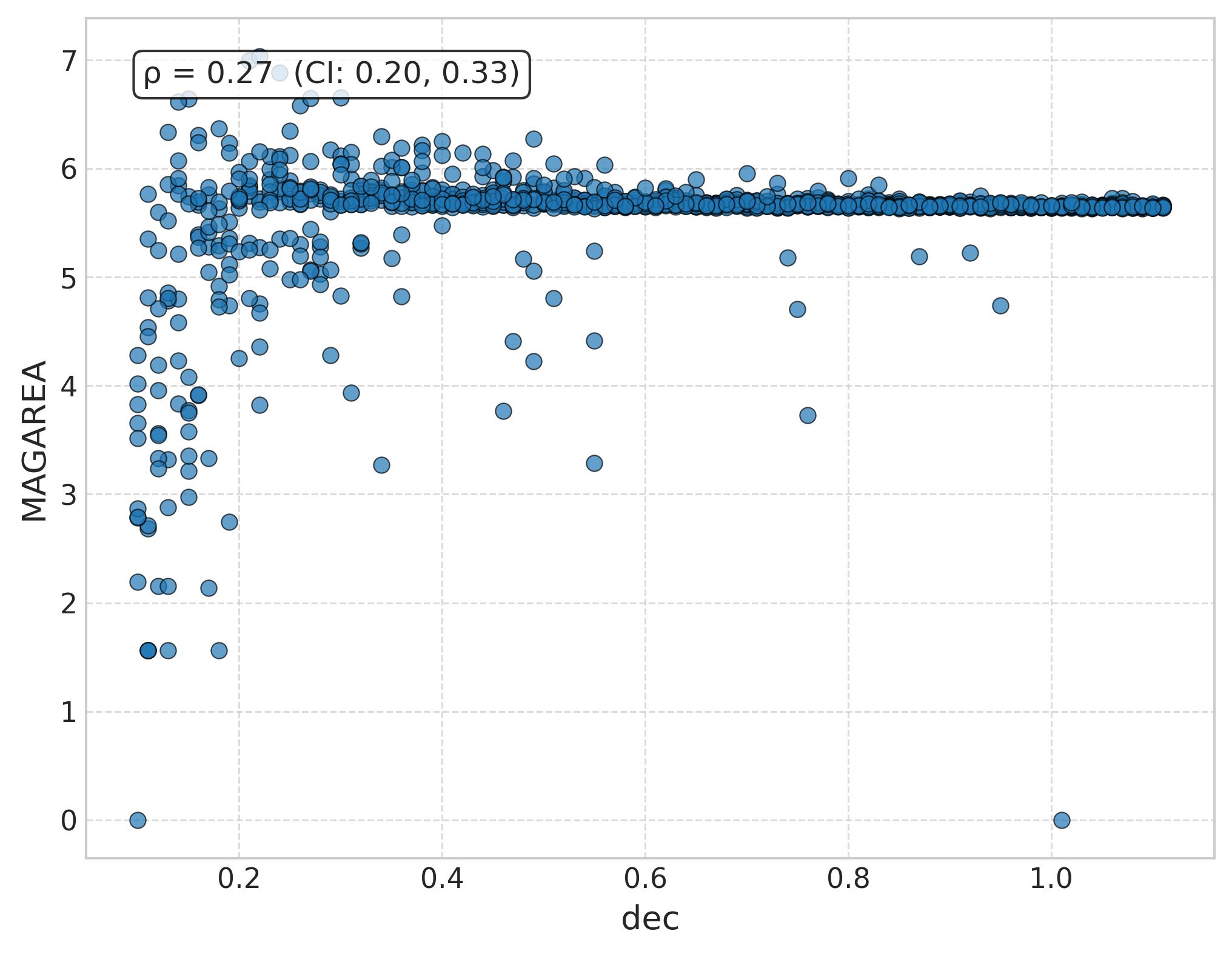}
    
    \vspace{2mm} 
    
     \includegraphics[width=0.24\textwidth]{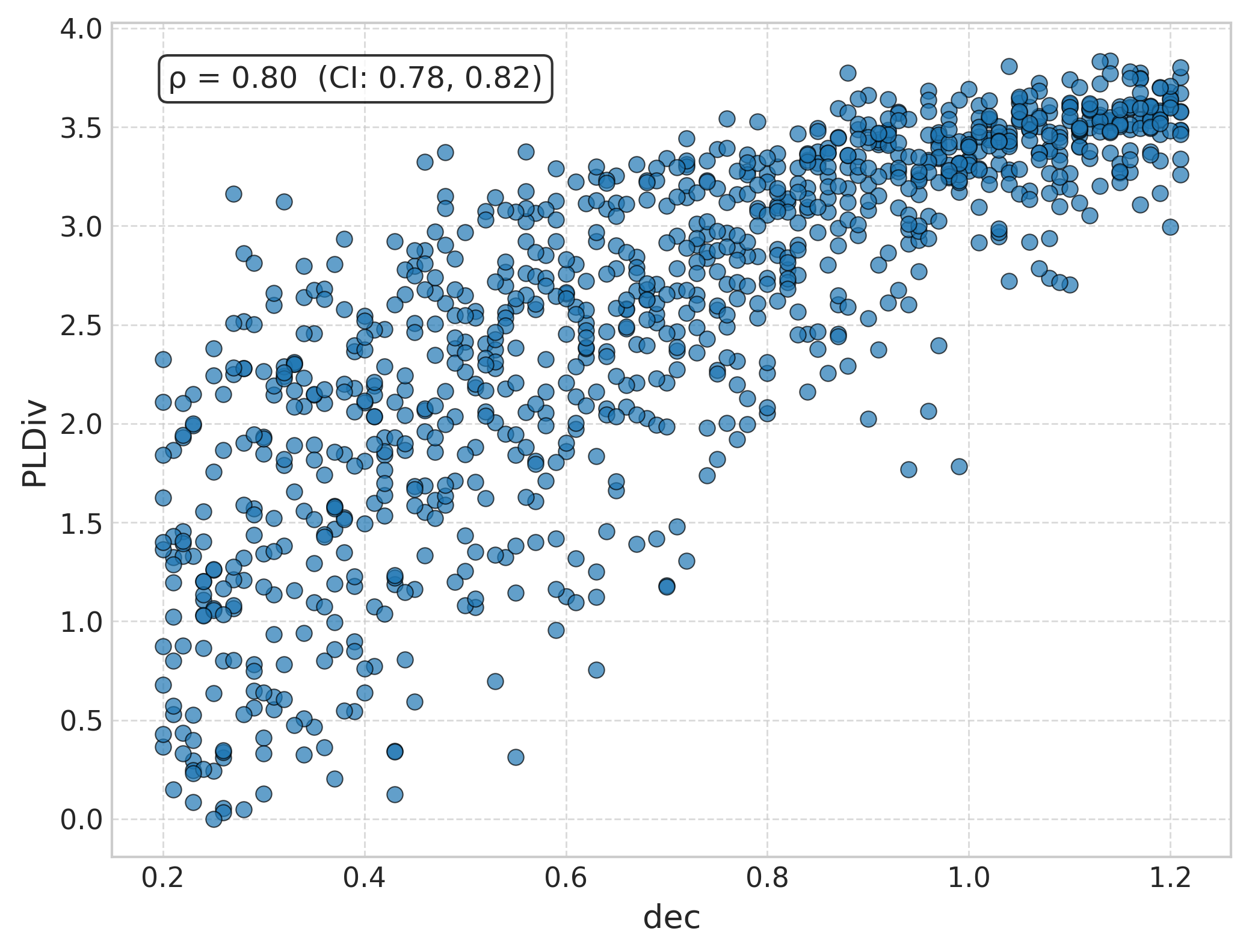}
    \includegraphics[width=0.24\textwidth]{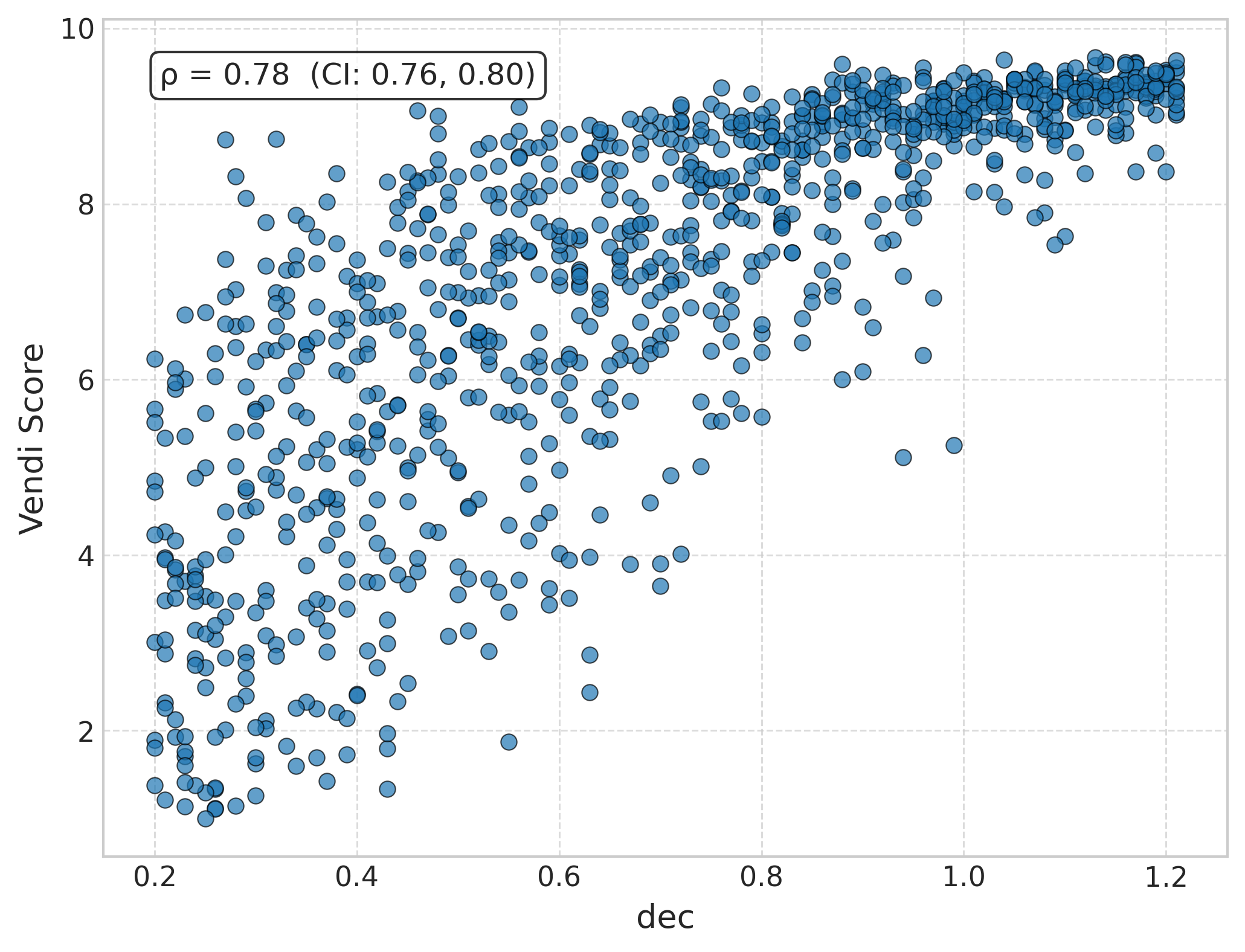}
    \includegraphics[width=0.24\textwidth]{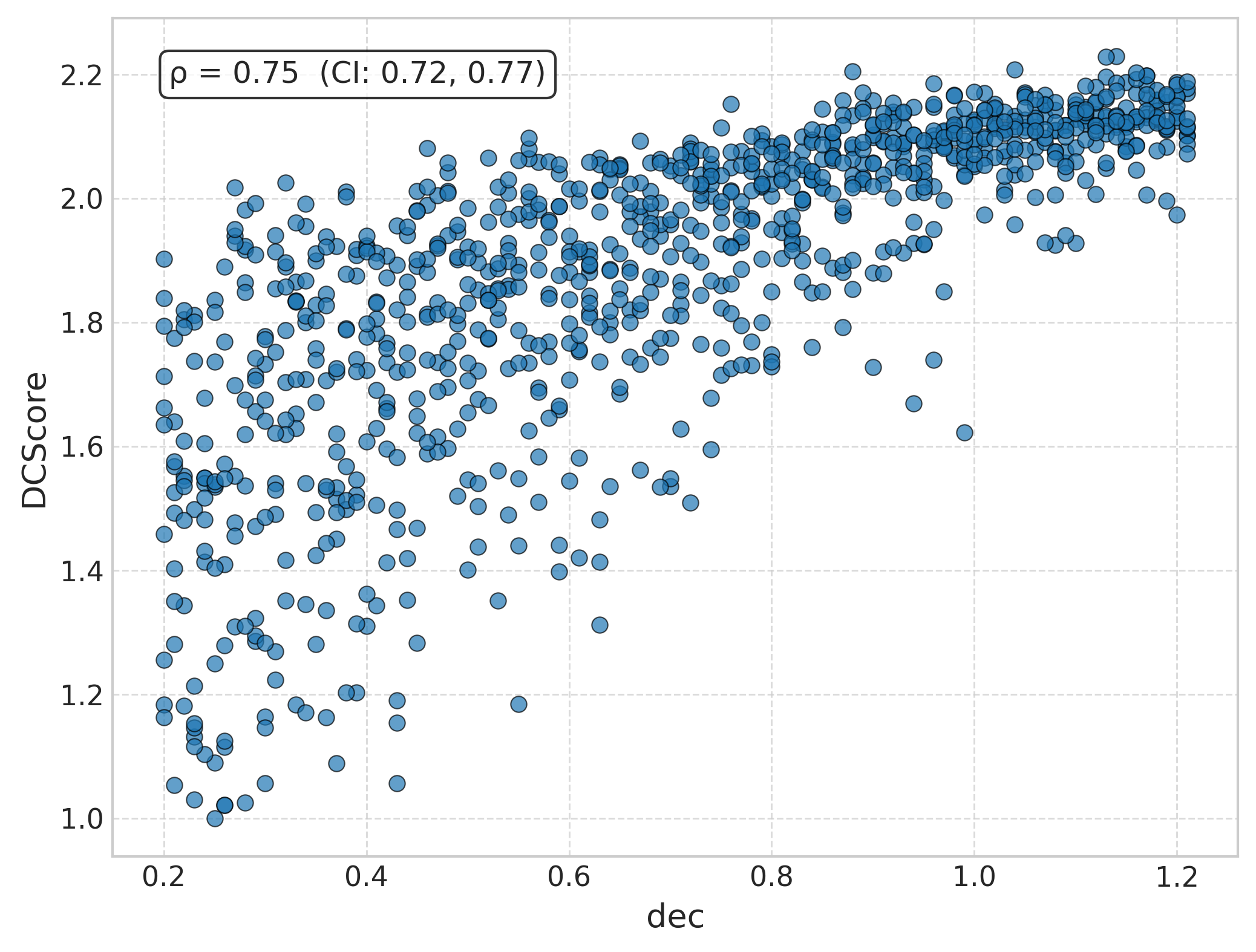}
    \includegraphics[width=0.24\textwidth]{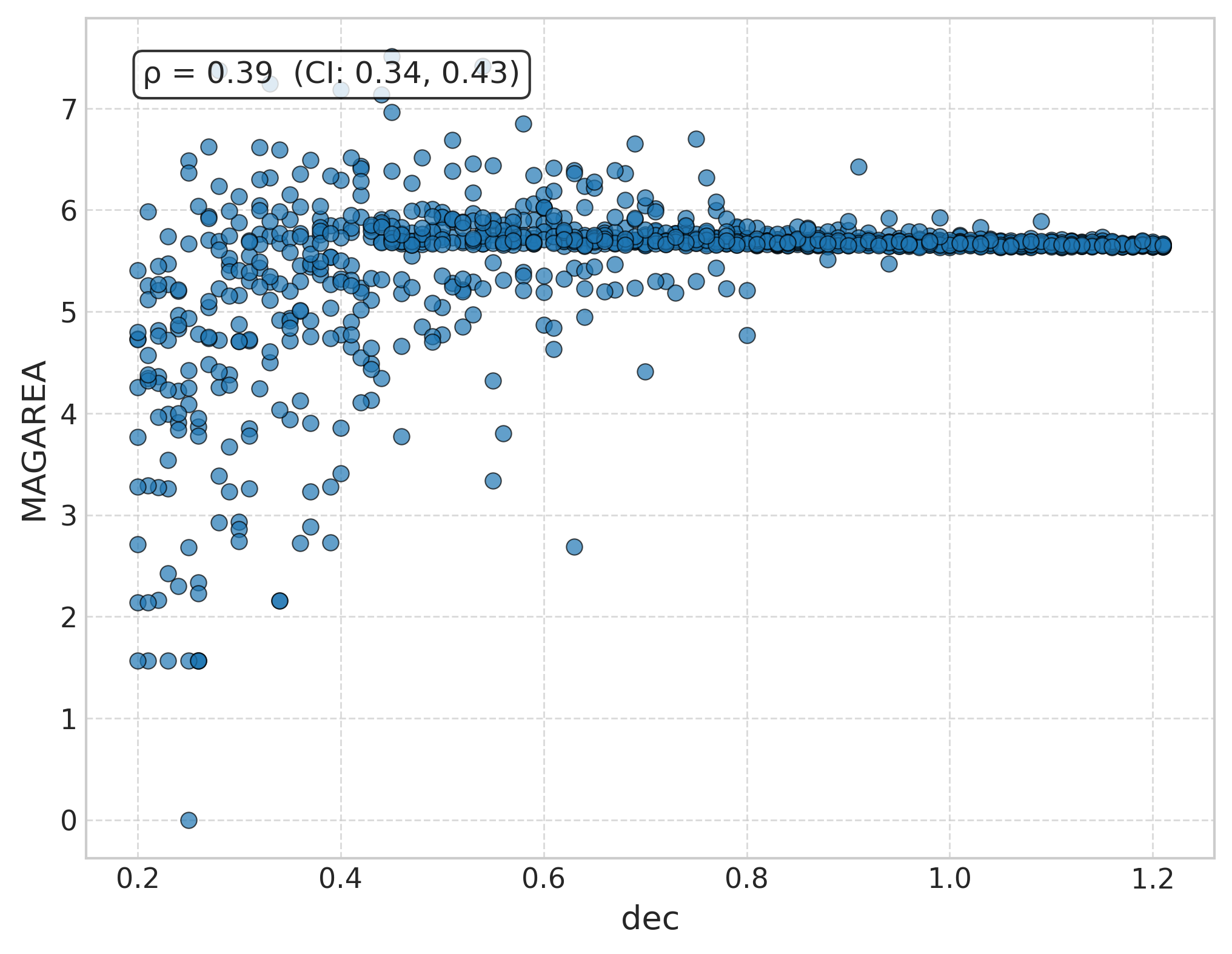}
    
    \vspace{2mm} 
    
     \includegraphics[width=0.24\textwidth]{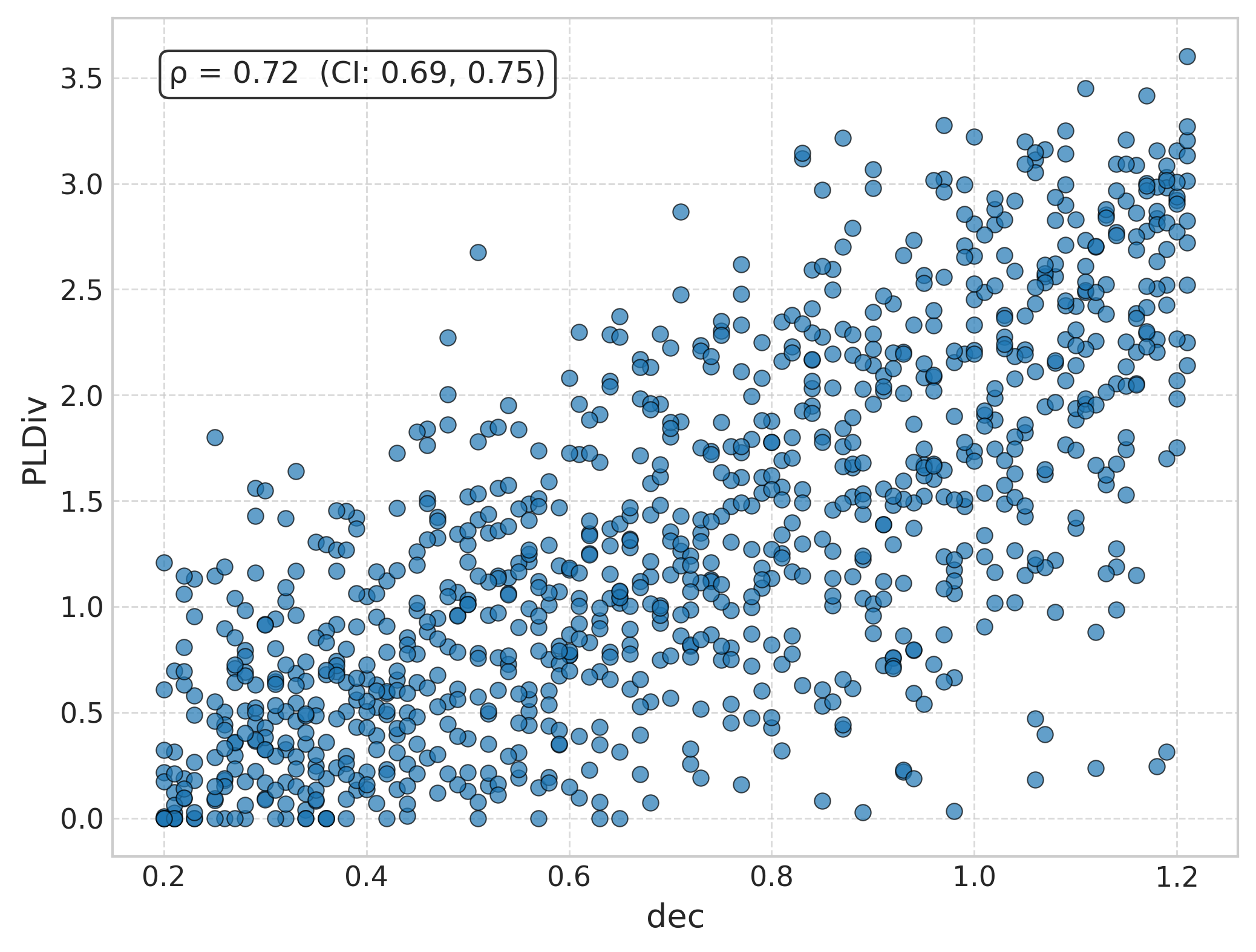}
    \includegraphics[width=0.24\textwidth]{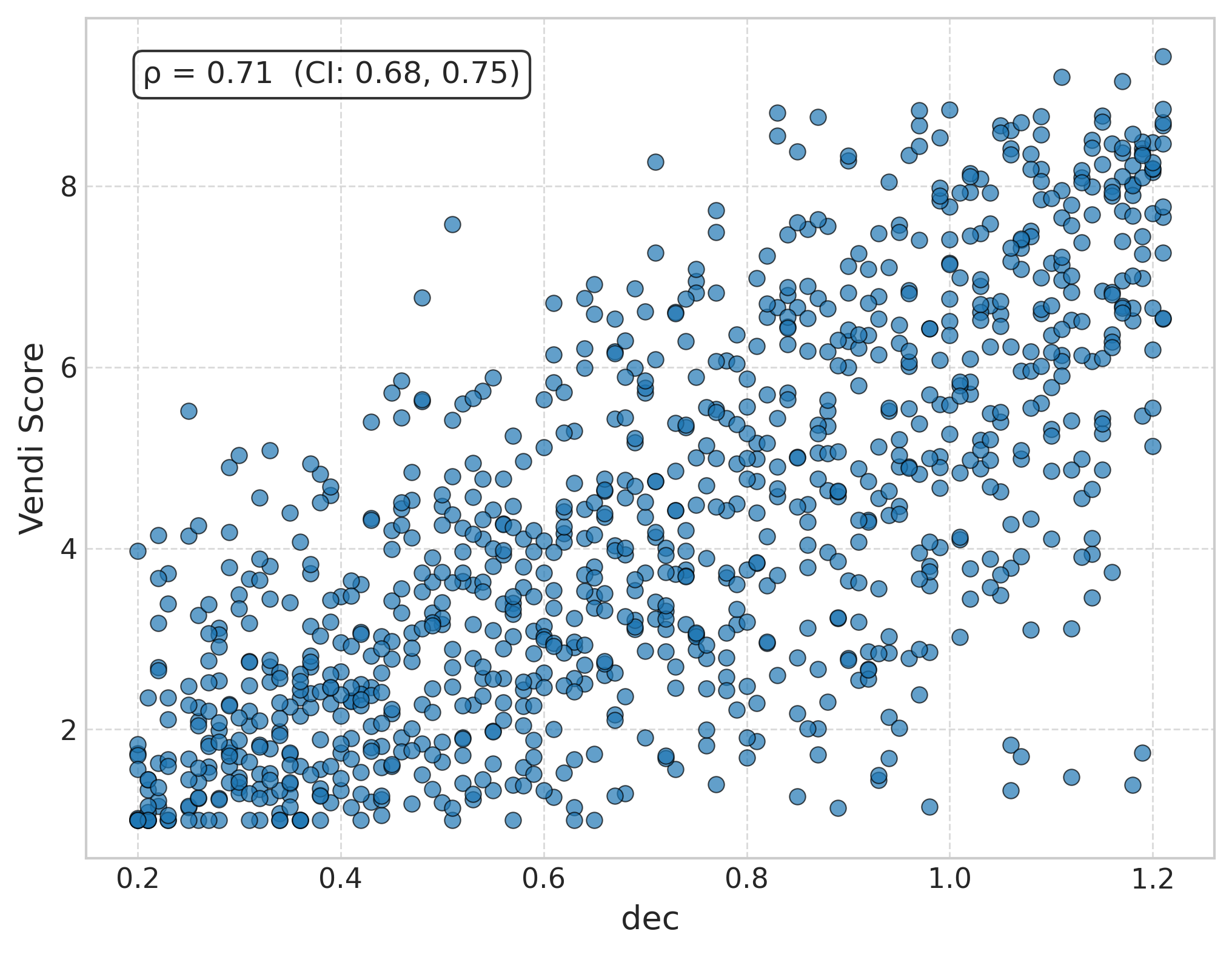}
    \includegraphics[width=0.24\textwidth]{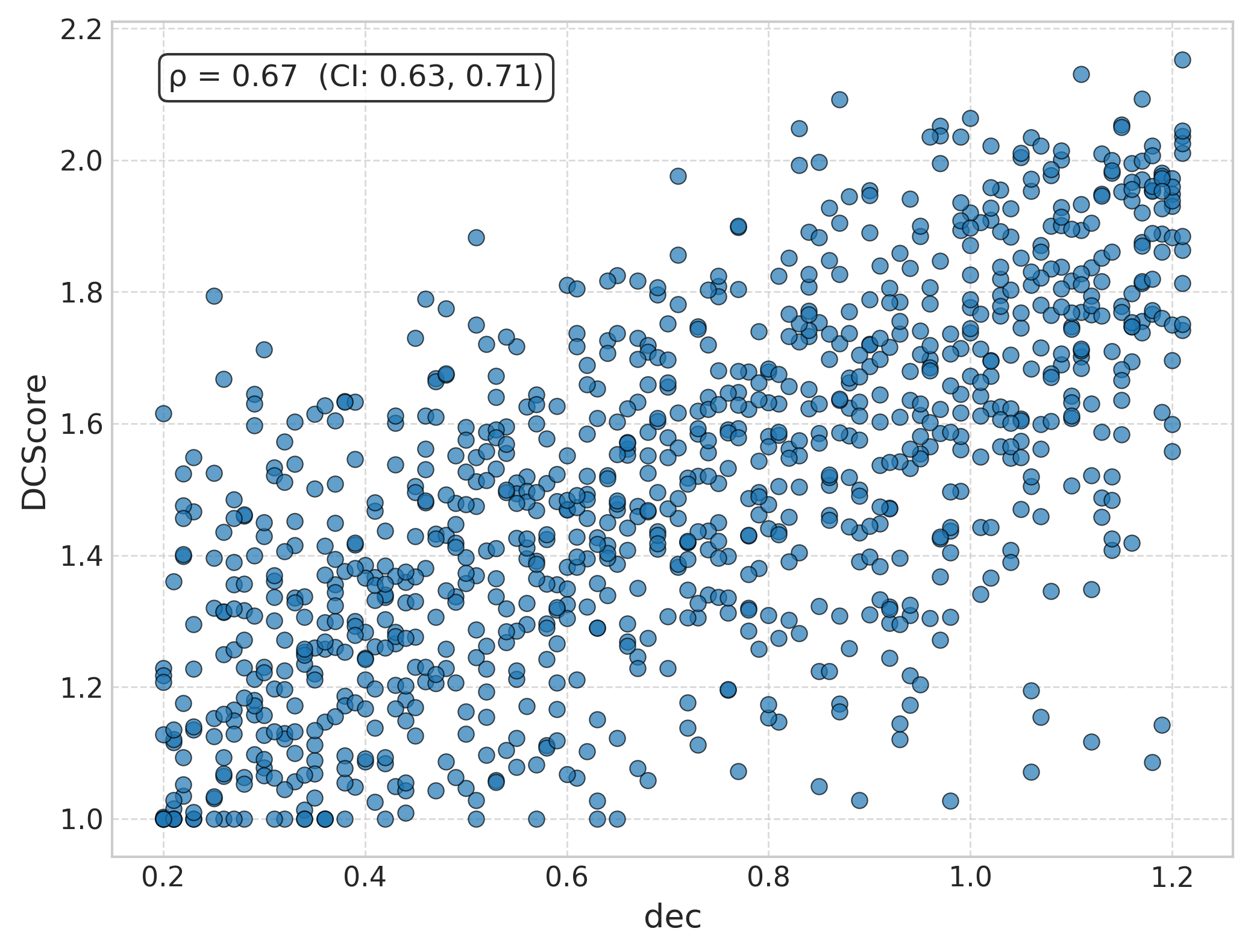}
    \includegraphics[width=0.24\textwidth]{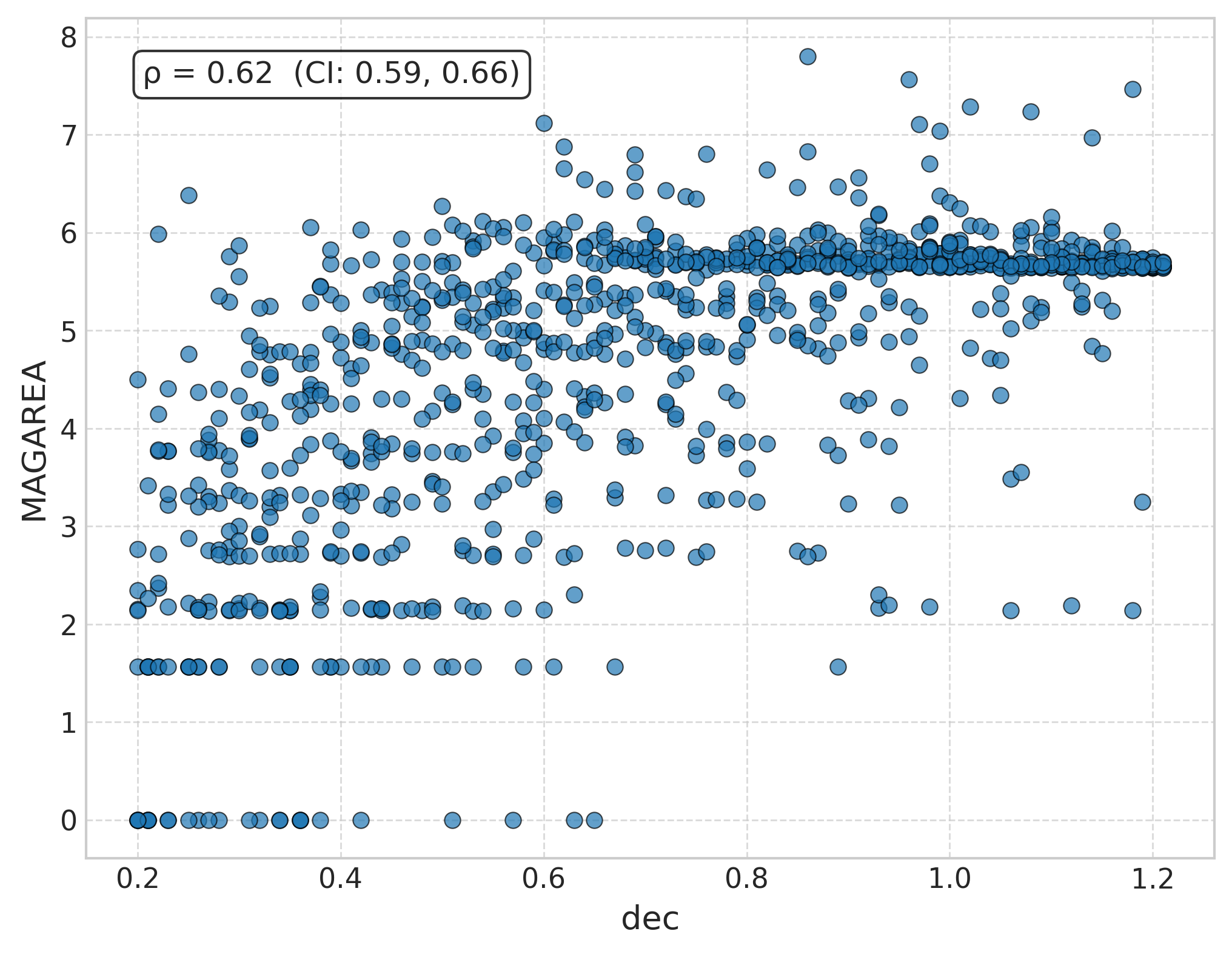}
    
    \caption{Correlation results for embeddings model: ``all-mpnet-base-v2" across three tasks: Row 1 shows prompt, Row 2 shows response, and Row 3 shows story. Columns 1–4 represent the results for PLDiv, VS, DCS, and MagArea, respectively.}
    \label{fig:mpnet}
\end{figure*}

We present the correlation plots for text embedding temperature $dec$ evaluation tasks in Figs. \ref{fig:bert}, \ref{fig:minilm}, and \ref{fig:mpnet}. Across the three embedding tasks, $\mathrm{PLDiv}$ shows the best performance on all three tasks: prompt, response, and story, exhibiting a linear relationship, while providing a non-linear relationship with softmax temperature $dec$ . 

\subsection{Implementation of Image Embeddings}

In Section 5.4, we evaluated the diversity measure to determine whether it can effectively capture the diversity introduced by the richness of labels.  We employed Colored MNIST \cite{deng2012mnist}.  Following the methodology of \citet{ospanov2024towards}, the number of labels served as the ground truth for diversity, where a higher label count signifies a more diverse set. We sampled half of the data from each class. Starting from class 1, we incrementally added samples from one additional class at a time, up to class 10, thereby forming 10 subsets. Comparisons are conducted against Vendi Score, Magnitude, and DCScore, using two embedding models: Inception V3 and ResNet-18.  All metrics are tested on cosine distance or cosine similarity. Figure \ref{fig:image} and Table \ref{tab:imag_corr} show that PLDiv can effectively capture diversity encoded in image embeddings.  PLDiv achieved comparable results with MagArea but is more computationally efficient.

\begin{table}[h!]
\centering
\caption{Pearson Correlation Comparison among diversity measures}
\label{tab:imag_corr}
\begin{tabular}{lcc}
\hline
\textbf{Metric} & \textbf{CLIP Model} & \textbf{Inception Model} \\ \hline
PLDiv & 0.998 & 0.998 \\
Vendi Score    & 0.371 & 0.222 \\
DCScore   & 0.901 & 0.984 \\
MagArea  & 0.997 & 0.998 \\ \hline
\end{tabular}
\end{table}

\section{Limitations}

While PLDiv demonstrates strong theoretical grounding and robust empirical performance across modalities, we acknowledge several limitations and areas for future improvement. First, computational cost is not the primary focus of this work. Although we proposed a sparse computation that significantly reduces both time and memory requirements, PLDiv remains computationally intensive than lightweight alternatives such as DCScore. Our contribution emphasizes accuracy and geometric faithfulness rather than speed, and we recognize that there is room for further algorithmic optimization. 

Second, PLDiv currently employs the Vietoris–Rips filtration as its default topological construction. While this choice offers broad applicability and simplicity, alternative filtrations, such as Čech, Alpha Complex, etc, may capture structure more effectively in specific domains. Exploring these variants could further increase the flexibility of PLDiv.

Third, PLDiv balances fine-grained local feature capture with preservation of global geometric structure, governed by the maximum-edge parameter. In our experiments, a single global setting was sufficient, though in other specific cases, this parameter may need tuning to balance local sensitivity and computational efficiency.



\section{Computational Environment}

All experiments were conducted on a high-performance computing server equipped with an AMD EPYC 7413 24-Core Processor and an NVIDIA A100-80GB GPU. The software environment was built using Python 3.11. For text embedding, we utilized Hugging Face Sentence Transformers as the embedding model framework.

\end{document}